\documentclass[runningheads]{llncs}

\usepackage{comment}
\usepackage{epsfig}
\usepackage{graphicx}
\usepackage{amsmath}
\usepackage{amssymb}
\usepackage{color}

\usepackage{tabularx,ragged2e}
\usepackage{booktabs,calc}
\usepackage{multirow}
\usepackage{hhline}
\usepackage{array}
\usepackage{float}
\usepackage[hyphens]{url}
\usepackage{makecell}
\usepackage[dvipsnames]{xcolor}
\usepackage{mathtools}
\usepackage{cite}
\usepackage{dsfont}
\usepackage{microtype}
\usepackage{subcaption}
\usepackage{comment}

\usepackage{float}
\usepackage{paralist}
\usepackage{siunitx}
\usepackage{pifont}
\usepackage{wrapfig}

\usepackage{enumitem}

\setlist[enumerate]{itemsep=0mm}
\setlist[itemize]{noitemsep, topsep=0pt}
\newcommand{\etal}{\textit{et al}.}
\newcommand{\ie}{\textit{i}.\textit{e}.}
\newcommand{\eg}{\textit{e}.\textit{g}.}

\usepackage[export]{adjustbox}
\usepackage{arydshln}

\usepackage[width=122mm,left=12mm,paperwidth=146mm,height=193mm,top=12mm,paperheight=217mm]{geometry}

\newcommand{\minus}{\scalebox{0.65}[1.0]{$-$}}

\newcommand{\figref}[1]{Fig.~\ref{#1}}
\newcommand{\tabref}[1]{Tab. ~\ref{#1}}
\newcommand{\V}[1]{\mathbf{#1}}
\newcommand{\R}[0]{\rm I\!R}

\newcommand{\rulesep}{\unskip\ \vrule width0.01cm\ }
\newcommand{\rulesepwhite}{\unskip\ \color{white}\vrule width0.01cm\ }

\newcommand{\cropparamhorz}{14cm}
\newcommand{\cropparamvert}{3.5cm}

\begin{document}
\pagestyle{headings}
\mainmatter
\def\ECCVSubNumber{2715}  %

\title{Weakly Supervised 3D Hand Pose Estimation \\ via Biomechanical Constraints}

\titlerunning{Weakly Supervised 3D Hand Pose Estimation via Biomechanical Constraints}
\author{Adrian Spurr\inst{1,2}* \and
Umar Iqbal\inst{2} \and
Pavlo Molchanov\inst{2} \and
\\Otmar Hilliges\inst{1} \and
Jan Kautz\inst{2}}
\authorrunning{A. Spurr et al.}
\institute{Advanced Interactive Technologies, ETH Zurich, Switzerland\\ 
\and
NVIDIA, Santa Clara, USA \\
\email{\{adrian.spurr, otmar.hilliges\}@inf.ethz.ch}\\
\email{\{uiqbal, pmolchanov, jkautz\}@nvidia.com}\\
}

\setlength{\intextsep}{0pt}

\maketitle
\captionsetup[figure]{font=small}
\captionsetup[table]{font=small}
\begin{abstract}
Estimating 3D hand pose from 2D images is a difficult, inverse problem due to the inherent scale and depth ambiguities. Current state-of-the-art methods train fully supervised deep neural networks with 3D ground-truth data. 
However, acquiring 3D annotations is expensive, typically requiring calibrated multi-view setups or labour intensive manual annotations. 
While annotations of 2D keypoints are much easier to obtain, how to efficiently leverage such \textit{weakly-supervised} data to improve the task of 3D hand pose prediction remains an important open question. 
The key difficulty stems from the fact that direct application of additional 2D supervision mostly benefits the 2D proxy objective but does little to alleviate the depth and scale ambiguities. 
Embracing this challenge we propose a set of novel losses that constrain the prediction of a neural network to lie within the range of biomechanically feasible 3D hand configurations. 
We show by extensive experiments that our proposed constraints significantly reduce the depth ambiguity and allow the network to more effectively leverage additional 2D annotated images.
For example, on the challenging freiHAND dataset, using additional 2D annotation without our proposed biomechanical constraints reduces the depth error by only $15\%$, whereas the error is reduced significantly by $50\%$ when the proposed biomechanical constraints are used. 

{\let\thefootnote\relax\footnote{{*This work was done during an internship at NVIDIA.}}}

\end{abstract}
\newcommand{\mysize}{\tiny}

\begin{figure*}[t]
    \centering
    
    \begin{subfigure}[b]{0.13\linewidth}
        \hspace{\linewidth}
    \end{subfigure}
    \begin{subfigure}[b]{0.13\linewidth}
        \hspace{\linewidth}
    \end{subfigure}
    \begin{subfigure}[b]{0.13\linewidth}
    \caption*{\textbf{Front view}}
        \hspace{\linewidth}
    \end{subfigure}
    \begin{subfigure}[b]{0.13\linewidth}
        \hspace{\linewidth}
    \end{subfigure}
    \begin{subfigure}[b]{0.13\linewidth}
        \hspace{\linewidth}
    \end{subfigure}
    \begin{subfigure}[b]{0.13\linewidth}
    \caption*{\textbf{Top view}}
        \hspace{\linewidth}
    \end{subfigure}
    \begin{subfigure}[b]{0.13\linewidth}
        \hspace{\linewidth}
    \end{subfigure}
    
    \begin{subfigure}[b]{0.13\linewidth}
        \includegraphics[height=1\linewidth]{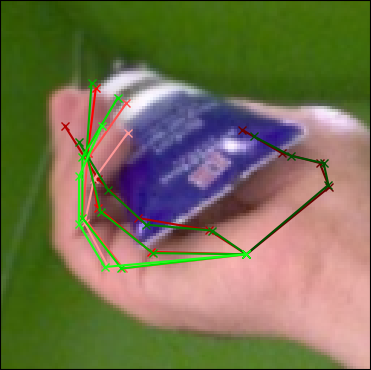}
    \end{subfigure}
    \begin{subfigure}[b]{0.13\linewidth}
        \includegraphics[trim={15.5cm 8cm 20cm 4cm},clip,width=\linewidth]{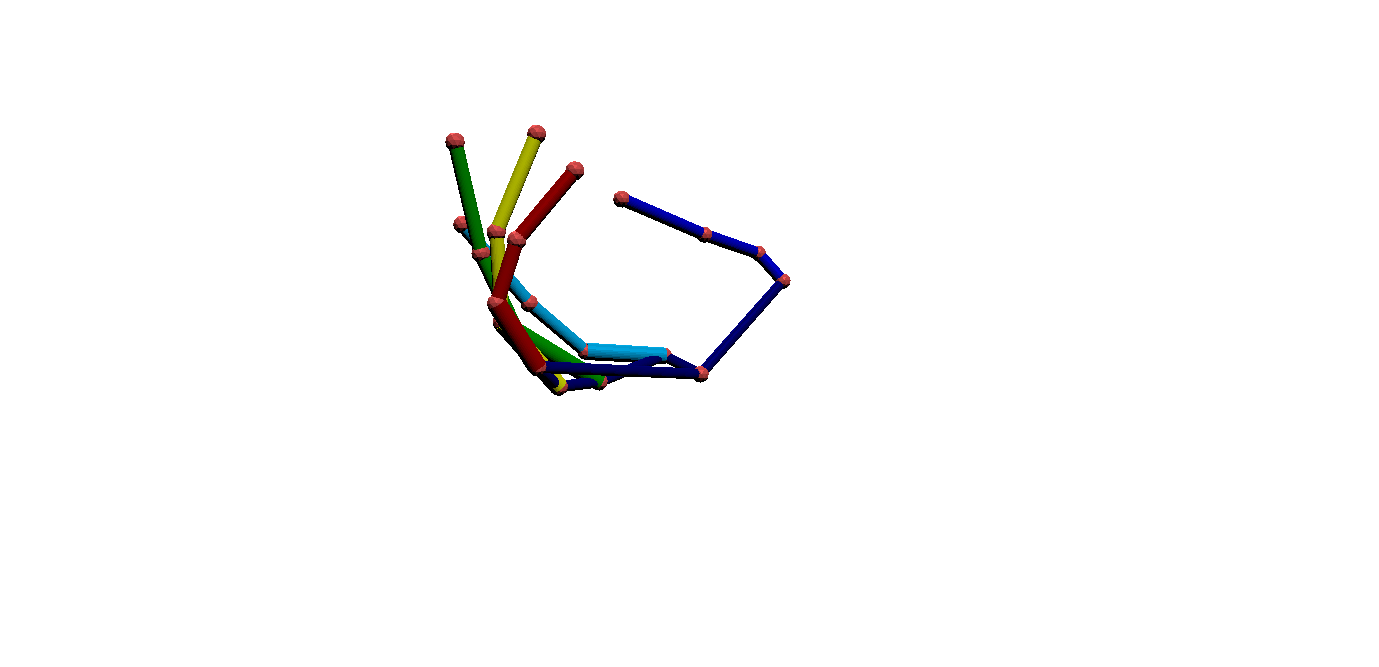}
    \end{subfigure}
    \begin{subfigure}[b]{0.13\linewidth}
        \includegraphics[trim={15.5cm 8cm 20cm 4cm},clip,width=\linewidth]{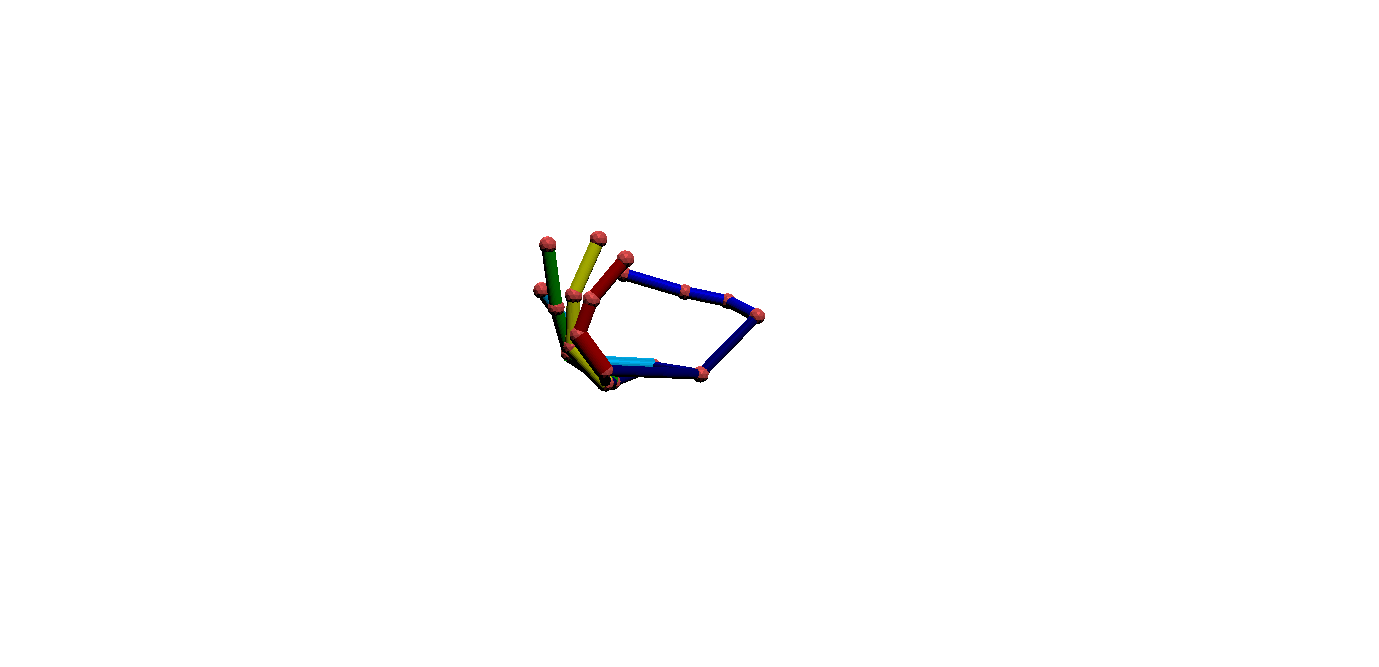}
    \end{subfigure}
    \begin{subfigure}[b]{0.13\linewidth}
        \includegraphics[trim={15.5cm 8cm 20cm 4cm},clip,width=\linewidth]{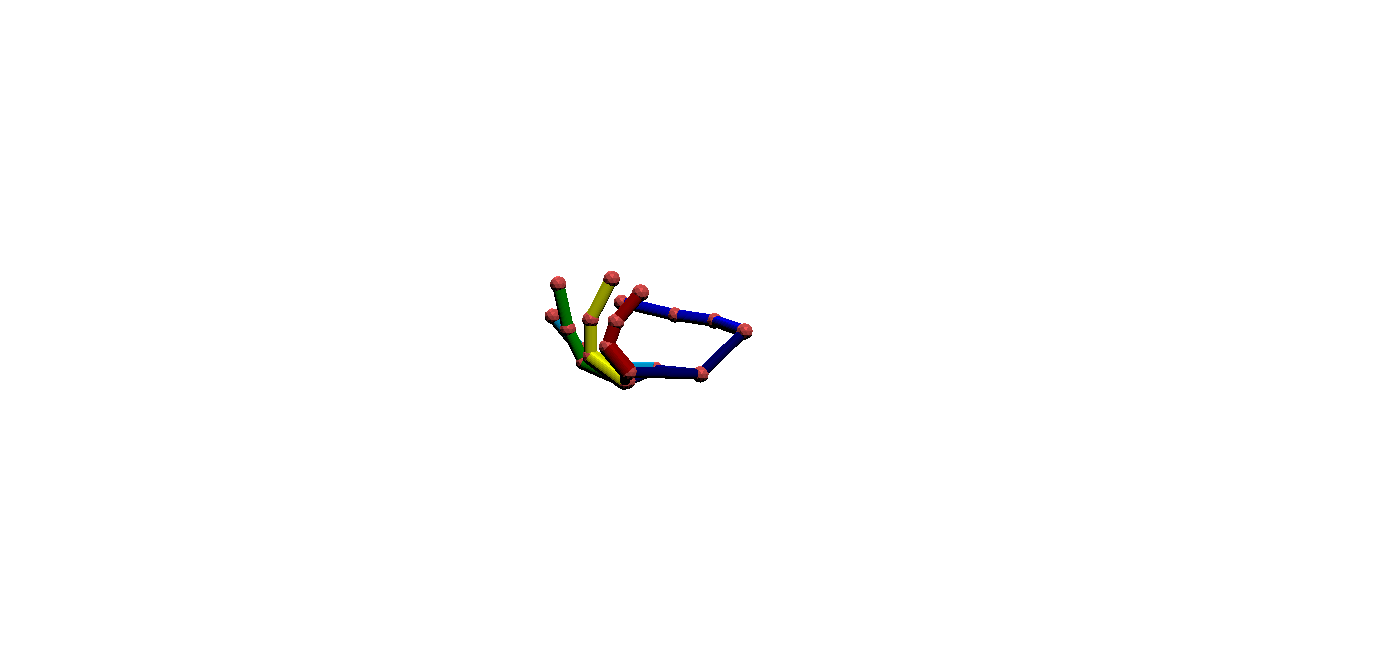}
    \end{subfigure}
     \rulesep 
    \begin{subfigure}[b]{0.13\linewidth}
        \includegraphics[trim={16cm 3.5cm 20cm 8.5cm},clip,width=\linewidth]{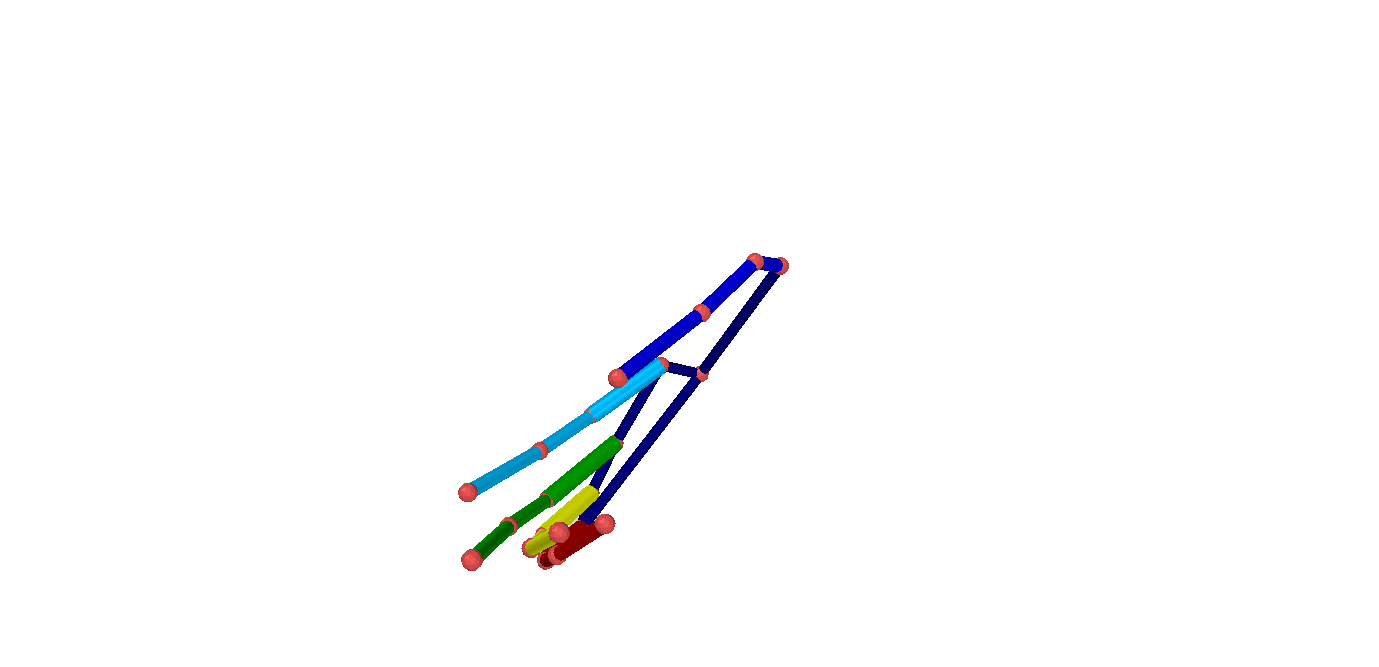}
    \end{subfigure}
    \begin{subfigure}[b]{0.13\linewidth}
        \includegraphics[trim={16cm 3.5cm 20cm 8.5cm},clip,width=\linewidth]{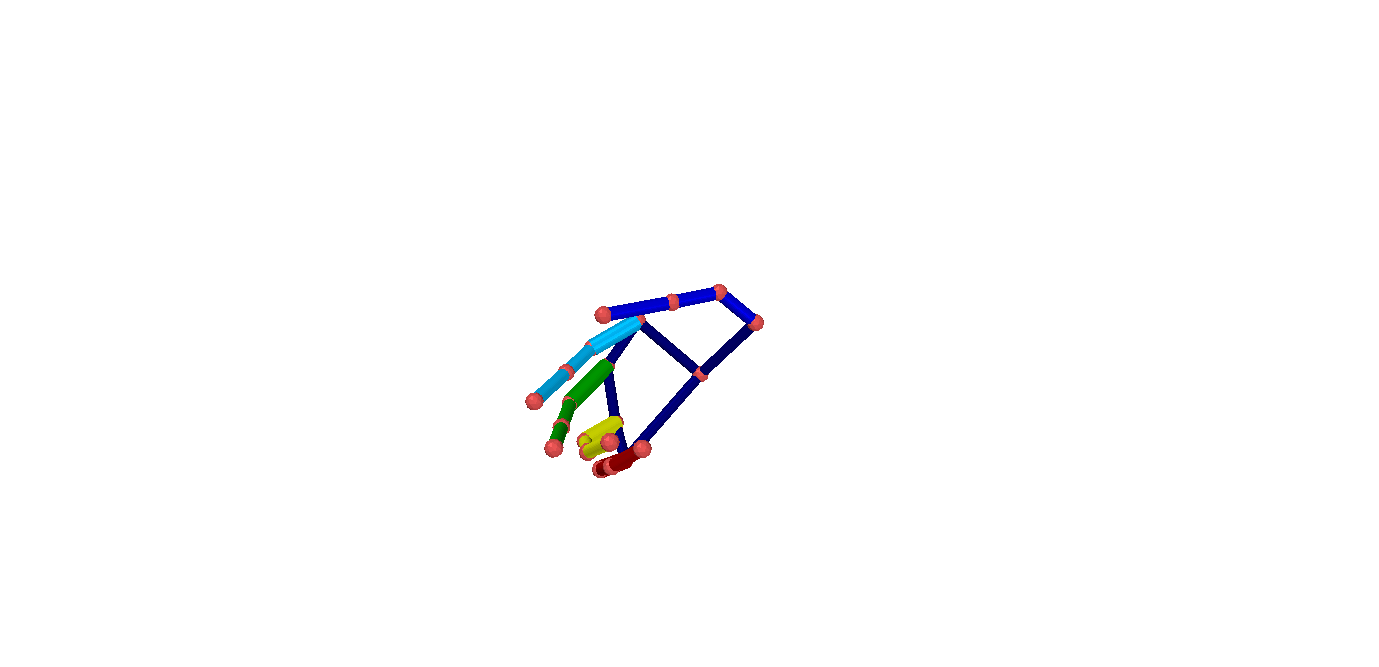}
    \end{subfigure}
    \begin{subfigure}[b]{0.13\linewidth}
        \includegraphics[trim={16cm 3.5cm 20cm 8.5cm},clip,width=\linewidth]{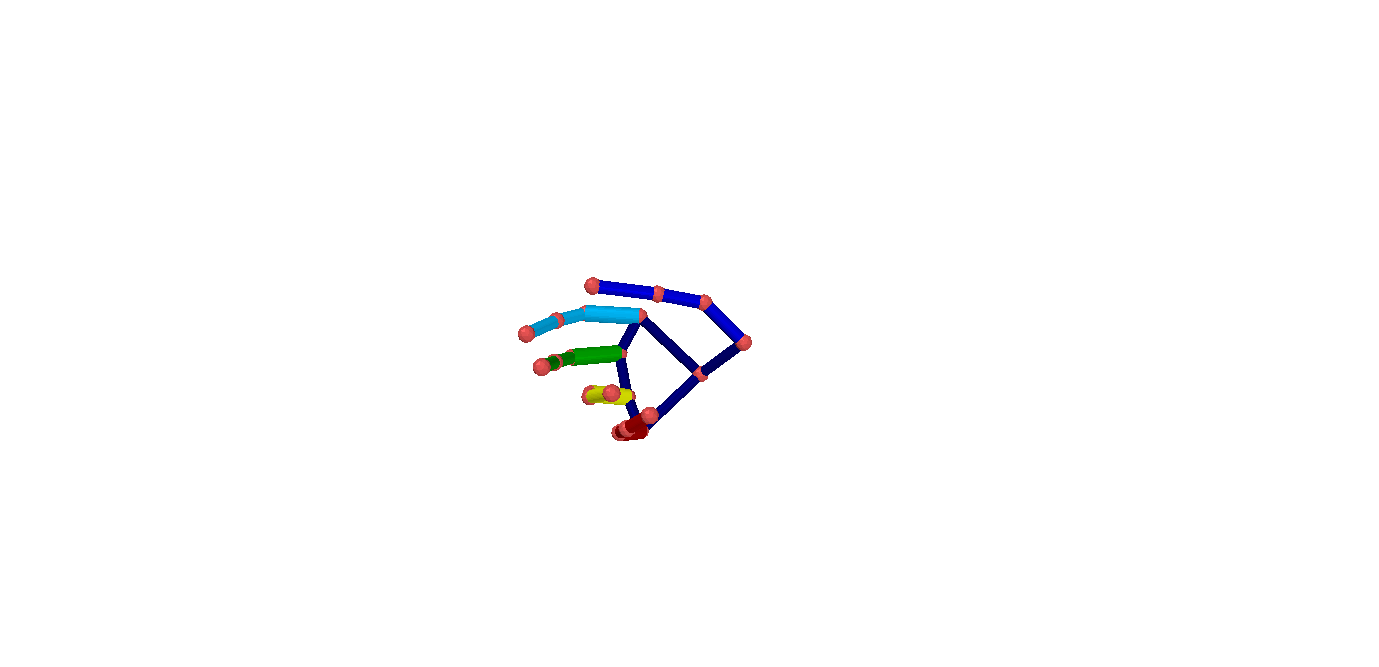}
    \end{subfigure}
    
    \begin{subfigure}[b]{0.13\linewidth}
        \includegraphics[height=1\linewidth]{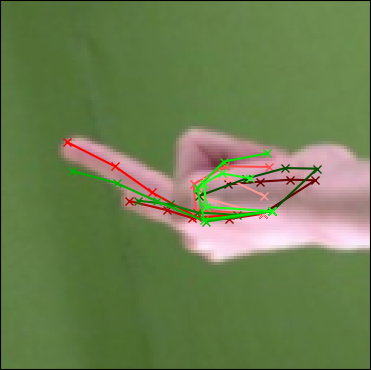}
    \end{subfigure}
    \begin{subfigure}[b]{0.13\linewidth}
        \includegraphics[trim={12cm 7cm 21.5cm 9cm},clip,width=\linewidth]{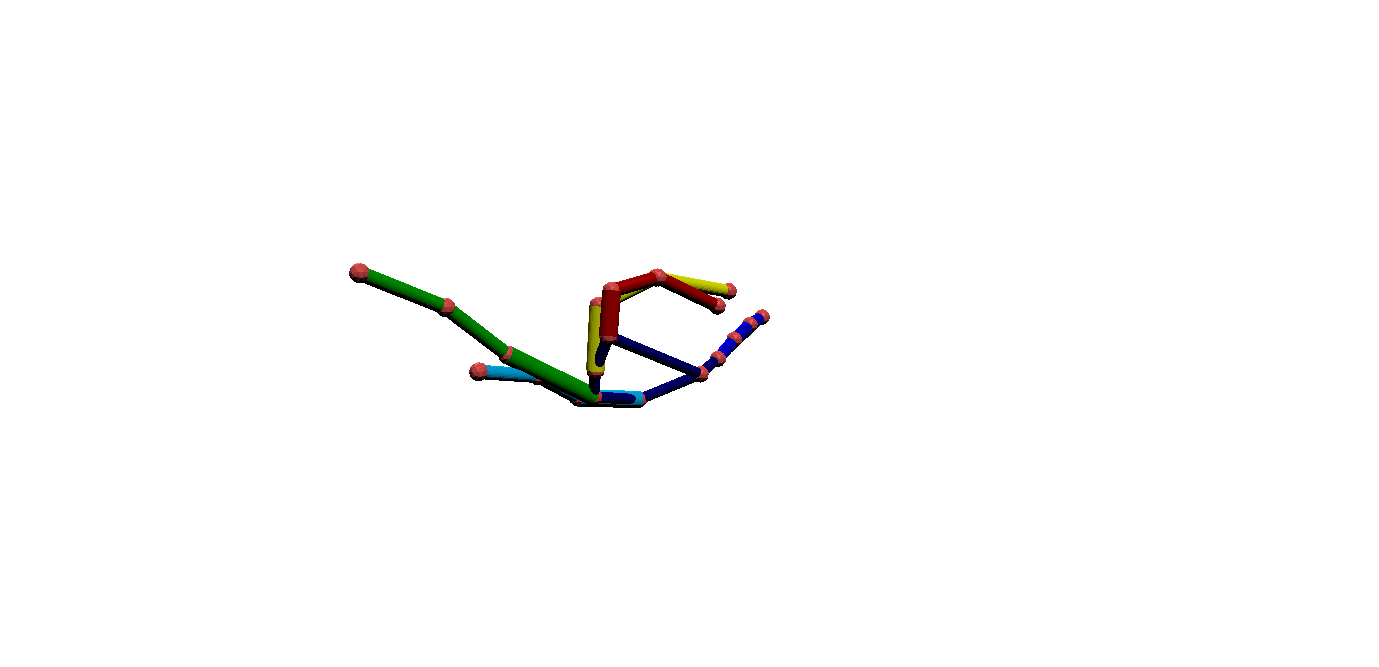}
        \label{fig:gt_1}
    \end{subfigure}
    \begin{subfigure}[b]{0.13\linewidth}
        \includegraphics[trim={12cm 7cm 21.5cm 9cm},clip,width=\linewidth]{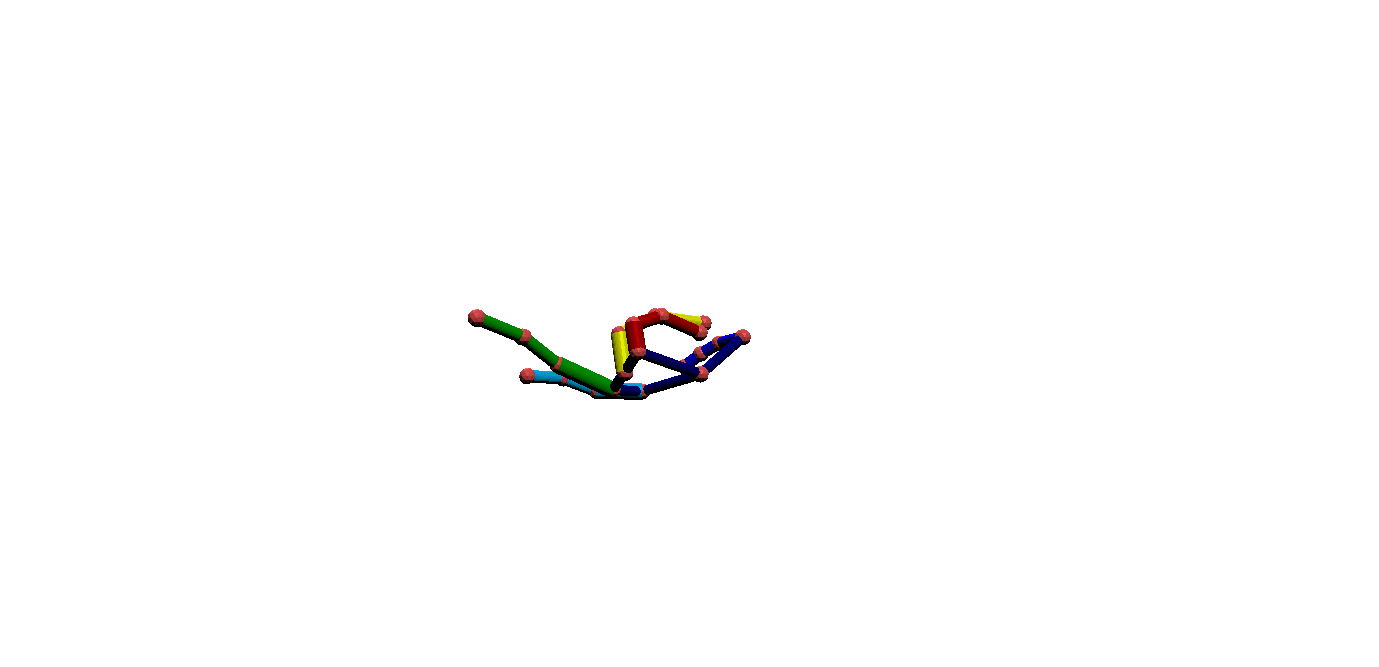}
        \label{fig:no_bmc_1}
    \end{subfigure}
    \begin{subfigure}[b]{0.13\linewidth}
        \includegraphics[trim={12cm 7cm 21.5cm 9cm},clip,width=\linewidth]{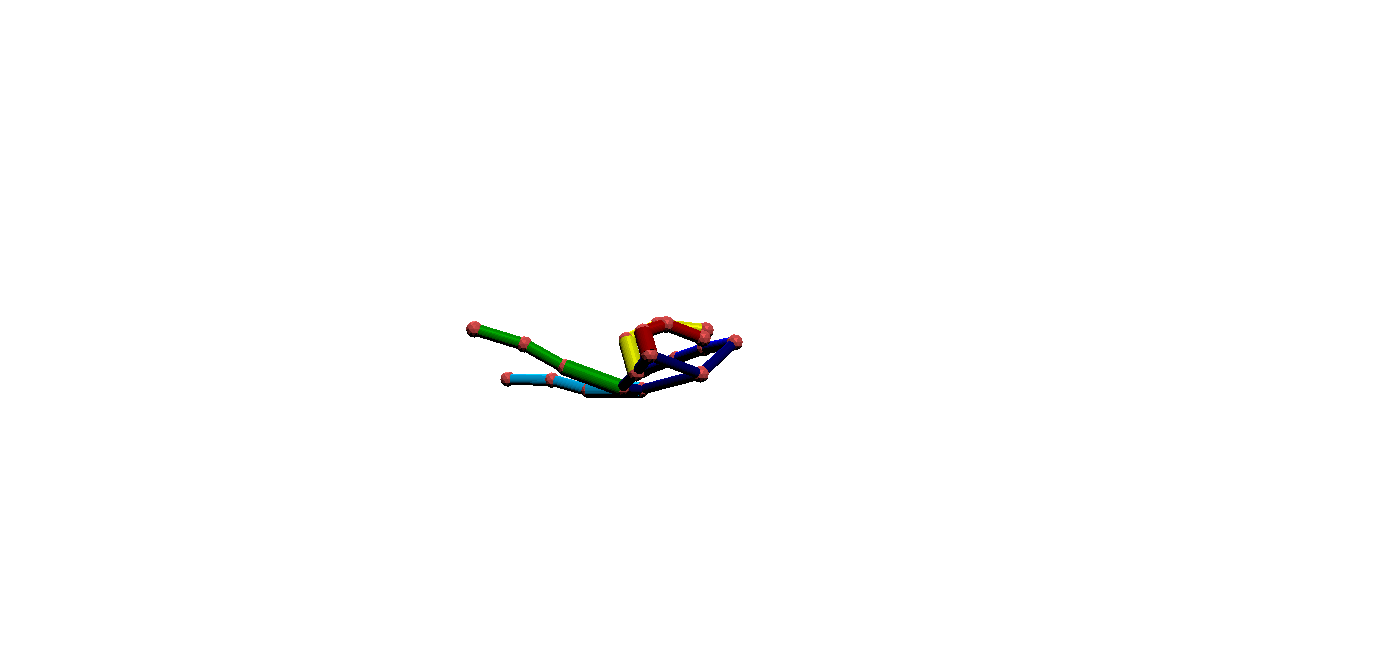}
        \label{fig:with_dhm_1}
    \end{subfigure}
      \rulesep 
    \begin{subfigure}[b]{0.13\linewidth}
        \includegraphics[trim={10cm 0.5cm 17cm 7cm},clip,width=\linewidth]{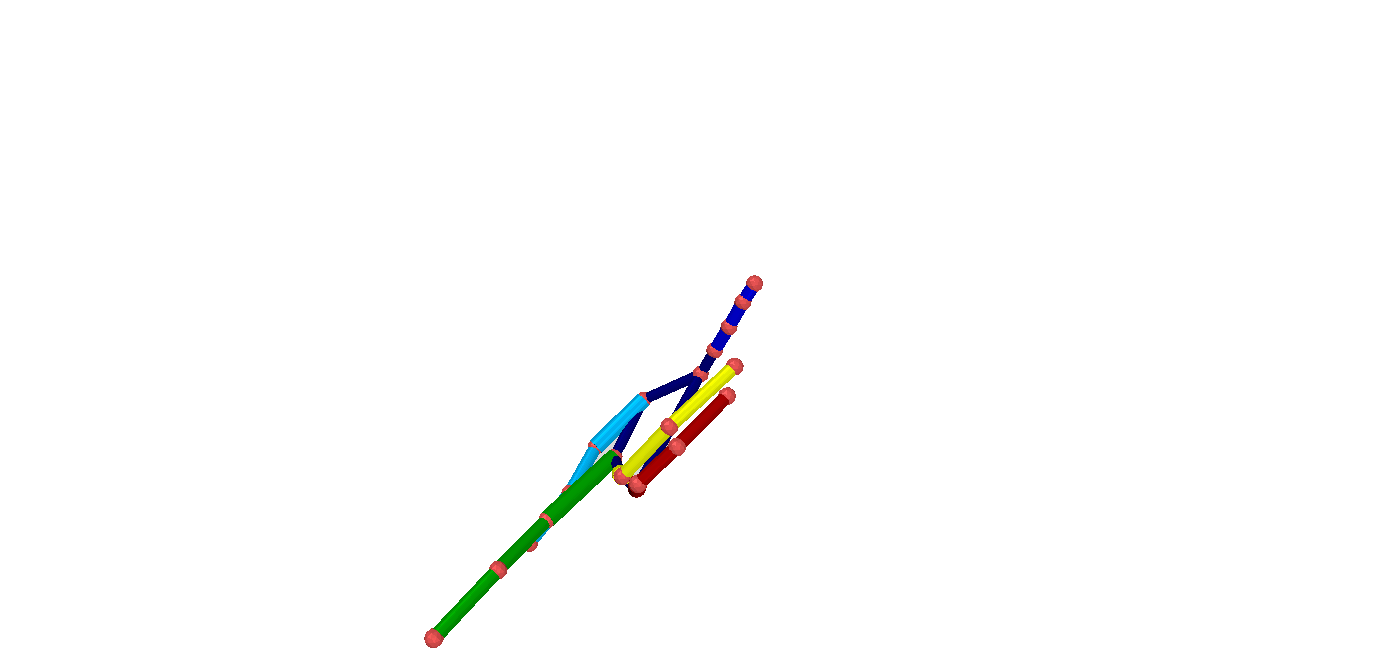}
        \label{fig:gt_2}
    \end{subfigure}
    \begin{subfigure}[b]{0.13\linewidth}
        \includegraphics[trim={10cm 0.5cm 17cm 7cm},clip,width=\linewidth]{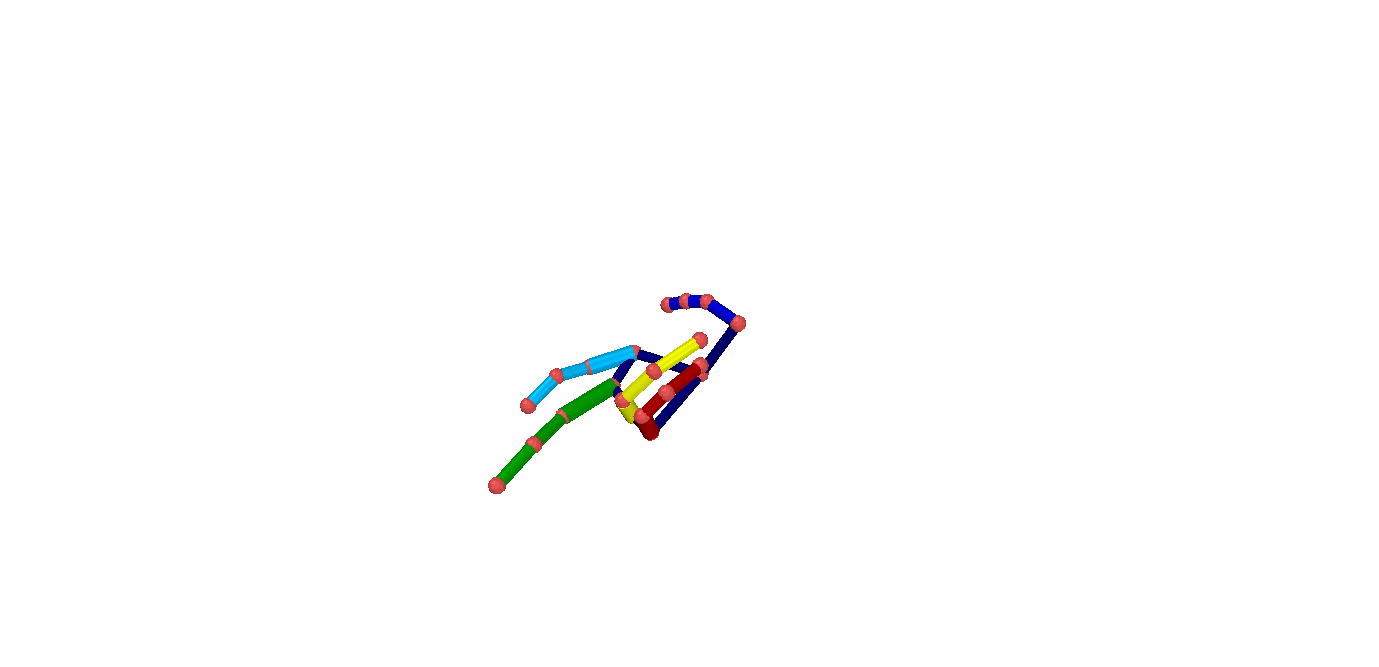}
        \label{fig:with_dhm_2}
    \end{subfigure}
    \begin{subfigure}[b]{0.13\linewidth}
        \includegraphics[trim={10cm 0.5cm 17cm 7cm},clip,width=\linewidth]{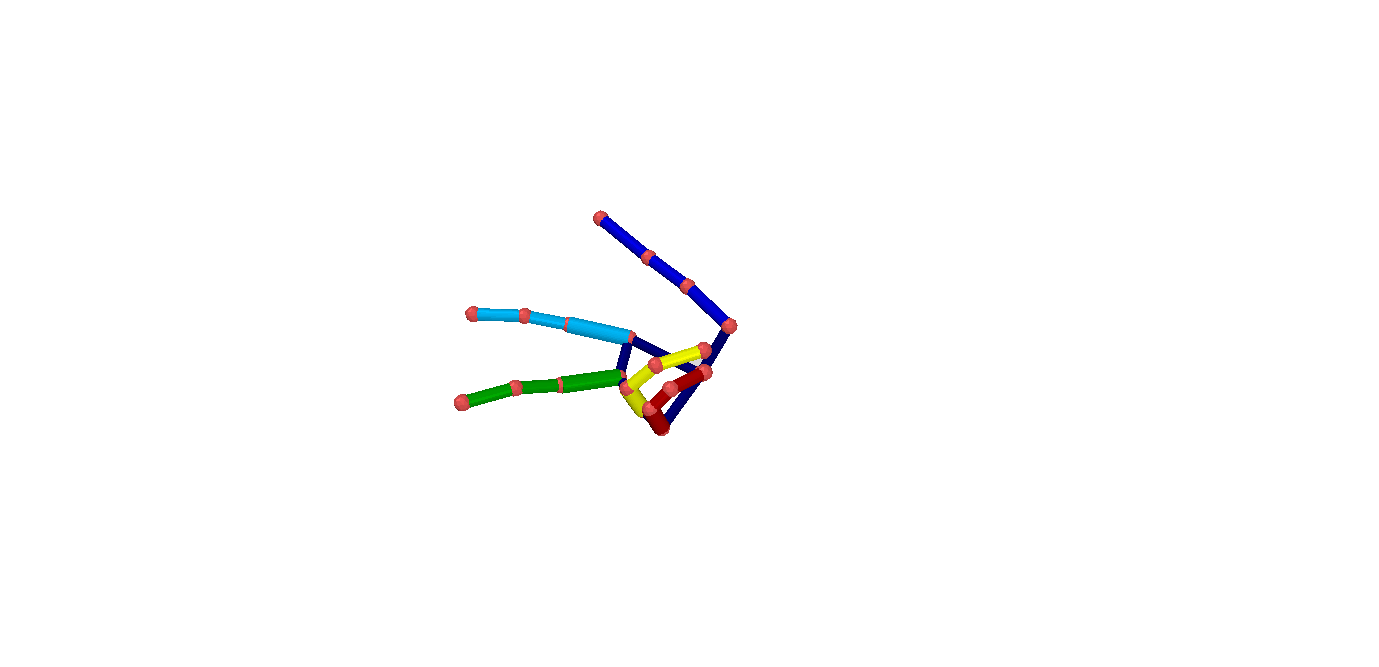}
        \label{fig:no_dhm_2}
    \end{subfigure}
    
    \begin{subfigure}[b]{0.13\linewidth}
        \hspace{\linewidth}
        \caption{\mysize $\hat{\V{J}}^{2D}$}
        \label{fig:rgb}
    \end{subfigure}
    \rulesepwhite
    \begin{subfigure}[b]{0.13\linewidth}
        \hspace{\linewidth}
        \caption{{\mysize w/o $\mathrm{\mathbf{BMC}}$}}
        \label{fig:wo_hp_1}
    \end{subfigure}
    \begin{subfigure}[b]{0.13\linewidth}
        \hspace{\linewidth}
        \caption{{\mysize w. $\mathrm{\mathbf{BMC}}$}}
         \label{fig:w_hp_1}
    \end{subfigure}
    \begin{subfigure}[b]{0.13\linewidth}
        \hspace{\linewidth}
        \caption{\mysize GT} %
        \label{fig:gt_1}
    \end{subfigure}
    \rulesepwhite
    \begin{subfigure}[b]{0.13\linewidth}
        \hspace{\linewidth}
        \caption{{\mysize w/o $\mathrm{\mathbf{BMC}}$}}
        \label{fig:wo_hp_2}
    \end{subfigure}
    \begin{subfigure}[b]{0.13\linewidth}
        \hspace{\linewidth}
        \caption{{\mysize w. $\mathrm{\mathbf{BMC}}$}}
        \label{fig:w_hp_2}
    \end{subfigure}
    \begin{subfigure}[b]{0.13\linewidth}
        \hspace{\linewidth}
        \caption{\mysize GT} %
        \label{fig:gt_2}
    \end{subfigure}
    
    \caption{Impact of the proposed biomechanical constraints (BMC). (b,e) Supplementing fully supervised data with 2D annotated data yields 3D poses with correct 2D projections, yet they are anatomically implausible. (c,f) Adding our biomechanical constraints significantly improves the pose prediction quantitatively and qualitatively. The resulting 3D poses are anatomically valid and display more accurate depth/scale even under severe self- and object occlusions, thus are closer to the ground-truth (d,g).} 
    \label{fig:teaser}
\end{figure*}

\section{Introduction}

Vision-based reconstruction of the 3D pose of human hands is a difficult problem that has applications in many domains.
Given that RGB sensors are ubiquitous, recent work has focused on estimating the full 3D pose \cite{spurr2018cvpr,mueller2018ganerated,iqbal2018hand,cai2018weakly,yang2019disentangling} and dense surface \cite{hasson2019learning,boukhayma20193d,ge20193d} of human hands from 2D imagery alone. This task is challenging due to the dexterity of the human hand, self-occlusions, varying lighting conditions and interactions with objects. 
Moreover, any given 2D point in the image plane can correspond to multiple 3D points in world space, all of which project onto that same 2D point. This makes 3D hand pose estimation from monocular imagery an ill-posed inverse problem in which depth and the resulting scale ambiguity pose a significant difficulty. 

Most of the recent methods use deep neural networks for hand pose estimation and rely on a combination of fully labeled real and synthetic training data (e.g., \cite{zimmermann2017,spurr2018cvpr,mueller2018ganerated,hasson2019learning,zhang2019end,baek2019pushing,iqbal2018hand, cai2018weakly, hasson2019learning}). However, acquiring full 3D annotations for real images is very difficult as it requires complex multi-view setups and labour intensive manual annotations of 2D keypoints in all views~\cite{hampali2019ho,zimmermann2019iccv,zhang2016}. On the other hand, synthetic data does not generalize well to realistic scenarios due to domain discrepancies. Some works attempt to alleviate this by leveraging additional 2D annotated images \cite{iqbal2018hand, boukhayma20193d}. Such kind of \textit{weakly-supervised} data is far easier to acquire for real images as compared to full 3D annotations. These methods use these annotations in a straightforward way in the form of a reprojection loss~\cite{boukhayma20193d} or supervision for the 2D component only~\cite{iqbal2018hand}. However, we find that the improvements stemming from including the weakly-supervised data in such a manner are mainly a result of 3D poses that agree with the 2D projection. Yet, the uncertainties arising due to depth ambiguities remain largely unaddressed and the resulting 3D poses can still be implausible. Therefore, these methods still rely on large amounts of fully annotated training data to reduce these ambiguities. In contrast, our goal is to \emph{minimize} the requirement of 3D annotated data as much as possible and \emph{maximize} the utility of weakly-labeled real data.

To this end, we propose a set of biomechanically inspired constraints (\textbf{BMC}) which can be integrated in the training of neural networks to enable anatomically plausible 3D hand poses even for data with 2D supervision only. Our key insight is that the human hand is subject to a set of limitations imposed by its biomechanics. We model these limitations in a differentiable manner as a set of \textit{soft constraints}. 
Note that this is a challenging problem. While the bone length constraints have been used successfully \cite{sun2017compositional,zhou2017towards}, capturing other biomechanical aspects is more difficult.
Instead of fitting a hand model to the predictions, we \emph{extract} the quantities in question directly from the predictions to impose our constraints. As such, the method of extraction has to be carefully designed to work under noisy and malformed 3D joint predictions while simultaneously being fully differentiable under any pose.
We propose to encode these constraints into a set of losses that are fully differentiable, interpretable and which can be incorporated into the training of any deep learning architecture that predicts 3D joint configurations. Due to this integration, we do not require a post-refinement step during test time. 
More specifically, our set of soft constraints consists of three equations that define 
\begin{inparaenum}[i)]
\item the range of valid bone lengths,
\item the range of valid palm structure, and
\item the range of valid joint angles of the thumb and fingers.
\end{inparaenum}
The main advantage of our set of constraints is that all parameters are interpretable and can either be set manually, opening up the possibility of personalization, or be obtained from a small set of data points for which 3D labels are available. 
As backbone model, we use the 2.5D representation proposed by Iqbal \etal~\cite{iqbal2018hand} due to its superior performance. We identify an issue in absolute depth calculation and remedy it via a novel refinement network. In summary, we contribute: 
\begin{itemize}
    \item A novel set of differentiable soft constraints inspired by the biomechanical structure of the human hand.
    \item Quantitative and qualitative evidence that demonstrates that our proposed set of constraints improves 3D prediction accuracy in \textit{weakly supervised settings}, resulting in an improvement of $55\%$ as opposed to $32\%$ as yielded by straightforward use of weakly-supervised data.
    \item A neural network architecture that extends \cite{iqbal2018hand} with a refinement step. 
    \item Achieving state-of-the-art performance on Dexter+Object using only synthetic and weakly-supervised real data, indicating cross-data generalizability. 
\end{itemize}
The proposed constraints require no special data nor are they specific to a particular backbone architecture.

\section{Related work}
Hand pose estimation from monocular RGB has gained traction in recent years due numerous possible applications. Generally there are two trains of thought. 

\textbf{Model-based methods} ensure plausible poses by fitting a hand model to the observation via optimization. As they are not learning-based, they are sensitive to initial conditions, rely on temporal information \cite{heap1996towards, oikonomidis2011efficient, oikonomidis2011full, panteleris2019using} or do not take the image into consideration during optimization \cite{panteleris2019using}. Whereas some make use of geometric primitives \cite{oikonomidis2011efficient, oikonomidis2011full, panteleris2019using}, other simply model the joint angles directly \cite{lee1995model, kuch1995vision, rhee2006human, cerveri2007finger, cordella2012bio, wu1999capturing}, learn a lower dimensional embedding of the joints \cite{lin2000modeling}, pose \cite{heap1996towards} or go a step further and model muscles of the hand \cite{albrecht2003construction}. Different to these methods, we propose to incorporate these constraints \emph{directly} into the training procedure of a neural network in a fully differentiable manner. As such, we do not fit a hand model to the prediction, but \emph{extract} and constrain the biomechanical quantities from them directly. The resulting network predicts biomechanically-plausible poses and does not suffer from the same disadvantages.

\textbf{Learning-based methods} utilize neural networks that either directly regress the 3D positions of the hand keypoints~\cite{zimmermann2017, spurr2018cvpr, yang2019disentangling, mueller2018ganerated, iqbal2018hand, tekin2019cvpr} or predict the parameters of a deformable hand model~\cite{baek2019pushing, boukhayma20193d, zhang2019end, hasson2019learning, xiang2019cvpr}. Zimmermann \etal~\cite{zimmermann2017} are the first to use deep neural network for root-relative 3D hand pose estimation from RGB images via a multi-staged approach. Spurr \etal \cite{spurr2018cvpr} learn a unified latent space that projects multiple modalities into the same space, learning a lower level embedding of the hands. Similarly, Yang \etal\cite{yang2019disentangling} learn a latent space that disentangles background, camera and hand pose. However, all these methods require large numbers of fully labeled training data. Cai \etal \cite{cai2018weakly} try to alleviate this problem by introducing an approach that utilizes paired RGB-D images to regularize the depth predictions.  
Mueller \etal\cite{mueller2018ganerated} attempt to improve the quality of synthetic training data by learning a GAN model that minimizes the discrepancies between real and synthetic images. Iqbal \etal~\cite{iqbal2018hand} decompose the task into learning 2D and root-relative depth components. This decomposition allows to use weakly-labeled real images with only 2D pose annotations which are cheap to acquire. While these methods demonstrate better generalization by adding a large number weakly-labeled training samples, the main drawback of this approach is that the depth ambiguities remain unaddressed. As such, training using only 2D pose annotations does not impact the depth predictions. This may result in 3D poses with accurate 2D projections, but due to depth ambiguities the 3D poses can still be implausible. In contrast, in this work, we propose a set of biomechanical constraints that ensures that the predicted 3D poses are always anatomically plausible during training (see~Fig.~\ref{fig:teaser}). We formulate these constraints in form of a fully-differentiable loss functions which can be incorporated into any deep learning architecture that predicts 3D joint configurations. 
We use a variant of Iqbal \etal~\cite{iqbal2018hand} as a baseline and demonstrate that the requirement of fully labeled real images can be significantly minimized while still maintaining performance on par with fully-supervised methods. 

Other recent methods directly predict the parameters of a deformable hand model, \eg, MANO~\cite{romero2017}, from RGB images~\cite{boukhayma20193d, zhang2019end, hasson2019learning, xiang2019cvpr, pavlakos2019expressive}. The predicted parameters consist of the shape and pose deformations wrt. a mean shape and pose that are learned using large amounts of 3D scans of the hand. Alternatively, \cite{ge20193d, kulon2019single} circumvent the need for a parametric hand model by directly predicting the mesh vertices from RGB images. These methods require both shape and pose annotations for training, therefore obtaining such kind of training data is even harder. Hence, most methods rely on synthetic training data. Some methods~\cite{boukhayma20193d, zhang2019end, baek2019pushing} alleviate this by introducing re-projection losses that measure the discrepancy between the projection of 3D mesh with labeled 2D poses~\cite{boukhayma20193d} or silhouettes~\cite{zhang2019end, baek2019pushing}. Even though they utilize strong hand priors in form of a mean hand shape and by operating on a low-dimensional PCA space, using re-projection losses with weakly-labeled data still does not guarantee that the resulting 3D poses will be anatomically plausible. Therefore, all these methods rely on a large number of fully labeled training data. In body pose estimation, such methods generally resort to adversarial losses to ensure plausibility~\cite{hmrKanazawa18}.

\textbf{Biomechanical constraints} have also been used in the literature to encourage plausible 3D poses by imposing biomechanical limits on the structure of the hands~\cite{cobos2008efficient, chen2013constraint, ryf1995neutral, wan2019self, dibra2017refine, tompson2014real, melax2013dynamics, sridhar2013interactive, xu2013efficient} or via a learned refinement model\cite{cai2019exploiting}. Most methods \cite{cobos2008efficient, chen2013constraint, ryf1995neutral, tompson2014real, melax2013dynamics, sridhar2013interactive, xu2013efficient, aristidou2018hand} impose these limits via inverse kinematic in a post-processing step, therefore the possibility of integrating them for neural network training remains unanswered. Our proposed soft-constraints are fully integrated into the network, which does not require a post-refinement step during test time. Similar to our method, \cite{wan2019self, dibra2017refine} also penalize invalid bone lengths. However, we additionally model the joint limits and palmar structure.

\section{Method}

\begin{figure*}[t]
    \centering
    \includegraphics[width=\textwidth]{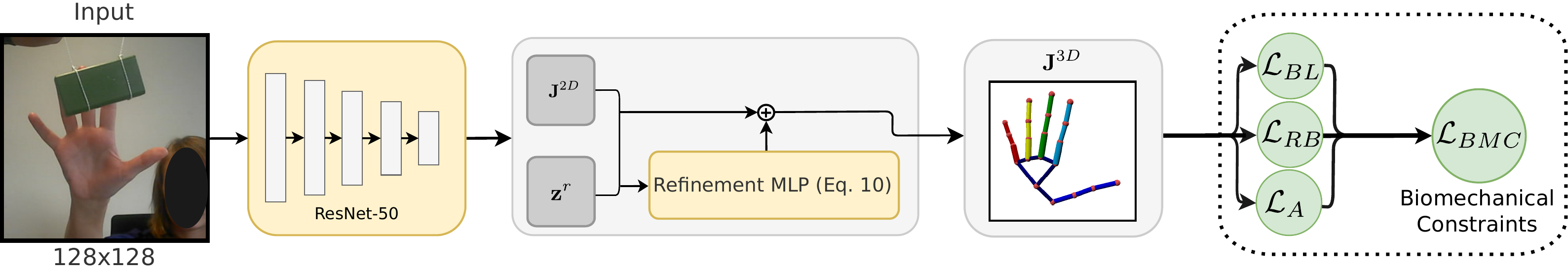}
    \caption{Method overview. A model takes an RGB image and predicts the 3D joints on which we apply our proposed BMC. These guide the model to predict plausible poses.
    }
    \label{fig:network_architecture}
\end{figure*}

\label{sec:method}

Our method is summarized in Figure \ref{fig:network_architecture}. Our key contribution is a set of novel constraints that constitute a biomechanical model of the human hand and capture the bone lengths, joint angles and shape of the palm. We emphasize that we do not fit a kinematic model to the predictions, but instead extract the quantities in question directly from the predictions in order to constrain them. Therefore the method of extraction is carefully designed to work under noisy and malformed 3D joint predictions while simultaneously being fully differentiable in any configuration. These biomechanical constraints provide an inductive bias to the neural network. Specifically, the network is guided to predict anatomically plausible hand poses for weakly-supervised data (\ie~2D only), which in turn increases generalizability. The model can be combined with any backbone architecture that predicts 3D keypoints. 
We first introduce the notations used in this paper followed by the details of the proposed biomechanical losses. Finally, we discuss the integration with a variant of \cite{iqbal2018hand}.

\begin{figure*}[t]
  \centering
  \setlength{\tabcolsep}{5pt}
\scalebox{0.84}{
  \begin{tabular}{c c c c}
    \includegraphics[trim={2cm 0 0 0},clip, height=0.25\textwidth]{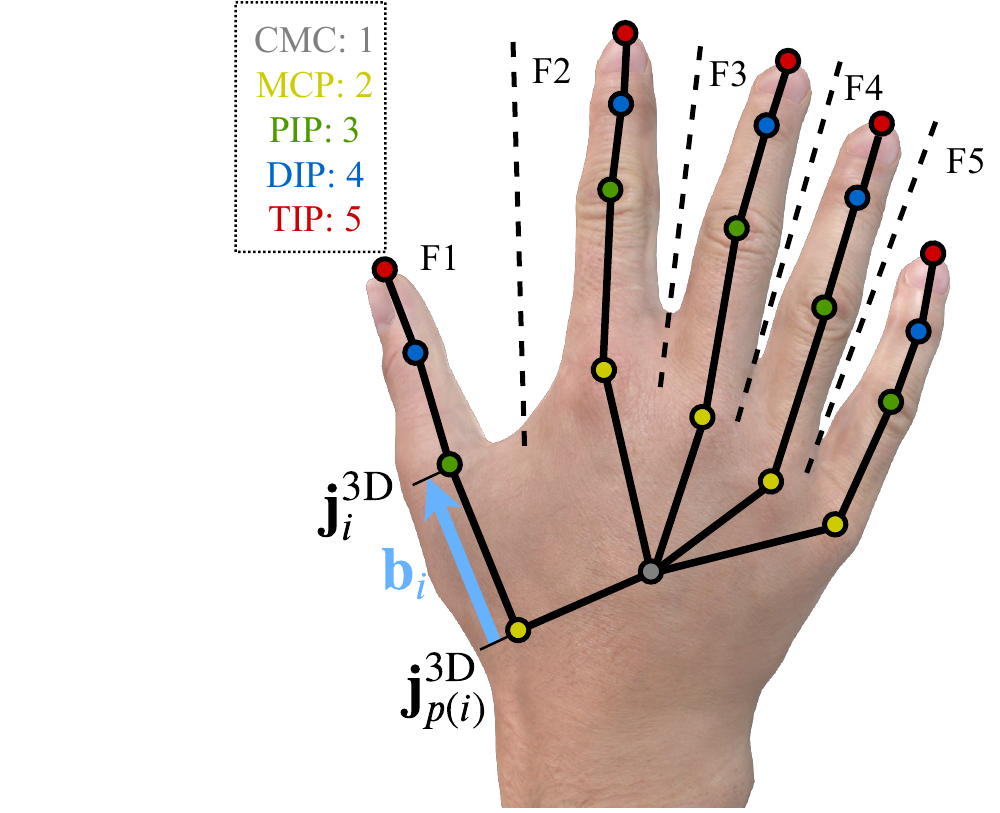} & 
    \includegraphics[height=0.2\textwidth]{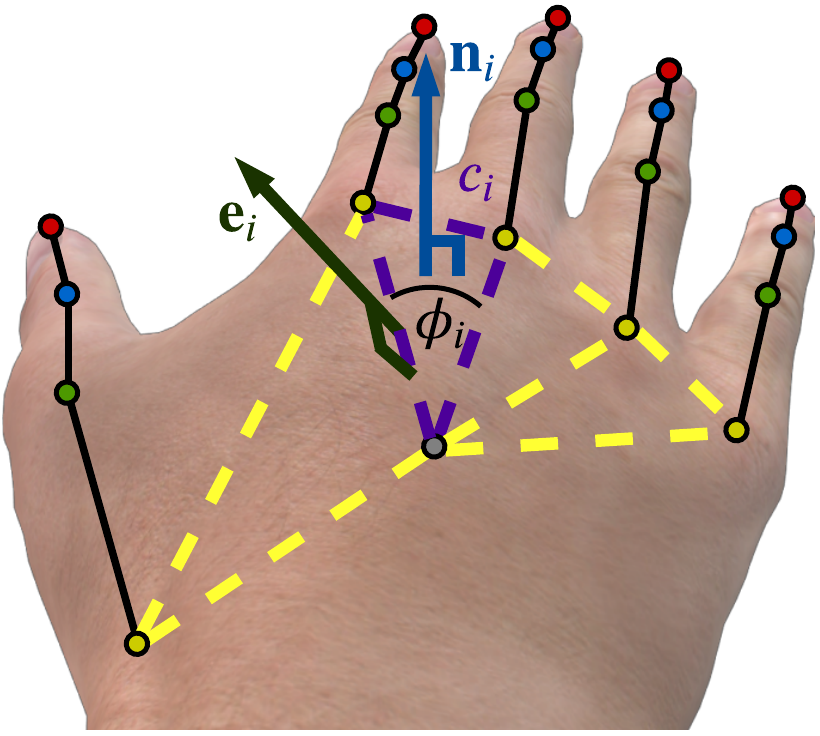} &  
    \includegraphics[height=0.25\textwidth]{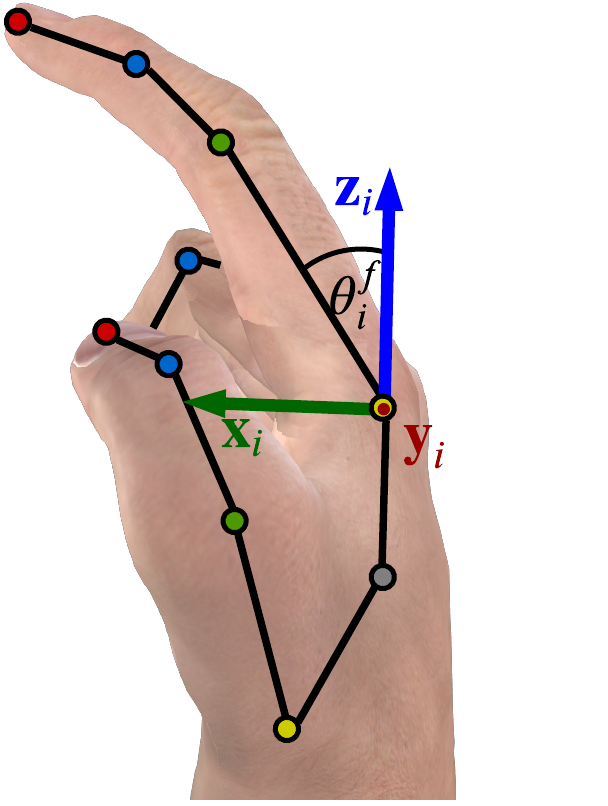} &         \includegraphics[height=0.25\textwidth]{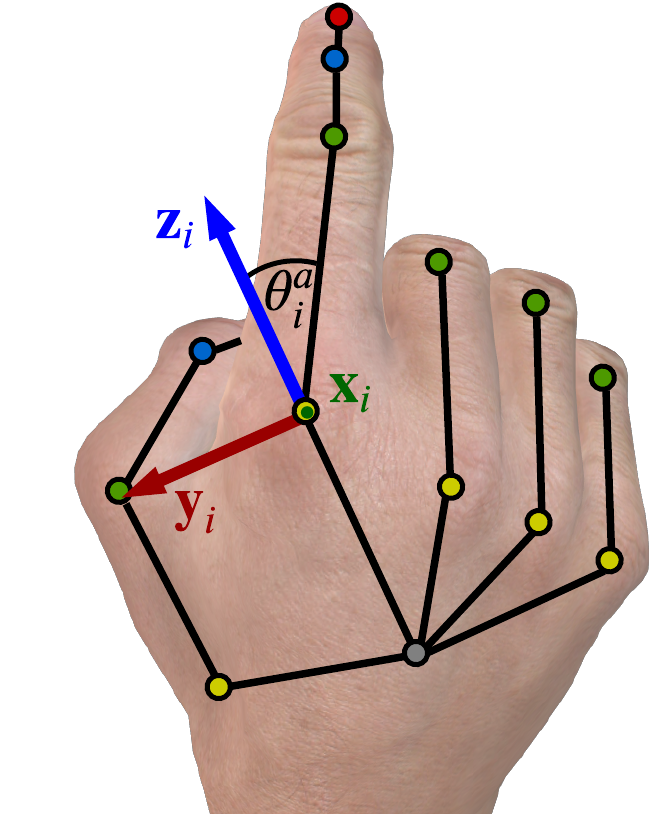} \\
    a) Joint skeleton structure & b) Root bone structure & \multicolumn{2}{c}{c) Angles. Flexion: Left -- Abduction: Right} \\
  \end{tabular}
}
\caption{Illustration of our proposed biomechanical structure.}
\label{fig:handmodel}
\end{figure*}

\textbf{Notation. }
We use bold capital font for matrices, bold lowercase for vector and roman font for scalars. We assume a right hand. The joints $[\V{j}^{3D}_1, \dots, \V{j}^{3D}_{21}]=\V{J}^{3D} \in \R^{21 \times 3}$ define a kinematic chain of the hand starting from the root joint $\V{j}^{3D}_1$ and ending in the fingertips. For the sake of simplicity, the joints of the hands are grouped by the fingers, denoted as the respective set $F1,\dots,F5$, visualized in \figref{fig:handmodel}a. Each $\V{j}^{3D}_i$, except the root joint (CMC), has a parent, denoted as $p(i)$. We define a bone $\V{b}_i = \V{j}^{3D}_{i+1} - \V{j}^{3D}_{p(i+1)}$ as the vector pointing from the parent joint to its child joint. Hence $[\V{b}_1,\dots,\V{b}_{20}] = \V{B} \in \R^{20 \times 3}$. The bones are named according to the child joint. For example, the bone connecting MCP to PIP is called PIP bone. We define the five root bones as the MCP bones, where one endpoint is the root $\V{j}^{3D}_1$. Intuitively, the root bones are those that lie within and define the palm. We define the bones $\V{b}_i$ with $i=1,\dots,5$ to correspond to the root bones of fingers $F1, \dots, F5$. We denote the angle $\alpha(v_1, v_2) = \mathrm{arccos}\big(\frac{\V{v}_1^T \V{v}_2}{||\V{v}_1||_2 \, ||\V{v}_2||_2}\big)$ between the vectors $\V{v}_1, \V{v}_2$. The interval loss is defined as $\mathcal{I}(x;a,b) = \max(a-x,0) + \max(x-b,0)$. The normalized vector is defined as $\mathrm{norm}(\V{x}) = \frac{\V{x}}{||\V{x}||_2}$. Lastly, $\mathrm{P}_{\V{xy}}(\V{v})$ is the orthogonal projection operator, projecting $\V{v}$ orthogonally onto the $\V{x}$-$\V{y}$ plane where $\V{x}$,$\V{y}$ are vectors.

\subsection{Biomechanical constraints}
Our goal is to integrate our biomechanical soft constraints (BMC) into the training procedure that encourages the network to predict feasible hand poses. We seek to avoid iterative optimization approaches such as inverse kinematics in order to avert significant increases in training time.

The proposed model consists of three functional parts, visualized in \figref{fig:handmodel}. First, we consider the length of the bones, including the root bones of the palm. Second, we model the structure and shape of the palmar region, consisting of a rigid structure made up of individual joints. %
To account for inter-subject variability of bones and palm structure, it is important to not enforce a specific mean shape. Instead, we allow for these properties to lie within a valid range. Lastly, the model describes the articulation of the individual fingers.
The finger motion is described via modeling of the flexion and abduction of individual bones. As their limits are interdependent, they need to be modeled jointly. As such, we propose a novel constraint that takes this interdependence into account.

The limits for each constraint can be attained manually from measurements, from the literature (e.g \cite{chen2013constraint, ryf1995neutral}), or acquired in a data-driven way from 3D annotations, should they be available.

\textbf{Bone length. }
For each bone $i$, we define an interval $[b^{\min}_i, b^{\max}_i]$ of valid bone length and penalize if the length $||\V{b}_i||_2$ lies outside of this interval:
\begin{equation}
\nonumber
    \mathcal{L}_{\mathrm{BL}}(\V{J}^{3D}) = \frac{1}{20} \sum_{i=1}^{20} \mathcal{I}(||\V{b}_i||_2; b^{\min}_i, b^{\max}_i) 
\end{equation}
This loss encourages keypoint predictions that yield valid bone lengths. \figref{fig:handmodel}a shows the length of a bone in blue.

\textbf{Root bones.} To attain valid palmar structures we first interpret the root bones as spanning a mesh and compute its curvature by following \cite{reed15stackexchange}:
\begin{equation}
    c_i = \frac{(\V{e}_{i+1} - \V{e}_i)^T(\V{b}_{i+1} - \V{b}_i)}{||\V{b}_{i+1} - \V{b}_i||^2}, \text{ for } i \in \{1,2,3,4\}
\end{equation}
Where $\V{e}_i$ is the edge normal at bone $\V{b}_i$:
\begin{equation}
\begin{split}
\label{eq:edge_normal}
    \V{n}_i &= \mathrm{norm}(\V{b}_{i+1} \times \V{b}_i), \text{for } i \in \{1,2,3,4\} \\
        \V{e}_i &= 
        \begin{dcases}
        \V{n}_1, & \text{if } i = 1 \\
        \mathrm{norm}(\V{n}_{i} + \V{n}_{i-1}), &\text{if } i \in \{2,3,4\}\\
        \V{n}_4, & \text{if } i = 5
        \end{dcases}
\end{split}
\end{equation}
Positive values of $c_i$ denote an arched hand, for example when pinky and thumb touch. A flat hand has no curvature. \figref{fig:handmodel}b visualizes the mesh in dashed yellow and the triangle over which the curvature is computed in dashed purple.

We ensure that the root bones fall within correct angular ranges by defining the angular distance between neighbouring $\V{b}_i$,$\V{b}_{i+1}$ across the plane they span:
\begin{equation}
    \phi_i = \alpha(\V{b}_i, \V{b}_{i+1})
\end{equation}
We constrain both the curvature $c_i$ and angular distance $\phi_i$ to lie within a valid range $[c_i^{\min}, c_i^{\max}]$ and $[\phi_i^{\min}, \phi_i^{\max}]$:
\begin{equation}
\nonumber
    \mathcal{L}_{\mathrm{RB}}(\V{J}^{3D}) = \frac{1}{4}\sum_{i=1}^4 \big(\mathcal{I}(c_i;c_i^{\min},c_i^{\max}) + \mathcal{I}(\phi_i;\phi_i^{\min}, \phi_i^{\max})\big)
\end{equation}
$\mathcal{L}_{\mathrm{RB}}$ ensures that the predicted joints of the palm define a valid structure, which is crucial since the kinematic chains of the fingers originate from this region. 

\textbf{Joint angles. }
To compute the joint angles, we first need to define a consistent frame $\V{F}_i$ of a local coordinate system for each finger bone $\V{b}_i$. $\V{F}_i$ must be consistent with respect to the movements of the finger. In other words, if one constructs $\V{F}_i$ given a pose $\V{J}^{3D}_1$, then moves the fingers and corresponding $\V{F}_i$ into pose $\V{J}^{3D}_2$, the resulting $\V{F}_i$ should be the same as if constructed from $\V{J}^{3D}_2$ directly. 

We assume right-handed coordinate systems. To construct $\V{F}_i$, we define two out of three axes based on the palm. We start with the first layer of fingers bones (PIP bones). We define their respective $z$-component of $\V{F}_i$ as the normalized bone of their respective parent bone (in this case, the root bones): $\V{z}_i = \text{norm}(\V{b}_{p(i)})$. Next, we define the $x$-axis, based on the plane normals spanned by two neighbouring root bones:
\begin{equation}
\begin{split}
        \V{x}_i &= 
        \begin{dcases}
        -\V{n}_{p(i)}, &\text{if } p(i) \in \{1,2\}\\
        -\mathrm{norm}(\V{n}_{p(i)} + \V{n}_{p(i)-1}), &\text{if } p(i) \in \{3,4\}\\
        -\V{n}_4, &\text{if } p(i) = 5
        \end{dcases}\\
\end{split}
\end{equation}
Where $\V{n}_i$ is defined as in Eq. \ref{eq:edge_normal}. Lastly, we compute the last axis $\V{y}_i = \text{norm}(\V{z}_i \times \V{x}_i)$. Given $\V{F}_i$, we can now define the flexion and abduction angles. Each of these angles are given with respect to the local $z$-axis of $\V{F}_i$. Given $\V{b}_i$ in its local coordinates $\V{b}_i^{\V{F}_i}$ wrt. $\V{F}_i$, we define the flexion and abduction angles as:
\begin{equation}
    \begin{split}
        \theta^{\mathrm{f}}_i &= \alpha(\mathrm{P}_{xz}(\V{b}_i^{\mathbf{F}_i}), \V{z}_i)\\
        \theta^{\mathrm{a}}_i &= \alpha(\mathrm{P}_{xz}(\V{b}_i^{\mathbf{F}_i}), \V{b}_i^{\mathbf{F}_i})
    \end{split}
    \label{eq:angle_comp}
\end{equation}
\figref{fig:handmodel}c visualizes $\V{F}_i$ and the resulting angles. Note that this formulation leads to ambiguities, where different bone orientations can map to the same ($\theta^{\mathrm{f}}_i$, $\theta^{\mathrm{a}}_i$)-point. We resolve this via an octant lookup, which leads to angles in the intervals $\theta^{\mathrm{f}}_i \in [-\pi,\pi]$ and $\theta^{\mathrm{a}}_i \in [-\pi/2,\pi/2]$ respectively. See appendix for more details.

Given the angles of the first set of finger bones, we can then construct the remaining two rows of finger bones. Let $\V{R}^{\theta_i}$ denote the rotation matrix that rotates by $\theta^{\mathrm{f}}_i$ and $\theta^{\mathrm{a}}_i$ such that $\V{R}^{\theta_i} \V{z}_i = \V{b}_i^{\V{F}_i}$, then we iteratively construct the remaining frames along the kinematic chain of the fingers:
\begin{equation}
    \begin{split}
        \mathbf{F}_i = \V{R}^{\theta_i}\V{F}_{p(i)}
    \end{split}
\end{equation}
This method of frame construction via rotating by $\theta^{\mathrm{f}}_i$ and $\theta^{\mathrm{a}}_i$ ensures consistency across poses. The remaining angles can be acquired as described in Eq. \ref{eq:angle_comp}.

Lastly, the angles need to be constrained. One way to do this is to consider each angle independently and penalize them if they lie outside an interval. This corresponds to constraining them within a box in a 2D space, where the endpoints are the min/max of the limits. However, finger angles have inter-dependency, therefore 
we propose an alternative approach to account for this. Given points $\theta_i = (\theta^{\mathrm{f}}_i, \theta^{\mathrm{a}}_i)$ that define a range of motion, we approximate their convex hull on the $(\theta^{\mathrm{f}}, \theta^{\mathrm{a}})$-plane with a fixed set of points $\mathcal{H}_i$. The angles are constrained to lie within this structure by minimizing their distance to it:
\begin{equation}
    \begin{split}
        \mathcal{L}_\mathrm{A}(\V{J}^{3D}) = \frac{1}{15} \sum_{i=1}^{15} D_H(\theta_i, \mathcal{H}_i)
    \end{split}
\end{equation}
Where $D_H$ is the distance of point $\theta_i$ to the hull $\mathcal{H}_i$. Details on the convex hull approximation and implementation
can be found in the appendix.

\subsection{$\mathbf{Z}^{\mathrm{root}}$ Refinement}
\label{sec:refinement}
The 2.5D joint representation allows us to recover the value of the absolute pose $Z^{root}$ up to a scaling factor . This is done by solving a quadratic equation dependent on the 2D projection $\V{J}^{2D}$ and relative depth values $\V{z}^r$, as proposed in \cite{iqbal2018hand}. In practice, small errors in $\V{J}^{2D}$ or $\V{z}^r$ can result in large deviations of $Z^{root}$. This leads to big fluctuations in the translation and scale of the predicted pose, which is undesirable. To alleviate these issues, we employ an MLP to refine and smooth the calculated $\hat{Z}^{root}$:
\begin{equation}
\label{eq:zroot_ref}
    \hat{Z}^{root}_\mathrm{ref} = \hat{Z}^{root} + M_{\mathrm{MLP}}(\V{z}^r, \V{K}^{-1}\V{J}^{2D}, \hat{Z}^{root}; \V{\omega})
\end{equation}
Where $M_\mathrm{MLP}$ is a multilayered perceptron with parameters $\V{\omega}$ that takes the predicted and calculated values $\V{z}^r \in \R^{21}$, $\V{K}^{-1}\V{J}^{2D} \in \R^{21\times 3}$, $Z^{root} \in \R$ and outputs a residual term. Alternatively, one could predict $Z^{root}$ directly using an MLP with the same input. However, as the exact relationship between the predicted variables and $Z^{root}$ is known, we resort to the refinement approach instead of requiring a model to learn what is already known.

\subsection{Final loss}
The biomechanical soft constraints is constructed as follows:
\begin{equation}
      \mathcal{L}_{\mathrm{BMC}} = \lambda_{\mathrm{BL}}\mathcal{L}_{\mathrm{BL}} + \lambda_{\mathrm{RB}}\mathcal{L}_{\mathrm{RB}} + \lambda_{\mathrm{A}}\mathcal{L}_{\mathrm{A}}
\end{equation}
Our final model is trained on the following loss function:
\begin{equation}
    \mathcal{L} = \lambda_{\V{J}^{2D}}\mathcal{L}_{\V{J}^{2D}} + \lambda_{\mathbf{z}^r}\mathcal{L}_{\mathbf{z}^r} + \lambda_{\mathrm{Z_\mathrm{ref}^{root}}}\mathcal{L}_{\mathrm{Z^{root}}} + \mathcal{L}_{\mathrm{BMC}}
\end{equation}
Where $\mathcal{L}_{\V{J}^{2D}}$, $\mathcal{L}_{\mathbf{z}^r}$ and $\mathcal{L}_{\mathrm{Z^{root}}}$ are the L$1$ loss on any available $\V{J}^{2D}$, $\mathbf{z}^r$ and $Z^{root}$ labels respectively. The weights $\lambda_i$ balance the individual loss terms.

\section{Implementation}

We use a ResNet-50 backbone~\cite{he2016}. The input to our model is a $128 \times 128$ RGB image from which the 2.5D representation is directly regressed. The model and its refinement step is trained on fully supervised and weakly-supervised data. The network was trained for 70 epochs using SGD with a learning rate of $5e\minus3$ and a step-wise learning rate decay of $0.1$ after every 30 epochs. We apply the biomechanical constraints directly on the predicted 3D keypoints $\V{J}^{3D}$.

\section{Evaluation}

\label{sec:evaluation}
Here we introduce the datasets used, show the performance of our proposed $\mathcal{L}_{\mathrm{\mathbf{BMC}}}$ and compare in extensive settings. Specifically, we study the effect of adding weakly supervised data to complement fully supervised training. All experiments are conducted in a setting where we assume access to a fully supervised dataset, as well as a supplementary weakly supervised real dataset. Therefore we have access to 2D ground-truth annotations and the computed constraint limits. We study two cases of 3D supervision sources:

\noindent{\bf Synthetic data.} We choose RHD. Acquiring fully labeled synthetic data is substantially easier as compared to real data. Section \ref{sec:weaksup_effect}-\ref{sec:weaksup_synth} consider this setting.%

\noindent{\bf Partially labeled real data.} In Section \ref{sec:weaksup_real} we gradually increase the number of real 3D labeled samples to study how the proposed approach works under different ratio of fully to weakly supervised data.

To make clear what kind of supervision is used we denote $\mathbf{3D}_\mathrm{A}$ if 3D annotation is used from dataset $\mathrm{A}$. We indicate usage of 2D from dataset $\mathrm{A}$ as $\mathbf{2D}_A$. Section \ref{sec:weaksup_effect} and \ref{sec:ablation} are evaluated on FH. 
\\
\subsection{Datasets}
\vspace{-1mm}

\begin{wraptable}{r}{0.58\textwidth}
\begin{center}
\vspace{-4mm}
\caption{Overview of datasets used for evaluation.}
\scriptsize
    \begin{tabular}{lcccl}
        \toprule
        \multirow{2}{*}{Name} &  \multirow{2}{*}{Type} & joints & train/test \\
             &  & \# & \#  \\
        \midrule
        Rendered Hand Pose (RHD) \cite{zimmermann2017} & Synth & 21 & 42k / 2.7k   \\
        FreiHAND (FH) \cite{zimmermann2019iccv} & Real & 21 & 33k / 4.0k         \\
        Dexter+Object (D+O) \cite{sridhar2016eccv}  & Real & 5 & ~~-~ / 3.1k           \\
        Hand-Object 3D (HO-3D) \cite{hampali2019ho}  & Real & 21 & 11k / 6.6k     \\
        \bottomrule
    \end{tabular}
    \label{tab:datasets}
\end{center}
\end{wraptable}

Each dataset that provides 3D labels comes with the camera intrinsics. Hence the 2D pose can be easily acquired from the 3D pose. \tabref{tab:datasets} provides an overview of datasets used. The test set of HO-3D and FH are available only via a submission system with limited number of total submissions. Therefore for the ablation study (Section \ref{sec:ablation}) and inspecting the effect of weak-supervision (Section \ref{sec:weaksup_effect}), we divide the training set into a training and validation split. For these sections, we choose to evaluate on FH due to its large number of samples and variability in both hand pose and shape.

\subsection{Evaluation Metric}
\noindent{\bf HO-3D.} The error given by the submission system is the mean joint error in mm. The INTERP is the error on test frames sampled from training sequences that are not present in the training set. The EXTRAP is the error on test samples that have neither hand shapes nor objects present in the training set. We used the version of the dataset that was available at the time \cite{armagan2020measuring}.

\noindent{\bf FH.} The error given by the submission system is the mean joint error in mm. Additionally, the area under the curve (AUC) of the percentage of correct keypoints (PCK) plot is reported. The PCK values lie in an interval from 0 mm to 50 mm with 100 equally spaced thresholds. Both the aligned (using procrustes analysis) and unaligned scores are given. We report the aligned score. The unaligned score can be found in the appendix.

\noindent{\bf D+O.} We report the AUC for the PCK thresholds of 20 to 50 mm comparable with prior work \cite{zimmermann2019iccv, zhang2019end, boukhayma20193d}. For \cite{iqbal2018hand, spurr2018cvpr, mueller2018ganerated, zimmermann2017} we report the numbers as presented in \cite{zhang2019end} as they consolidate all AUC of related work in a consistent manner using the same PCK thresholds. For \cite{baek2019pushing}, we recomputed the AUC for the same interval based on the values provided by the authors.

\subsection{Effect of Weak-Supervision}
\label{sec:weaksup_effect}

\begin{table}[t]
\centering
\caption{The effect of weak-supervision on the \textbf{validation} split of FH. Training on synthetic data (RHD) leads to poor accuracy on real data (FH). Adding real 2D labeled data reduces 3D prediction error due to better alignment with the 2D projection. Adding our proposed $\mathcal{L}_\mathrm{\mathbf{BMC}}$ significantly reduces the 3D error due to more accurate $\V{Z}$.}
\label{tbl:weaksup_effect}
\scriptsize
\begin{tabularx}{\columnwidth}{XXccc}
\toprule
 \multirow{1}{4cm}{Effect of weak-supervision}  & \multirow{2}{16cm}{Description} & \multicolumn{3}{c}{Mean Error $\downarrow$}  \\ 
 & &\multirow{1}{1.3cm}{2D (px)} & \multirow{1}{1.3cm}{Z (mm)} & \multirow{1}{1.3cm}{3D (mm)} \\
\midrule
$\mathbf{3D_\mathrm{RHD} + 3D_\mathrm{FH}}$                         & Fully supervised, synthetic+real                    & 3.72  & 5.69  & 8.78  \\ 
+ $\mathbf{\mathcal{L}_\mathrm{\mathbf{BMC}}}$ (\textbf{ours})& ~~~+ BMC  & \textbf{3.70}  & \textbf{5.44}  & \textbf{8.60}  \\ \hdashline
$\mathbf{3D_\mathrm{RHD}}$                                          & Fully supervised, synthetic only                    & 12.35 & 20.02 & 30.82 \\
+ $\mathbf{2D_\mathrm{FH}}$                         & + Weakly supervised, real                            & 3.80  & 17.02 & 20.92 \\ 
~~~+ $\mathcal{L}_\mathrm{\mathbf{BMC}}$ (\textbf{ours})  & ~~~+ BMC                    & \textbf{3.79}  & \textbf{9.97}  & \textbf{13.78} \\ %

\bottomrule
\end{tabularx}
\label{tbl:weaksup_effect}
\end{table}

We first inspect how weak-supervision affects the performance of the model. We decompose the 3D prediction error on the validation set of FH in terms of its 2D ($\V{J}^{2D}$) and depth component ($\V{Z}$) via the pinhole camera model $\V{Z}^{-1}\V{K}\V{J}^{3D} = \V{J}^{2D}$ and evaluate their individual error. 

We train four models using different data sources.
\begin{inparaenum}[1)]
\item Full 3D supervision on both synthetic RHD and real FH ($\mathbf{3D}_\mathrm{RHD}+\mathbf{3D}_\mathrm{FH}$), which serves as an upper bound for when all 3D labels are available
\item Fully supervised on RHD which constitutes our lower bound on accuracy ($\mathbf{3D}_\mathrm{RHD}$)
\item Fully supervised on RHD with naive application of weakly-supervised FH ($+\mathbf{2D}_\mathrm{FH}$)
\item Like setting 3) but adding our proposed constraints ($\mathbf{+ \mathcal{L}_\mathrm{\mathbf{BMC}}}$).
\end{inparaenum}
 
\tabref{tbl:weaksup_effect} shows the results. The model trained with full 3D supervision from real and synthetic data reflects the best setting. Adding $\mathcal{L}_\mathrm{\mathbf{BMC}}$ during training slightly reduces 3D error ($8.78$mm to $8.6$mm) primarily due to a regularization effect. When the model is trained only on synthetic data ($\mathbf{3D}_\mathrm{RHD}$) we observe a significant rise ($8.78$mm to $30.82$mm) in 3D error due to the poor generalization from synthetic data. When weak-supervision is provided from the real data ($+\mathbf{2D}_\mathrm{FH}$), the error is reduced ($30.82$mm to $20.92$mm). However, inspecting this more closely we observe that the improvement comes mainly from 2D error reduction ($12.35$px to $3.8$px), whereas the depth component is improved marginally ($20.02$mm to $17.02$mm). Observing these samples qualitatively (\figref{fig:teaser}), we see that many do not adhere to biomechanical limits of the human hand. By penalizing such violations via our proposed losses $\mathcal{L}_\mathrm{\mathbf{BMC}}$ to the weakly supervised setting we see a significant improvement in 3D error ($20.92$mm to $13.78$mm) which is due to improved depth accuracy ($20.02$mm to $9.97$mm). Inspecting  (\eg~\figref{fig:teaser}) closer, we see that the model predicts the correct 3D pose in challenging settings such as heavy self- and object occlusion, despite having never seen such samples in 3D. Since $\mathcal{L}_\mathrm{\mathbf{BMC}}$ describes a valid range, rather than a specific pose, slight deviations from the ground truth 3D pose have to be expected which explains the small remaining quantitative gap from the fully supervised model.

\subsection{Ablation Study}
\label{sec:ablation}
We quantify the individual contributions of our proposals on the validation set of FH and reproduce these results on HO-3D in supplementary. Each error metric is computed for the root-relative 3D pose.

\begin{figure*}[t]
    \centering

    \begin{subfigure}[b]{0.16\textwidth}
        \includegraphics[width=\textwidth]{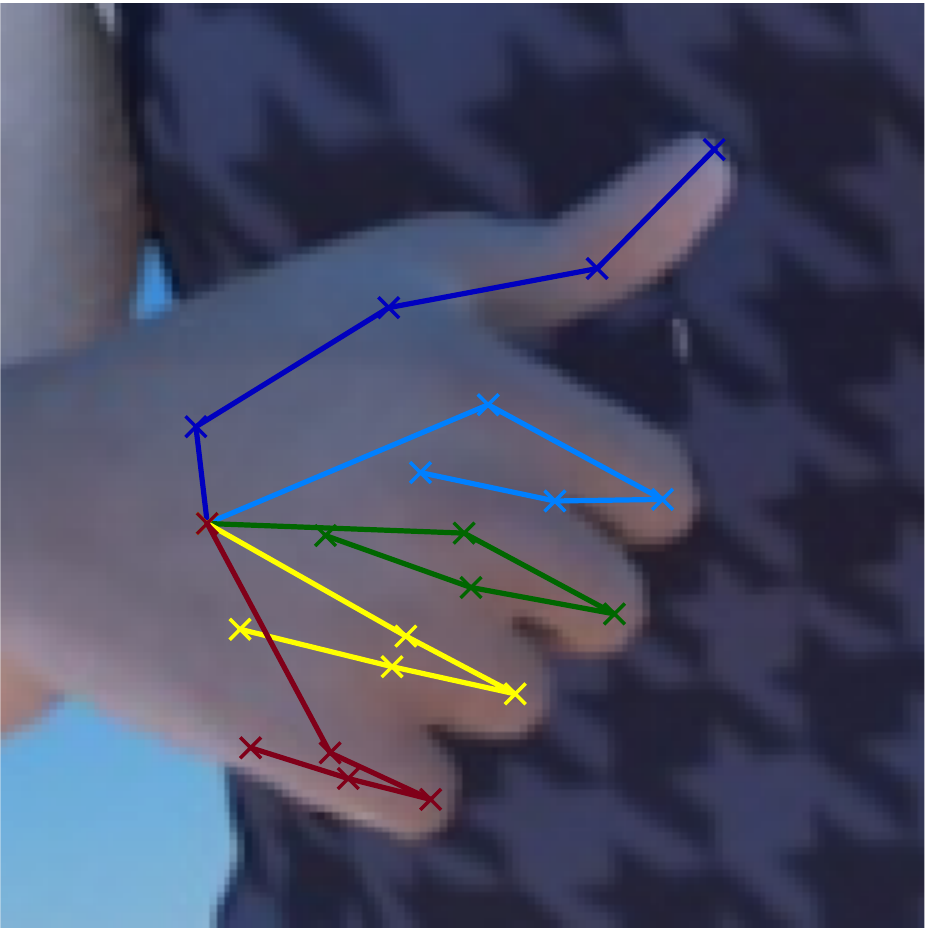}
        \caption{Input image}
        \label{fig:ablation_rgb}
    \end{subfigure}%
    \begin{subfigure}[b]{0.2\textwidth}
        \vspace{-1cm}
        \includegraphics[width=\textwidth, clip, trim=14cm 8.5cm 10.5cm 1cm]{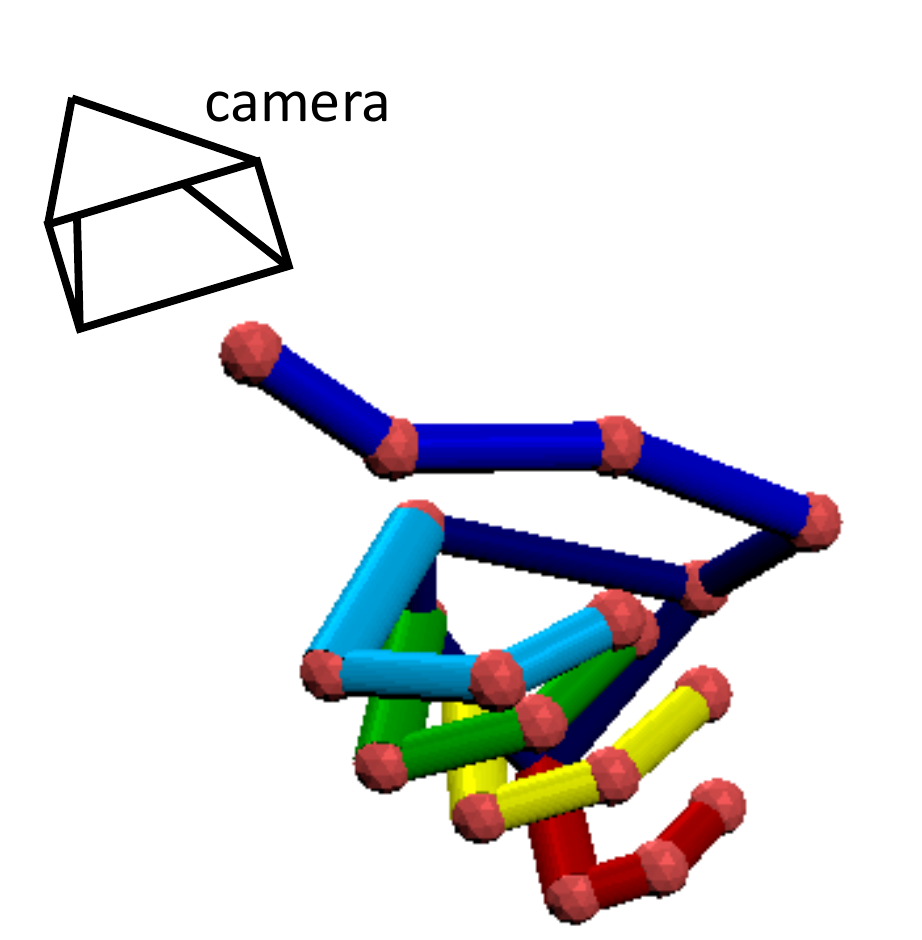}
        \caption{Ground-truth}
        \label{fig:ablation_gt}
    \end{subfigure}%
    \begin{subfigure}[b]{0.2\textwidth}
        \includegraphics[width=\textwidth, clip, trim=17cm 9.5cm 15cm 3cm]{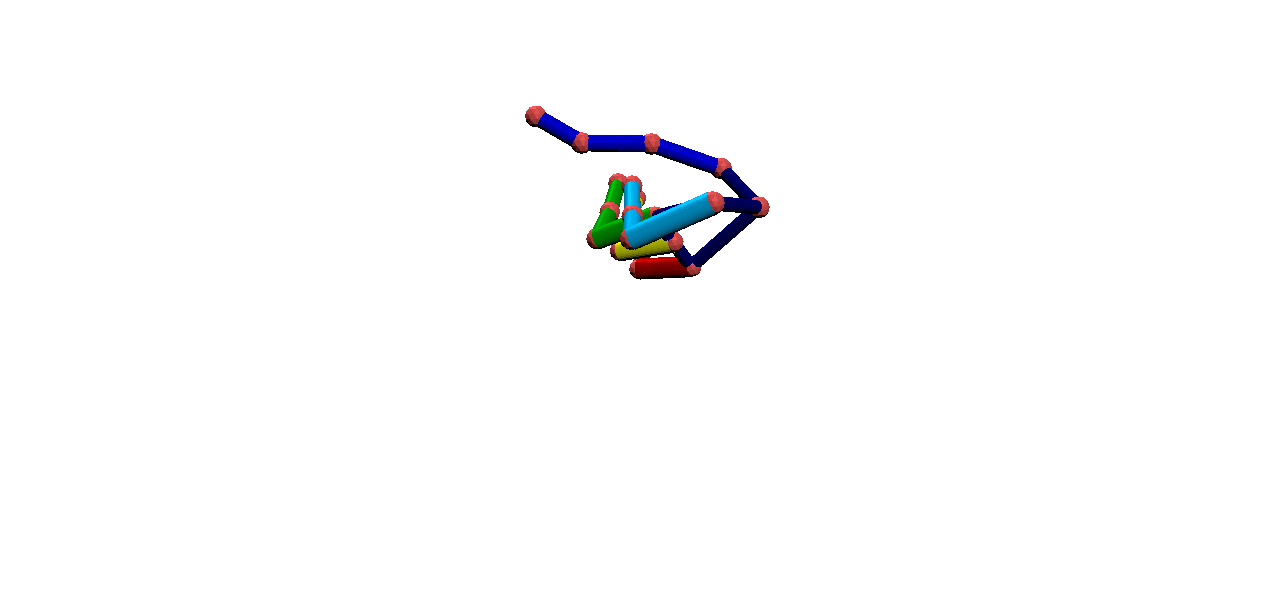}
        \caption{$\mathcal{L}_{\mathrm{BL}}$}
        \label{fig:ablation_bl}
    \end{subfigure}%
    \begin{subfigure}[b]{0.2\textwidth}
        \includegraphics[width=\textwidth, clip, trim=17cm 9.5cm 15cm 3cm]{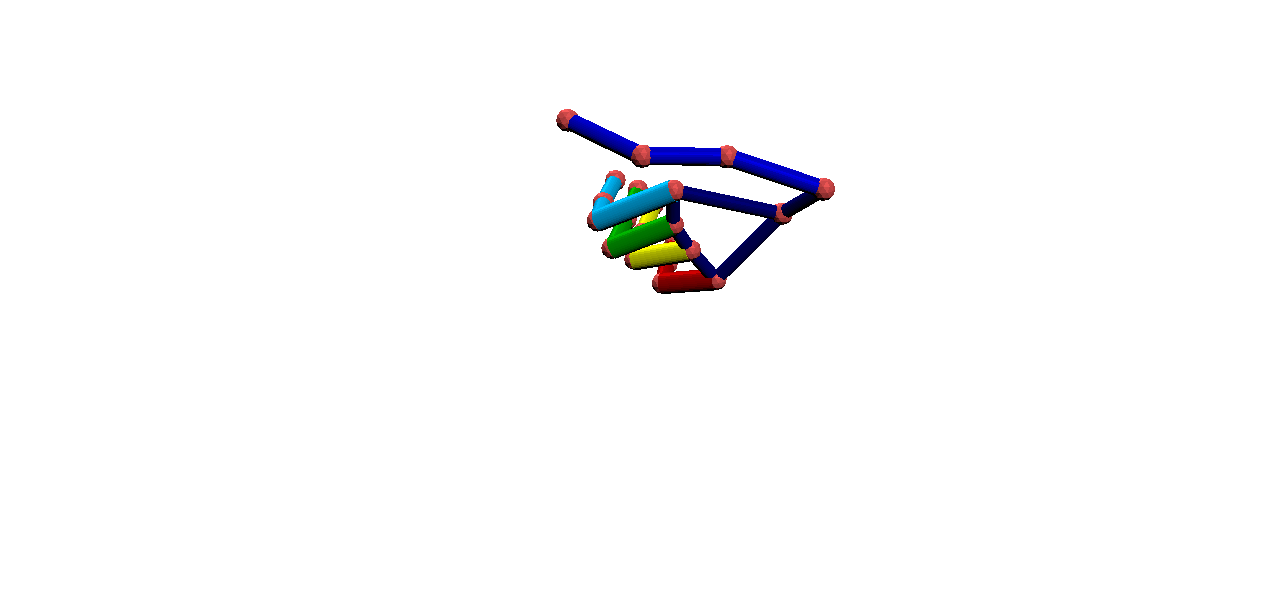}
        \caption{$\mathcal{L}_{\mathrm{BL}} + \mathcal{L}_{\mathrm{RB}}$}
        \label{fig:ablation_blrb}
    \end{subfigure}%
    \begin{subfigure}[b]{0.2\textwidth}
        \includegraphics[width=\textwidth, clip, trim=17cm 9.5cm 15cm 3cm]{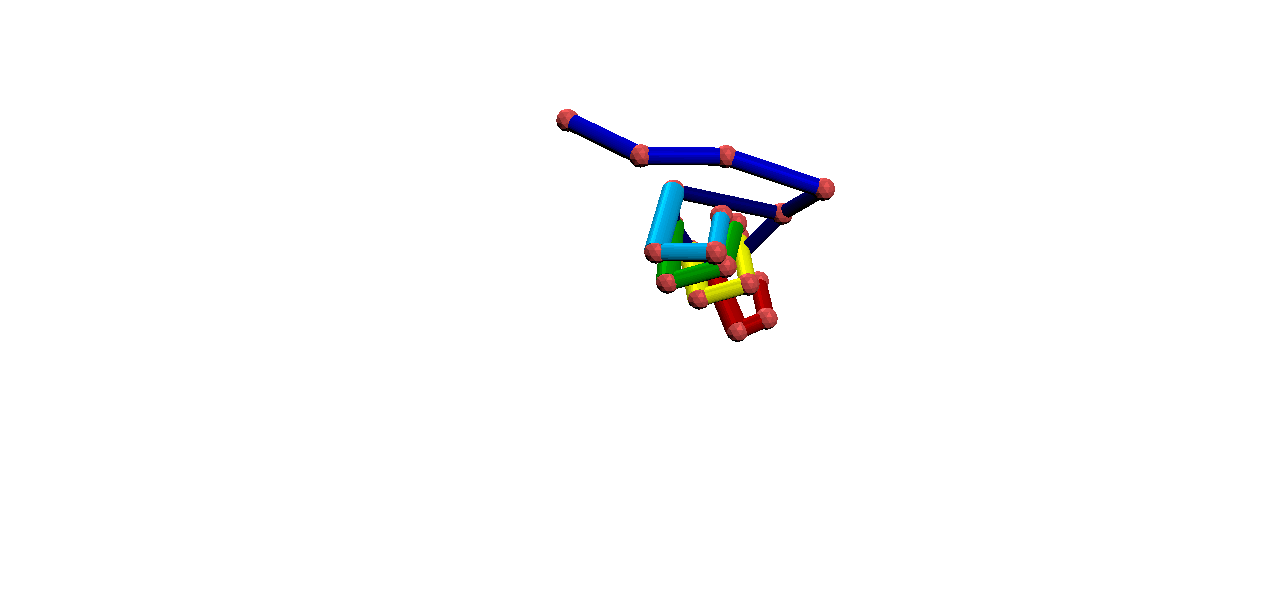}
        \caption{$\mathcal{L}_{\mathrm{BL}} + \mathcal{L}_{\mathrm{RB}}+ \mathcal{L}_{\mathrm{A}}$}
        \label{fig:ablation_blrba}
    \end{subfigure}

    \caption{Impact of our proposed losses. (a) All predicted 3D poses project to the same 2D pose. (b) Ground-truth pose. (c) $\mathcal{L}_{\mathrm{BL}}$ results in poses that have correct bone lengths, but may have invalid angles and palm structure. (d) Including $\mathcal{L}_{\mathrm{RB}}$ imposes a correct palm, but the fingers are still articulated wrong. (e) Adding $\mathcal{L}_{\mathrm{A}}$ leads to the finger bones having correct angles. The resulting hand is plausible and close to the ground-truth.
    }
    \label{fig:ablation_qualitative}
\end{figure*}

\begin{wraptable}{r}{0.5\textwidth}
\label{tab:ablation_fh_refinement}
\begin{center}
\vspace{-4mm}
\caption{Effect of $Z^{root}$ refinement}
\scriptsize
\begin{tabular}{lccc}
\toprule
 \multirow{2}{2.5cm}{Ablation Study} & \multicolumn{2}{c}{EPE (mm)}    & \multirow{2}{1.0cm}{AUC $\uparrow$}   \\
 & \multirow{1}{1.0cm}{mean $\downarrow$} & \multirow{1}{1.0cm}{median~$\downarrow$ }   \\
\midrule
w/o refinement     & 11.20 & 8.62 & 0.95 \\
w. refinement (\textbf{ours})     & \textbf{9.76} & \textbf{8.14} & \textbf{0.97} \\ 
\midrule
\end{tabular}
\label{tab:ablation_fh_refinement}

\end{center}
\end{wraptable}
\textbf{Refinement network.} \tabref{tab:ablation_fh_refinement} shows the impact of $Z^{root}$ refinement (Sec.~\ref{sec:refinement}). We train two models that include (w. refinement) or omit (w/o refinement) the refinement step, using full supervision on FH ($\mathbf{3D}_\mathrm{FH}$). Using refinement, the mean error is reduced by $1.44$mm which indicates that refining effectively reduces outliers. 

\begin{wraptable}{r}{0.5\textwidth}
\begin{center}
\vspace{-4mm}
\caption{Effect of BMC components.}
\scriptsize
\begin{tabular}{lccc}
\toprule
  \multirow{2}{2.5cm}{Ablation Study} & \multicolumn{2}{c}{EPE (mm)}    & \multirow{2}{1.0cm}{AUC $\uparrow$}   \\
 & \multirow{1}{1.0cm}{mean $\downarrow$} & \multirow{1}{1.0cm}{median~$\downarrow$ }   \\
\midrule
$\mathbf{3D_\text{RHD}}$ + $\mathbf{2D_\text{FH}}$                                            & 20.92 & 16.93 & 0.81      \\ 
+ $\mathcal{L}_{\mathrm{BL}}$ (\textbf{ours})                                                       & 17.58 & 14.81 & 0.88    \\ 
~~~~~~+~$\mathcal{L}_{\mathrm{RB}}$ (\textbf{ours})                                                 & 15.48 & 13.49 & 0.91    \\ 
~~~~~~~~~~~~+~$\mathcal{L}_{\mathrm{A}}$ (\textbf{ours})                                            & \textbf{13.78} & \textbf{11.61} & \textbf{0.92}    \\ \hdashline
$\mathbf{3D_\text{RHD}}$ + $\mathbf{3D_\text{FH}}$                                      & 8.78  & 7.25  & 0.98  \\
\midrule
\end{tabular}
\label{tab:ablation_fh_component}
\end{center}
\end{wraptable}

\vspace{-3mm}
\textbf{Components of BMC.} In~\tabref{tab:ablation_fh_component}, we perform a series of experiments where we incrementally add each of the proposed constraints. For 3D guidance, we use the synthetic RHD and \emph{only} use the 2D labels of FH. We first run the baseline model trained only on this data ($\mathbf{3D}_\mathrm{RHD} + \mathbf{2D}_\mathrm{FH}$). Next, we add the bone length loss $\mathcal{L}_{\mathrm{BL}}$, followed by the root bone loss $\mathcal{L}_{\mathrm{RB}}$ and the angle loss $\mathcal{L}_\mathrm{A}$. An upper bound is given by our model trained fully supervised on both datasets ($\mathbf{3D}_\mathrm{RHD} + \mathbf{3D}_\mathrm{FH}$). Each component contributes positively towards the final performance, totalling a decrease of $6.24$mm in mean error as compared to our weakly-supervised baseline, significantly closing the gap to the fully supervised upper bound. A qualitative assessment of the individual losses can be seen in~\figref{fig:ablation_qualitative}.

\begin{wraptable}{r}{0.5\textwidth}
\begin{center}
\vspace{-4mm}
\caption{Effect of angle constraints}
\scriptsize
\begin{tabular}{lccc}
\toprule
\multirow{2}{2.5cm}{Ablation Study}  & \multicolumn{2}{c}{EPE (mm)}    & \multirow{2}{1.0cm}{AUC $\uparrow$}   \\
 & \multirow{1}{1.0cm}{mean $\downarrow$} & \multirow{1}{1.0cm}{median~$\downarrow$ }   \\
\midrule
Independent     & 15.57 & 13.45 & 0.91 \\ 
Dependent       & 13.78	& 11.61	& 0.92 \\
\midrule
\end{tabular}
\label{tab:ablation_fh_angles}

\end{center}
\end{wraptable}
\textbf{Co-dependency of angles.} In~\tabref{tab:ablation_fh_angles}, we show the importance of modeling the dependencies between the flexion and abduction angle limits (Sec. \ref{sec:method}), instead of regarding them independently. Co-dependent angle limits yield a decrease in mean error of $1.40$ mm.

\begin{wraptable}{r}{0.5\textwidth}
\begin{center}

\vspace{-4mm}
\caption{Effect of limits}
\scriptsize
\begin{tabular}{lccc}
\toprule
 \multirow{2}{2.5cm}{Ablation Study} & \multicolumn{2}{c}{EPE (mm)}    & \multirow{2}{1.0cm}{AUC $\uparrow$}   \\
 & \multirow{1}{1.0cm}{mean $\downarrow$} & \multirow{1}{1.0cm}{median~$\downarrow$ }   \\
\midrule
Approximated    & 16.14 & 13.93 & 0.90 \\ 
Computed        & 13.78	& 11.61	& 0.92 \\
\midrule
\end{tabular}
\label{tab:ablation_fh_limits}
\end{center}
\end{wraptable}

\textbf{Constraint limits.} In~\tabref{tab:ablation_fh_limits}, we investigate the effect of the used limits on the final performance, as one may have to resort to approximations. For this, we instead take the hand parameters from RHD and perform the same weakly-supervised experiment as before ($+\mathcal{L}_\mathrm{\mathbf{BMC}}$). Approximating the limits from another dataset slightly increases the error, but still clearly outperforms the 2D baseline.

\subsection{Bootstrapping with Synthetic Data}
\vspace{-2mm}
\label{sec:weaksup_synth}
We validate $\mathcal{L}_\mathrm{\mathbf{BMC}}$ on the \textbf{test set} of FH and HO-3D. We train the same four models like in Sec. \ref{sec:weaksup_effect} using fully supervised RHD and weakly-supervised real data R$\in$[FH,HO-3D].

For all results here we perform training on the \textit{full} dataset and evaluate on the official test split via the online submission system. Additionally, we evaluate the cross-dataset performance on D+O dataset to show how our proposed constraints improves generalizability and compare with prior work \cite{iqbal2018hand, mueller2018ganerated, boukhayma20193d, zhang2019end, baek2019pushing}.

\begin{table}[t]
\centering
\caption{Results on the respective \textbf{test} split, evaluated by the \textit{submission systems}. Training on RHD leads to poor accuracy on both FH and HO-3D. Adding weakly-supervised data improves results, as expected. By including our proposed $\mathbf{\mathcal{L}_\mathrm{\mathbf{BMC}}}$, our model incurs a significant boost in accuracy, especially evident for the INTERP score.}
\label{tbl:bootstrap_real}
\scriptsize
\begin{tabularx}{\columnwidth}{XX|cc|cc}
\toprule
 & \multirow{2}{1.2mm}{Description} & \multicolumn{2}{c}{R=FH}               & \multicolumn{2}{c}{R=HO-3D}                             \\
 & & mean $\downarrow$ & AUC $\uparrow$   &  EXTRAP $\downarrow$ & INTERP $\downarrow$                \\ 
\midrule
$\mathbf{3D_\mathrm{RHD} + 3D_\mathrm{R}}$                       & Fully sup. upper bound      & 0.90 & 0.82 & 18.22 & 5.02        \\ \hdashline
$\mathbf{3D_\mathrm{RHD}}$                                        & Fully sup. lower bound      & 1.60 & 0.69 & 20.84 & 33.57       \\ 
+$\mathbf{2D_\mathrm{R}}$                       & + Weakly sup.               & 1.26 & 0.75 & 19.57 & 25.16       \\ 
~~~~+~$\mathbf{\mathcal{L}_\mathrm{\mathbf{BMC}}}$ (\textbf{ours}) & ~~~~+~BMC     & \textbf{1.13} & \textbf{0.78} & \textbf{18.42} & \textbf{10.31}   \\ 
\bottomrule
\end{tabularx}

\end{table}

\textbf{FH.}
The second column of \tabref{tbl:bootstrap_real} shows the dataset performance for R = FH. Training solely on RHD ($\mathbf{3D_\mathrm{RHD}}$) performs the worst. Adding real data ($+ \mathbf{2D_\mathrm{FH}}$) with 2D labels reduces the error, as we reduce the real/synthetic domain gap. Including the proposed $\mathcal{L}_\mathrm{\mathbf{BMC}}$ results in an accuracy boost.

\textbf{HO-3D.}
The third column of \tabref{tbl:bootstrap_real} shows a similar trend for R = HO-3D. Most notably, our constraints  yield a decrease of $14.85$ mm for INTERP. This is significantly larger than the relative decrease the 2D data adds (-$8.41$mm). For EXTRAP, BMC yields an improvement of $1.15$mm, which is close to the $1.27$mm gained from 2D data. This demonstrates that $\mathcal{L}_\mathrm{\mathbf{BMC}}$ is beneficial in leveraging 2D data more effectively in unseen scenarios.

\begin{table}[t]
\centering
\vspace{-4mm}
\caption{Datasets used by prior work for evaluation on D+O. With solely fully-supervised synthetic and weakly-supervised real data, we outperform recent works and perform on par with \cite{zhang2019end}. All other works rely on full supervision from real and synthetic data. %
*These works report unaligned results.}
\label{tbl:weaksup_crossdata_datasets}
\scriptsize
\begin{tabularx}{\columnwidth}{Xccc|c}
\toprule
\multirow{2}{*}{D+O}                                                & \multicolumn{3}{c}{Annotations used}      &       \\
                                                                    & Synth.    & Real      & Scans             & AUC  $\uparrow$ \\
\midrule
\textbf{Ours} (weakly sup.)                                         & 3D        & \textbf{2D only}   &                   & \textbf{0.82}  \\ \hdashline
Zhang (2019) \cite{zhang2019end}                                    & 3D        & 3D        & 3D                & 0.82  \\
Boukhayma (2019) \cite{boukhayma20193d}                             & 3D        & 3D        & 3D                & 0.76  \\
Iqbal (2018)* \cite{iqbal2018hand}                                  & 3D        & 3D        &                   & 0.67  \\ 
Baek (2019)* \cite{baek2019pushing}                                 & 3D        & 3D        & 3D                & 0.61  \\
Zimmermann (2018)\cite{zimmermann2017}                             & 3D        & 3D        &                   & 0.57  \\
Spurr (2018) \cite{spurr2018cvpr}                                   & 3D        & 3D        &                   & 0.51  \\
Mueller (2018)* \cite{mueller2018ganerated}                         & 3D        & {\scriptsize Unlabeled\par}&  & 0.48  \\ 
\bottomrule
\end{tabularx}
\end{table}

\textbf{D+O.}
In \tabref{tbl:weaksup_crossdata_datasets} we demonstrate the cross-data performance on D+O for R = FH.
Most recent works have made use of MANO \cite{boukhayma20193d, zhang2019end, baek2019pushing}, leveraging a low-dimensional embedding of highly detailed hand scans and require custom synthetic data \cite{baek2019pushing, boukhayma20193d} to fit the shape. Using only fully supervised \textit{synthetic data} and \textit{weakly-supervised} real data in conjunction with $\mathcal{L}_\mathrm{\mathbf{BMC}}$, we reach state-of-the-art.

\subsection{Bootstrapping with Real Data}
\label{sec:weaksup_real}
We study the impact of our biomechanical constrains on reducing the number of labeled samples required in scenarios where few real 3D labeled samples are available. We train a model in a setting where a fraction of the data contains the full 3D labels and the remainder contains only 2D supervision.
\begin{wrapfigure}{r}{0.5\textwidth}
    \centering
    \vspace{4mm}
    \includegraphics[width=0.5\textwidth]{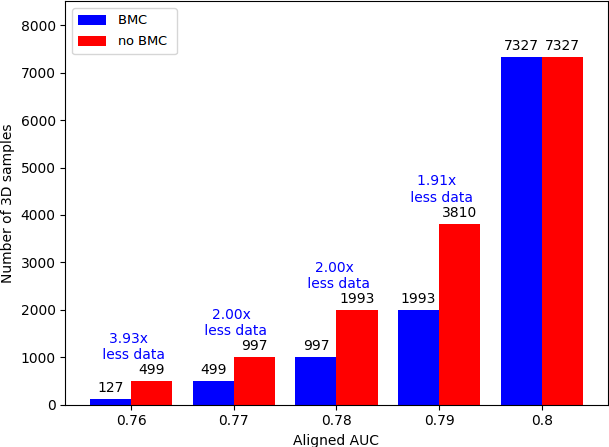}
    \caption{Number of 3D samples required to reach a certain aligned AUC on FH.}
    \label{fig:fh_semi_barplot}
\end{wrapfigure}

Here we choose $R = $ FH, use the entire training set and evaluate on the test set. For each fraction of fully labelled data we evaluate two models. The first is trained on both the fully and weakly labeled samples. The second is trained with the addition of our proposed constraints. We show the results in \figref{fig:fh_semi_barplot}. For a given AUC, we plot the number of labeled samples required to reach it. We observe that for lower labeling percentages, the amount of labeled data required is approximately \textit{half} using $\mathcal{L}_\mathrm{\mathbf{BMC}}$. This showcases its effectiveness in low label settings and demonstrates the decrease in requirement for fully annotated training data.

\section{Conclusion}
We propose a set of fully differentiable biomechanical losses to more effectively leverage weakly supervised data. Our method consists of a novel procedure to encourage anatomically correct predictions of a backbone network via a set of novel losses that penalize invalid bone length, joint angles as well as palmar structures. Furthermore, we have experimentally shown that our constraints can more effectively leverage weakly-supervised data, which show improvement on both within- and cross-dataset performance. Our method reaches state-of-the-art performance on the aligned D+O objective using 3D synthetic and 2D real data and reduces the need of training data by half in low label settings on FH.
\newline
\newline
\noindent\textbf{Acknowledgments.} We are grateful to Christoph Gebhardt and Shoaib Ahmed Siddiqui for the aid in figure creation and Abhishek Badki for helpful discussions.
\newpage

\Urlmuskip=0mu plus 1mu\relax
{\small
\bibliographystyle{splncs04}
\bibliography{egbib}
}

\newpage
\title{Supplementary:\\Weakly Supervised 3D Hand Pose Estimation via Biomechanical Constraints}
\titlerunning{Supplementary: Weakly Supervised 3D Hand Pose Estimation via BMC}
\authorrunning{A. Spurr et al.}
\author{}
\institute{}
\maketitle

Here we provide additional implementation details and more experimental comparisons. In Section 1 we describe details on how the angle loss is computed and the joint angle interdependence is modeled. Section 2 repeats the ablation study on additional datasets (HO-3D) to highlight the generalizability of results. Section 3 demonstrates the effect of weak-supervision in two additional settings, one using a real dataset as the fully-supervised data and the other using MPII in-the-wild data as weak-supervision. Section 4 compares BMC to an adversarial loss. Sections 5 and 6 provide additional results of bootstrapping via weak-supervision with synthetic or real data. Section 7 shows further qualitative results of using BMC. Sections 8 and 9 provide additional implementation details and results on HANDS2019 challenge, respectively. 

\section{Joint angle loss}
\textbf{Joint angle ambiguity.} The computation of the joint angles lead to ambiguities. More specifically, two different vectors on the unit sphere may map to the same joint angles. 

For example, given two bones $\V{b}^{\V{F}_i, 1}_i = [1,0,1]$ and $\V{{b}}^{\V{F}_i, 2}_i = [\minus1,0,1]$ in a coordinate frame $\V{F}_i$, we have using $\mathrm{P}_{xz}(\V{b}_i^{\mathbf{F}_i,1}) = [1,0,1]$ and $\mathrm{P}_{xz}(\V{b}_i^{\mathbf{F}_i,2}) = [\minus1,0,1]$:
\begin{equation}
    \begin{split}
        \theta^{\mathrm{f},1}_i &= \alpha(\mathrm{P}_{xz}(\V{b}_i^{\mathbf{F}_i,1}), \V{z}_i) = \alpha([1,0,1], \V{z}_i) \\
        &= \pi / 4 \\
        \theta^{\mathrm{a},1}_i &= \alpha(\mathrm{P}_{xz}(\V{b}_i^{\mathbf{F}_i,1}), \V{b}_i^{\mathbf{F}_i,1}) \\
        &= \alpha([1,0,1],[1,0,1]) = 0 \\
        \theta^{\mathrm{f},2}_i &= \alpha(\mathrm{P}_{xz}(\V{b}_i^{\mathbf{F}_i,2}), \V{z}_i) = \alpha([\minus1,0,1], \V{z}_i) \\
        &= \pi / 4 \\
        \theta^{\mathrm{a},2}_i &= \alpha(\mathrm{P}_{xz}(\V{b}_i^{\mathbf{F}_i,2}), \V{b}_i^{\mathbf{F}_i,2}) \\
        &= \alpha([\minus1,0,1],[\minus1,0,1]) = 0
    \end{split}
\end{equation}
Therefore, both bones map to the same angle pair $(\pi/4, 0)$. To resolve this, we perform an octant look up. Given the flexion angle $\theta^{\mathrm{f}}_i$ and abduction angle $\theta^{\mathrm{a}}_i$ of bone $i$, we negate the respective angle if the bone lies within the negative $x$-octant or negative $y$-octant:
\begin{equation}
    \begin{split}
        \theta^{\mathrm{f}}_i &= \begin{dcases}
            \minus\theta^{\mathrm{f}}_i, &\text{if } b_{i,x}^{\V{F}_i} < 0 \\
            \theta^{\mathrm{f}}_i, & \text{else}
        \end{dcases}\\
        \theta^{\mathrm{a}}_i &= \begin{dcases}
            \minus\theta^{\mathrm{a}}_i, &\text{if } b_{i,y}^{\V{F}_i} < 0\\
            \theta^{\mathrm{a}}_i, & \text{else}
        \end{dcases}\\
    \end{split}
    \label{eq:angle_lookup}
\end{equation}
Where $b_{i,x}^{\V{F}_i}$,$b_{i,y}^{\V{F}_i}$ is the $x$/$y$-component of the bone vector given in coordinates of its local coordinate frame $\V{F}_i$. This leads to angles in the range $\theta^{\mathrm{f}}_i \in [\minus\pi,\pi]$ and $\theta^{\mathrm{a}}_i \in [\minus\pi/2,\pi/2]$ respectively.

\begin{figure}[t]
    \centering
    \includegraphics[width=\columnwidth*2/3]{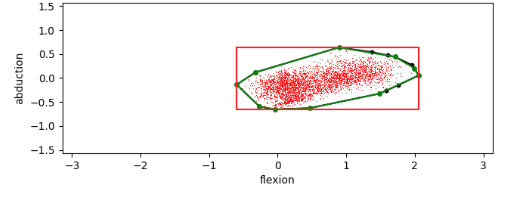}
    \caption{$(\theta^{\mathrm{f}}, \theta^{\mathrm{a}})$-plane. Green: $\mathcal{H}_i$. Red: min/max-box}
    \label{fig:angle_convex_hull}
\end{figure}

\textbf{Approximation of Convex Hull.} 
\figref{fig:angle_convex_hull} plots the distribution of the pinkys MCP flexion/extension angles of the FH dataset, visualized as red points. The red rectangle corresponds to the valid range of angles when considering both angle limits independently. Hence the corners correspond to $(\min^f_i,\min^a_i), (\min^f_i,\max^a_i), (\max^f_i,\max^f_i), (\max^f_i, \min^a_i)$ in counter-clockwise order, where $\min^k_i,\max^k_i$ corresponds to the minimum/maximum of angle $\theta^k_i$, where $k \in \{a,f\}$.

In order to take the dependence of the angle limits in account, we first compute the convex hull of the angle points. However, depending on the shape of the point cloud, the number of points lying on the hull can vary and be numerous. In order to keep the number of hull points low and consistent for all joint angles, we approximate this hull in two steps. We first employ the Ramer-Douglas-Peucker algorithm, a polygon simplification algorithm. This significantly reduces the number of vertices in the hull, but still results in a variable number. To ensure consistency, we apply a greedy algorithm that iteratively removes points such that the hull encompasses as many points as possible until we reach the desired number of points, resulting in our approximation $\mathcal{H}_i$. For all our experiments, we set number of points to be $10$. The green polygon in \figref{fig:angle_convex_hull} displays this approximation to the convex hull.

\textbf{Distance computation.}
To compute the distance $\mathcal{H}_i$, we compute two values. The first indicates if an angle point $\boldsymbol{\theta}_i$ is contained within the hull. The second corresponds to the distance to the hull. Here we detail how we compute both values.
For ease of notation, we assume that the points in $\mathcal{H}_i$ are ordered counter-clockwise beginning from any point in $\mathcal{H}_i$. Let $\mathcal{H}_{i,k}$ be the $k$-th point in $\mathcal{H}_i$. An edge $\V{v}_k$ of the hull is given as:
\begin{equation}
\begin{split}
    \V{v}_k &= \mathcal{H}_{i,k+1} - \mathcal{H}_{i,k}, \text{ for } k \in [1,10] \\
    \V{w}_k &= \boldsymbol{\theta}_i - \mathcal{H}_{i,k}, \text{ for } k \in [1,10]
\end{split}
\end{equation}
Where we define $\mathcal{H}_{i,11} = \mathcal{H}_{i,1}$ to wrap around the hull. 

To compute if a point $\boldsymbol{\theta}_i$ is contained within $\mathcal{H}_i$, we exploit the convexity of the hull and make use of the cross-product. Specifically, we compute the 2D cross-product between $\V{v}_k$ and $\V{w}_k$. Intuitively, if the cross-product $\V{w}_k \times \V{v}_k$ is positive for any given edge $k$, then the angle point lies outside of the hull. If its negative for all, it is contained within. If it lies on the hull, we consider it to be contained within it. More formally:
\begin{equation}
\label{eq:supp_contains}
    c = \prod_{k=1}^{10} \mathds{1}_{(\V{w}_k \times \V{v}_k) \leq 0 } 
\end{equation}

To compute the distance of $\boldsymbol{\theta}_i$ to the hull, we compute its distance to each edge and take the minimum. Given edge $\V{v}_k$ and point $\boldsymbol{\theta}_i$, their distance is the minimum distance between either endpoints of $\V{v}_k$ or the projection of $\V{w}_k$ onto $\V{v}_k$. Formally:
\begin{equation}
    \begin{split}
        &t = \max(0, \min(1, \V{w}_k^T\V{v}_k / ||\V{v}_k||^2_2)) \\ 
        &\V{p}_k = \mathcal{H}_{i,k} + t \V{v}_k \\ 
        &D(\V{v}_k, \boldsymbol{\theta}_i) = |\cos(\boldsymbol{\theta}_i) - \cos(\V{p}_k)| + |\sin(\boldsymbol{\theta}_i) - \sin(\V{p}_k)|
    \end{split}
\end{equation}
Where the min/max ensures that we do not extend beyond the endpoints of $\V{v}_k$. Given the distance to the edge, we can compute the distance to the hull $\mathcal{H}_i$:
\begin{equation}
    D(\boldsymbol{\theta}_i, \mathcal{H}_i) = \min_k D(\V{v}_k, \boldsymbol{\theta}_i)
\end{equation}
This formulation computes the distance towards $\mathcal{H}_i$, whether the point is contained or not. We do not want to penalize points that lie within the hull, as that constitutes our range of valid angles. Therefore we make use of the quantity $c$ computed in Eq. \ref{eq:supp_contains}, which leads to the final angle loss function:
\begin{equation}
    D_A(\boldsymbol{\theta}_i, \mathcal{H}_i) = (1 - \mathds{1}_c) D(\boldsymbol{\theta}_i, \mathcal{H}_i)
\end{equation}
This returns a loss of $0$ if the angle point $\boldsymbol{\theta}_i$ is contained, otherwise it returns the distance to the approximation of the convex hull $\mathcal{H}_i$. This constitutes our angle loss for bone $i$.

\section{Ablation study}
We repeat the ablation study with the HO-3D dataset. All evaluations are done on a custom split, where we manually extract two sequences for the test and use the remainder for the training set. Each error is computed for the root relative case. 

\textbf{Refinement network.} We train two models using full supervision on HO-3D ($\mathbf{3D}_\mathrm{HO3D}$). The first model (w/o refinement) does not use the proposed refinement network, whereas the second does (w.refinement). We showcase the performance difference in the first row of \tabref{tbl:ablation_ho3d}. We note a reduction of $2.97$mm mean error when using the refinement network. 

\textbf{BMC ablation.}
We study the individual contribution of the BMC losses. We bootstrap the 3D annotation from synthetic data and use only the 2D annotation of HO-3D. The first model constitutes our baseline, which is trained only on that data ($\mathbf{3D}_\mathrm{RHD} + \mathbf{2D}_\mathrm{HO3D}$). We incrementally add the bone length loss $\mathcal{L}_{\mathrm{BL}}$, the root bone loss $\mathcal{L}_{\mathrm{RB}}$ and lastly the angle loss $\mathcal{L}_\mathrm{A}$. We train a fully supervised model ($\mathbf{3D}_\mathrm{RHD} + \mathbf{3D}_\mathrm{HO3D}$) which is our upper bound. We refer to the second section of \tabref{tbl:ablation_ho3d}. Each loss contributes towards a reduction in mean error, culminating in a total decrease of $5.21$mm as compared to our 2D only baseline.

\begin{figure}[t]
    \centering
    \includegraphics[width=\columnwidth*2/3]{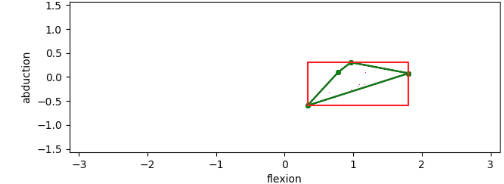}
    \caption{$(\theta^{\mathrm{f}}, \theta^{\mathrm{a}})$-plane. Green: $\mathcal{H}_i$. Red: min/max-box }
    \label{fig:angle_convex_hull_ho3d}
\end{figure}

\textbf{Co-dependency between angles.} 
We train two models. The first models the angle limits independently, whereas the second takes the dependency of the limits into account. The resulting performance is shown in \tabref{tbl:ablation_ho3d}. We note a minor performance degradation. We attribute this to the extremely limited angle range contained in the HO-3D dataset. As it contains subjects holding various object in a gripping pose while rotating it in front of the camera, the actual angles of the fingers do not change. Therefore the range of angles across the dataset is low, which leads to a very tight angle limit. This does not generalize well, which in turn hurts performance. \figref{fig:angle_convex_hull_ho3d} displays the angle-plane plot for HO-3D using the pinkys MCP flexion/extension angles. Comparing with \figref{fig:angle_convex_hull}, which plots the plane for the same finger for FH, we see that the resulting range of HO-3D is a lot more severely limited. This is to be expected, as HO-3D is a very constrained dataset due to the aforementioned reason. 

\textbf{BMC limits.} 
We study the effect of approximating the BMC limits when using a different dataset to compute these values. We compute the hand parameters from RHD and perform the same weakly-supervised experiment as previously ($\mathbf{3D}_\mathrm{RHD} + \mathbf{2D}_\mathrm{HO3D}$). As can be seen in the last row of \tabref{tbl:ablation_ho3d}, we note a slight increase in loss, however it still clearly outperforms the 2D baseline in mean error ($18.50$ mm vs $23.71$ mm).

\begin{table}[t]
\centering
\caption{Ablation studies on \textbf{validation} split of HO-3D. The models of the first section was trained on our train split of HO-3D.}
\label{tbl:ablation_ho3d}
\scriptsize
\begin{tabularx}{\columnwidth}{Xccc}
\toprule
 \multirow{3}{6cm}{HO-3D}  & \multicolumn{3}{c}{3D Pose Estimation (root-relative)}  \\
 & \multicolumn{2}{c}{EPE (mm)}    & \multirow{2}{1.2cm}{AUC $\uparrow$}   \\
 & \multirow{1}{1.2cm}{mean $\downarrow$} & \multirow{1}{1.2cm}{median~$\downarrow$ }   \\
\midrule
\multicolumn{4}{c}{Effect of $Z^{root}$ refinement} \\
\midrule
w/o refinement     & 25.34 & 24.39 & 0.79 \\
w. refinement      & 22.37 & 23.01 & 0.83 \\ 
\midrule
\multicolumn{4}{c}{Effect of BMC components} \\
\midrule
$\mathbf{3D_\text{RHD}}$ + $\mathbf{2D_\text{HO3D}}$                        & 23.71 & 22.07 & 0.78 \\ \hdashline
+ $\mathcal{L}_{\mathrm{BL}}$                                               & 22.15 & 20.27 & 0.80 \\ 
~~~~~~+~$\mathcal{L}_{\mathrm{RB}}$                                         & 18.83 & 17.79 & 0.87 \\ 
~~~~~~~~~~~~+~$\mathcal{L}_{\mathrm{A}}$                                    & 18.50 & 17.41 & 0.87 \\ \hdashline
$\mathbf{3D_\text{RHD}}$ + $\mathbf{3D_\text{HO3D}}$                        & 16.74 & 16.94 & 0.89 \\
\midrule
\multicolumn{4}{c}{Effect of angle co-dependency} \\
\midrule
Independent     & 18.30 & 17.40 & 0.87 \\ 
Dependent       & 18.50 & 17.41 & 0.87 \\
\midrule
\multicolumn{4}{c}{Effect of BMC limits} \\
\midrule
Approximated    & 19.21 & 17.88 & 0.86 \\ 
Computed        & 18.50 & 17.41 & 0.87 \\
\bottomrule
\end{tabularx}
\end{table}

\section{Effect of Weak-Supervision}
We repeat the experiments of Section 5.3 in the main paper using different datasets. We show that the effect of weak-supervision also holds when using fully labeled real data or weakly-labeled in-the-wild data.

\textbf{STB.} We reproduce the results of Section 5.3 in the main paper, but instead of using RHD we use STB \cite{zhang2016} as the fully supervised dataset. The weakly-supervised dataset remains FH. The purpose of this experiment is to demonstrate that the effect of weak supervision also takes place when using a real dataset for full supervision. Table \ref{tbl:weaksup_effect_both} (top) shows the result.

\textbf{MPII - in-the-wild dataset.} We reproduce the results of Section 5.3 in the main paper, but using MPII \cite{simon2017hand} as our weakly-supervised dataset. This is to demonstrate the effect of weak-supervision stemming from datasets collected in-the-wild, a potentially useful supervision source. We evaluate on the validation split of FH. Table \ref{tbl:weaksup_effect_both} (bottom) shows the result. Note that as the MPII dataset only contains 2D labels and no 3D annotation is provided, the fully supervised upper bound cannot be performed and is therefor omitted from the table.

\begin{table}[t]
\centering
\caption{This table show-cases the same effect of weak-supervision as Table 2 in the main paper but evaluated in different settings. All models are evaluated on the \textbf{validation} split of FH. (top) We use STB as the fully labeled dataset and supplement is using weakly-labeled FH. (bottom) We use RHD as the fully labeled dataset and MPII as the weakly-supervised data. The same trend can be observed in both settings. Adding weakly-supervised data improves 3D prediction performance due to predicted 3D poses with the correct 2D projection. By incorporating our proposed biomechanically constraints we significantly improve 3D pose accuracy due to more accurate $\V{Z}$. Note that as the MPII dataset only contains 2D labels and no 3D annotation is provided, the fully supervised upper bound cannot be performed and is therefor omitted from the table.}
\label{tbl:weaksup_effect_both}
\scriptsize
\begin{tabularx}{\columnwidth}{XX|ccc}
\toprule
\multirow{3}{6cm}{Effect of weak-supervision} & \multirow{3}{6cm}{Description} &  \\
                                                & &\multicolumn{3}{c}{mean $\downarrow$}         \\
  & &\multirow{1}{1.3cm}{2D (pixel)} & \multirow{1}{1.3cm}{Z (mm)} & \multirow{1}{1.3cm}{3D (mm)} \\
  \midrule
  \multicolumn{5}{c}{\textbf{3D labels: STB}} \\
\midrule
$\mathbf{3D_\mathrm{STB} + 3D_\mathrm{FH}}$                         & Fully supervised, real    & 3.85  & 5.68   & 9.05         \\
+ $\mathbf{\mathcal{L}_\mathrm{\mathbf{BMC}}}$ (\textbf{ours})      & + BMC                     & \textbf{3.83}  & \textbf{5.50}   & \textbf{8.89}          \\ \hdashline
$\mathbf{3D_\mathrm{STB}}$                                          & Fully sup. lower bound    & 20.45 & 36.80  & 54.92        \\
+ $\mathbf{2D_\mathrm{FH}}$                                         & + Weakly supervised, real             & \textbf{3.86}  & 35.41  & 42.02        \\ 
~~~+ $\mathbf{\mathcal{L}_\mathrm{\mathbf{BMC}}}$ (\textbf{ours})   & ~~~~+~BMC   & 3.88  & \textbf{11.17}  & \textbf{18.58}        \\
\midrule
 \multicolumn{5}{c}{\textbf{2D labels: MPII}} \\
 \midrule
$\mathbf{3D_\mathrm{RHD}}$                                          & Fully supervised, synthetic only      & 12.35 & 20.02 & 30.82 \\
+ $\mathbf{2D_\mathrm{MPII}}$                                       & + Weakly supervised, real             & 10.36 & 19.77 & 28.81 \\ 
~~~+ $\mathbf{\mathcal{L}_\mathrm{\mathbf{BMC}}}$ (\textbf{ours})   & ~~~~+~BMC                             & \textbf{10.35} & \textbf{17.72} & \textbf{27.10} \\
\bottomrule
\end{tabularx}
\end{table}

\section{Comparison with Adversarial loss}
It is intuitive to think of drawing parallels between BMC and an adversarial loss. BMC can be interpreted as a discriminator penalizing poses that do not adhere to the distribution of valid hand poses. However, BMC models the task at hand more closely and only requires the limits, whereas a discriminator requires access to a full dataset of 3D poses. In order to see how a discriminator performs against BMC, we perform an experiment in the same setting as the ablation study. We train on fully supervised RHD and weakly-supervised FH, and evaluate on the validation split of FH. As it has not been shown if and how the adversarial loss works for the task of 3D hand pose estimation, we adapt a model from literature applied to 2D body pose \cite{drover2018can}. In order to adjust to the new setting, we performed a search for the optimal hyperparameters to improve the performance of the discriminator. We show the results in Table \ref{tbl:adv_loss}. As can be seen, BMC outperforms the adversarial loss. We hypothesise this is due to BMC  modeling the task at hand more closely. 

\begin{table}[t]
\centering
\caption{We compare using BMC to an adversarial loss adapted from \cite{drover2018can}. BMC outperforms the adversarial loss. We hypothesise this is due to BMC modeling the task at hand more closely.}
\label{tbl:adv_loss}
\scriptsize
\begin{tabularx}{\columnwidth}{XXccc}
\toprule
 \multirow{3}{*}{Comparison to adversarial loss}  & \multirow{3}{*}{Description} &\multicolumn{3}{c}{3D Pose Estimation (root-relative)}  \\
 & &\multicolumn{2}{c}{EPE (mm)}    & \multirow{2}{*}{AUC $\uparrow$}   \\
 & &\multirow{1}{*}{mean $\downarrow$} & \multirow{1}{*}{median~$\downarrow$ } \\
\midrule
$\mathbf{3D_\text{RHD}} + \mathbf{2D_\text{FH}}$                                         & Baseline  & 20.92 & 16.93 & 0.81      \\ 
$\mathbf{3D_\mathrm{RHD} + 2D_\mathrm{FH}}$ + $\mathbf{\mathcal{L}_\mathrm{\mathbf{BMC}}}$  & BMC           & \textbf{15.48} & \textbf{13.49} & \textbf{0.91}    \\ 
$\mathbf{3D_\mathrm{RHD} + 2D_\mathrm{FH}}$ + $\mathbf{\mathcal{L}_\mathrm{\mathbf{adv}}}$  & Adversarial   & 17.60  & 14.38  & 0.87 \\ 
\bottomrule
\end{tabularx}
\end{table}

\section{Bootstrapping with Synthetic Data}
We show the full results of the online evaluation on FH and HO-3D in Table \ref{tbl:bootstrap_synth_all}.

\begin{table}[t]
\centering
\caption{Bootstrapping results on the respective \textbf{test split}, as evaluated by the \textit{online submission system}. Results are given in mm.}
\label{tbl:bootstrap_synth_all}
\scriptsize
\begin{tabularx}{\columnwidth}{lXcccc}
\toprule
\multirow{2}{*}{FH} & \multirow{2}{*}{Description}  & \multicolumn{2}{c}{aligned} & \multicolumn{2}{c}{unaligned} \\
 & & mean $\downarrow$ & AUC $\uparrow$ & mean $\downarrow$ & AUC $\uparrow$ \\ 
\midrule
Zimmermann et al.\cite{zimmermann2019iccv} & fully supervised FH           & 1.10 & 0.78 & \textbf{7.13} & 0.19 \\ 
$\mathbf{3D_\mathrm{RHD} + 3D_\mathrm{FH}}$ & fully supervised RHD/FH   & \textbf{0.90} & \textbf{0.82} &  7.54 & \textbf{0.20} \\  \midrule
$\mathbf{3D_\mathrm{RHD}}$ & fully supervised RHD                & 1.60 & 0.69 & 15.15 & 0.06 \\ 
+~$\mathbf{2D_\mathrm{FH}}$ & + weakly-supervised FH          & 1.26 & 0.75 & 13.02 & 0.14 \\ 
~~~+~$\mathbf{\mathcal{L}_\mathrm{\mathbf{BMC}}}$ & ~~~+~BMC     & \textbf{1.13} & \textbf{0.78} & \textbf{10.39} & \textbf{0.15} \\ 
\midrule
HO3D  & Description    & EXTRAP $\downarrow$   & INTERP $\downarrow$   & OBJECT $\downarrow$   & SHAPE $\downarrow$    \\
\midrule
$\mathbf{3D_\mathrm{RHD} + 3D_\mathrm{HO3D}}$ & fully supervised HO3D                              & 18.22     & 5.02      &   16.56   &   10.79   \\ \midrule
$\mathbf{3D_\mathrm{RHD}}$ & fully supervised RHD                                                 & 20.84     & 33.57     &   35.08   &   23.94   \\ 
+~$\mathbf{2D_\mathrm{HO3D}}$ & + weakly supervised HO3D                                              & 19.57     & 25.16     &   25.79   &   21.05   \\ 
~~~+~$\mathbf{\mathcal{L}_\mathrm{\mathbf{BMC}}}$ & ~~~+~BMC                         & \textbf{18.42}     & \textbf{10.31}     &   \textbf{19.91}   &   \textbf{12.51}   \\ 

\bottomrule
\end{tabularx}
\end{table}

\section{Bootstrapping with Real Data}
\tabref{tbl:weaksup_real_fh} shows the full result of Bootstrapping with real data, as evaluated by the online submission system \footnote{\url{https://competitions.codalab.org/competitions/21238}}. Recall that we assume the remainder of the data to be weakly-supervised, i.e it contains the 2D annotation. We list the exact number of 3D labeled samples used, in addition to the percentage wrt. to the entire dataset it corresponds to. Note that the percentage values have been rounded for readability, but the number of samples is exact. We divide the table according to three categories \begin{inparaenum}[a)]
\item \textbf{Aligned / Unaligned} - Procrustes analysis is used to align before computing the score
\item \textbf{Mean / AUC} - The AUC is given for PCK values that lie in an interval from 0 mm to 50 mm with 100 equally spaced thresholds.
\item \textbf{With / Without BMC} - Using our proposed biomechanical constraints.
\end{inparaenum}
We first focus on the aligned results. Using BMC, the required amount of 3D annotated data for a given AUC is approximately \textit{halved}. This trend continues for labeling percentages up to $\sim13\%$. For example, to achieve the same performance as a model that is trained without BMC on $3810$ 3D labeled data samples, BMC achieves the same performance with $1993$ 3D labeled samples, roughly half the amount. 

A similar trend can be observed for the unaligned score. For labeling percentages up to $6.8\%$ ($1993$), the required amount of data to reach the same performance is approximately halved ($997$).

\newcolumntype{Y}{>{\centering\arraybackslash}X}

\begin{table*}[t]
\centering
\caption{Scores as evaluated on the online submission system. The first column denotes the percentage (in brackets) of 3D annotated samples used during training, where the remainder is annotated only with 2D labels. Note that the percentages are rounded, but the number of samples are exact. \textbf{$+$ indicates the model trained with BMC, $-$ indicates the model trained without it.}}
\label{tbl:weaksup_real_fh}
\small
\begin{tabularx}{\columnwidth}{cc|YYYY|YYYY}
\toprule
\multirow{2}{*}{FH} & \multirow{2}{*}{\shortstack[l]{$+$: with BMC (\textbf{ours}) \\ $-$: without BMC}} & \multicolumn{4}{c}{Aligned} & \multicolumn{4}{c}{Unaligned} \\
 & & \multicolumn{2}{c}{mean $\downarrow$} & \multicolumn{2}{c}{AUC $\uparrow$} & \multicolumn{2}{c}{mean $\downarrow$} & \multicolumn{2}{c}{AUC $\uparrow$} \\
  3D samples: Number      & 3D samples: Perc.                     & $-$  & $+$    & $-$ & $+$ & $-$ & $+$ & $-$ & $+$ \\
\midrule
$1$     &   $(3.4e\minus 3 \%)$ & 1.96 & 1.64 & 0.62 & 0.68 & 34.86 & 18.40 & 0.08 & 0.11   \\
$5$     &   $(0.017 \%)$        & 1.85 & 1.41 & 0.64 & 0.72 & 26.40 & 15.26 & 0.11 & 0.13   \\
$14$    &   $(0.045 \%)$        & 1.78 & 1.39 & 0.65 & 0.73 & 25.24 & 12.98 & 0.11 & 0.13   \\
$27$    &   $(0.094 \%)$        & 1.75 & 1.34 & 0.66 & 0.73 & 23.90 & 11.93 & 0.12 & 0.14   \\
$127$   &   $(0.43 \%)$         & 1.54 & 1.24 & 0.70 & 0.76 & 21.83 & 12.08 & 0.13 & 0.16   \\ 
$499$   &   $(1.7 \%)$          & 1.23 & 1.18 & 0.76 & 0.77 & 11.68 & 10.88 & 0.17 & 0.18   \\ 
$997$   &   $(3.4 \%)$          & 1.14 & 1.12 & 0.77 & 0.78 & 9.85  & 9.42  & 0.18 & 0.19   \\
$1993$  &   $(6.8 \%)$          & 1.10 & 1.07 & 0.78 & 0.79 & 8.83  & 8.75  & 0.19 & 0.20   \\ 
$3810$  &   $(13 \%)$           & 1.06 & 1.04 & 0.79 & 0.79 & 8.01  & 7.90  & 0.21 & 0.21   \\ 
$7327$  &   $(25 \%)$           & 1.02 & 1.01 & 0.80 & 0.80 & 7.91  & 7.84  & 0.21 & 0.21   \\ 
$14653$ &   $(50 \%)$           & 0.99 & 1.00 & 0.80 & 0.80 & 7.46  & 7.56  & 0.22 & 0.22   \\
$29305$ &   $(100 \%)$          & 0.98 & 0.98 & 0.81 & 0.81 & 7.18  & 7.18  & 0.23 & 0.23   \\
\bottomrule
\end{tabularx}
\end{table*}

\section{Qualitative results}
\begin{figure*}[t!p]
    \centering
    \begin{subfigure}[b]{0.13\linewidth}
        \hspace{\linewidth}
    \end{subfigure}
    \begin{subfigure}[b]{0.13\linewidth}
        \hspace{\linewidth}
    \end{subfigure}
    \begin{subfigure}[b]{0.13\linewidth}
    \caption*{\textbf{Front view}}
        \hspace{\linewidth}
    \end{subfigure}
    \begin{subfigure}[b]{0.13\linewidth}
        \hspace{\linewidth}
    \end{subfigure}
    \begin{subfigure}[b]{0.13\linewidth}
        \hspace{\linewidth}
    \end{subfigure}
    \begin{subfigure}[b]{0.13\linewidth}
    \caption*{\textbf{Top view}}
        \hspace{\linewidth}
    \end{subfigure}
    \begin{subfigure}[b]{0.13\linewidth}
        \hspace{\linewidth}
    \end{subfigure}
    \begin{subfigure}[b]{0.13\linewidth}
        \includegraphics[trim={3.6cm 1.8cm 3.6cm 1.8cm},clip,height=0.81\linewidth]{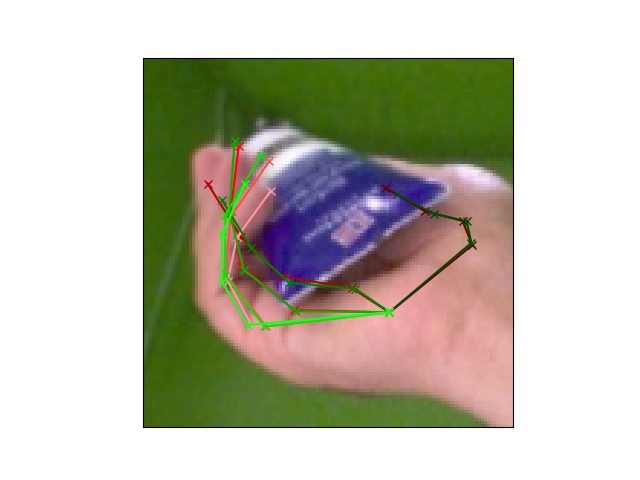}
    \end{subfigure}
    \begin{subfigure}[b]{0.13\linewidth}
        \includegraphics[trim={\cropparamhorz{} \cropparamvert{} \cropparamhorz{} \cropparamvert{}},clip,width=\linewidth]{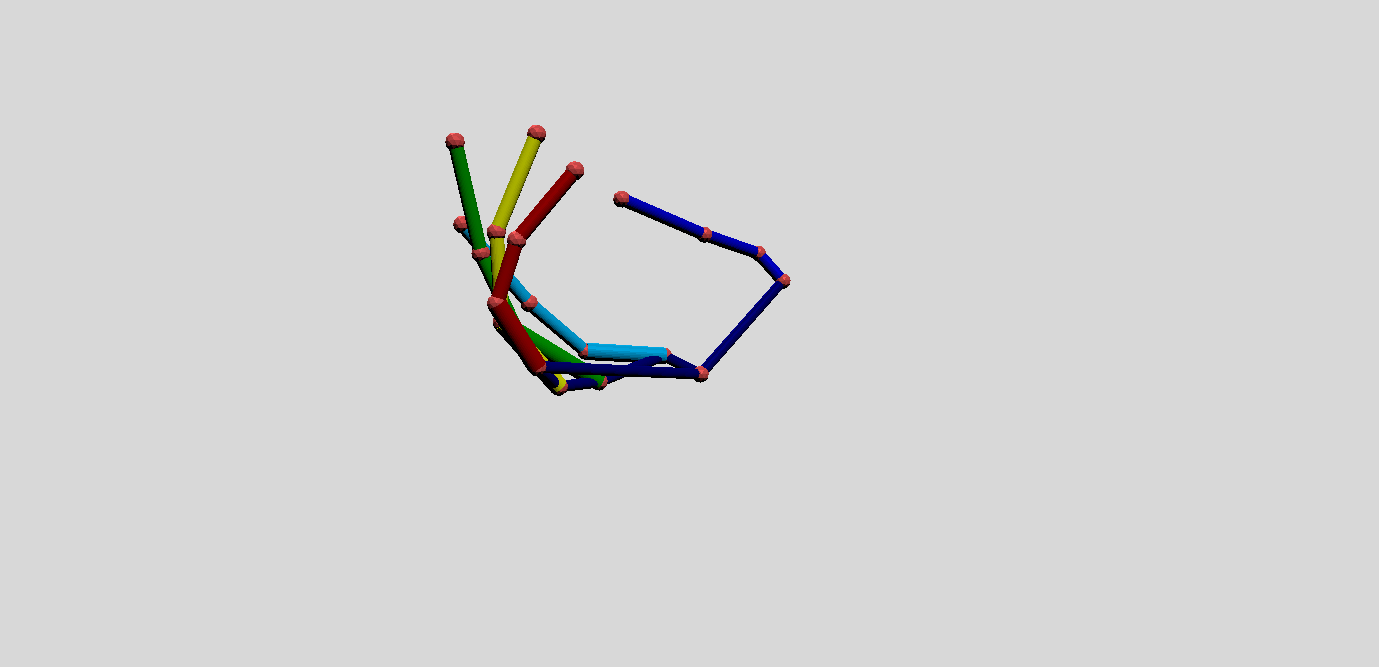}
    \end{subfigure}
    \begin{subfigure}[b]{0.13\linewidth}
        \includegraphics[trim={\cropparamhorz{} \cropparamvert{} \cropparamhorz{} \cropparamvert{}},clip,width=\linewidth]{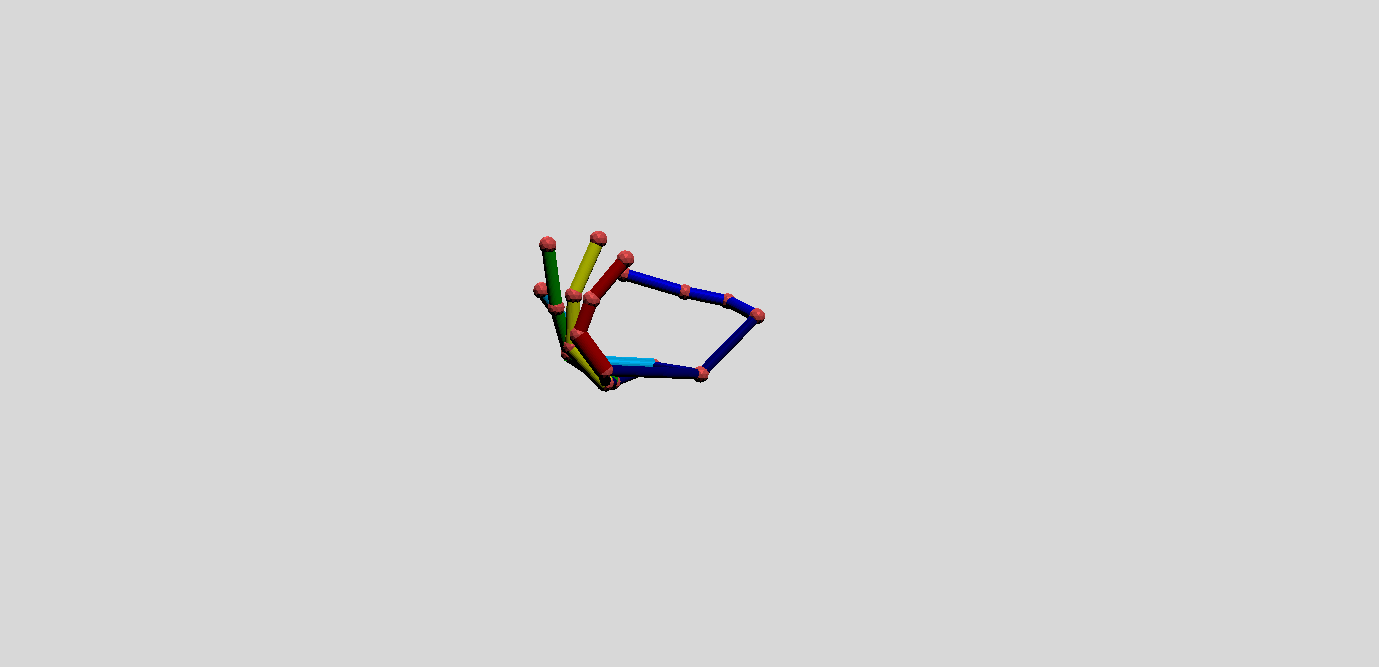}
    \end{subfigure}
    \begin{subfigure}[b]{0.13\linewidth}
        \includegraphics[trim={\cropparamhorz{} \cropparamvert{} \cropparamhorz{} \cropparamvert{}},clip,width=\linewidth]{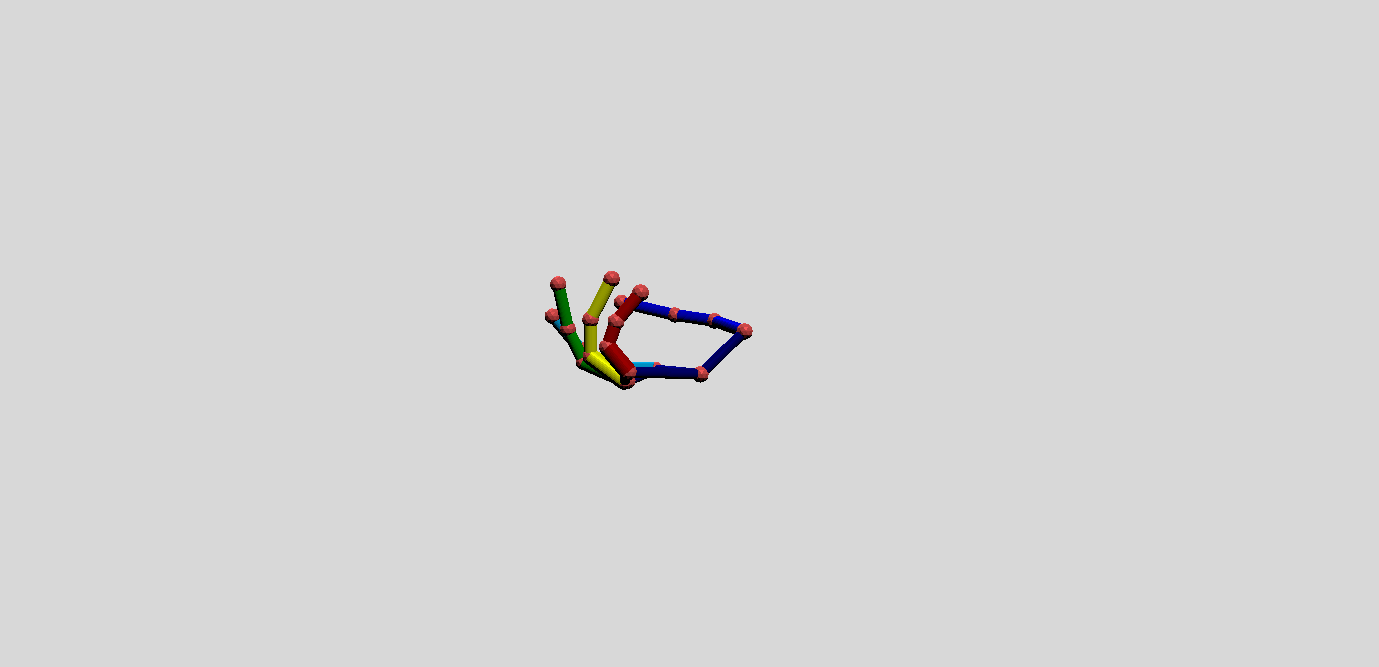}
    \end{subfigure}
    \ \ \ %
    \begin{subfigure}[b]{0.13\linewidth}
        \includegraphics[trim={\cropparamhorz{} \cropparamvert{} \cropparamhorz{} \cropparamvert{}},clip,width=\linewidth]{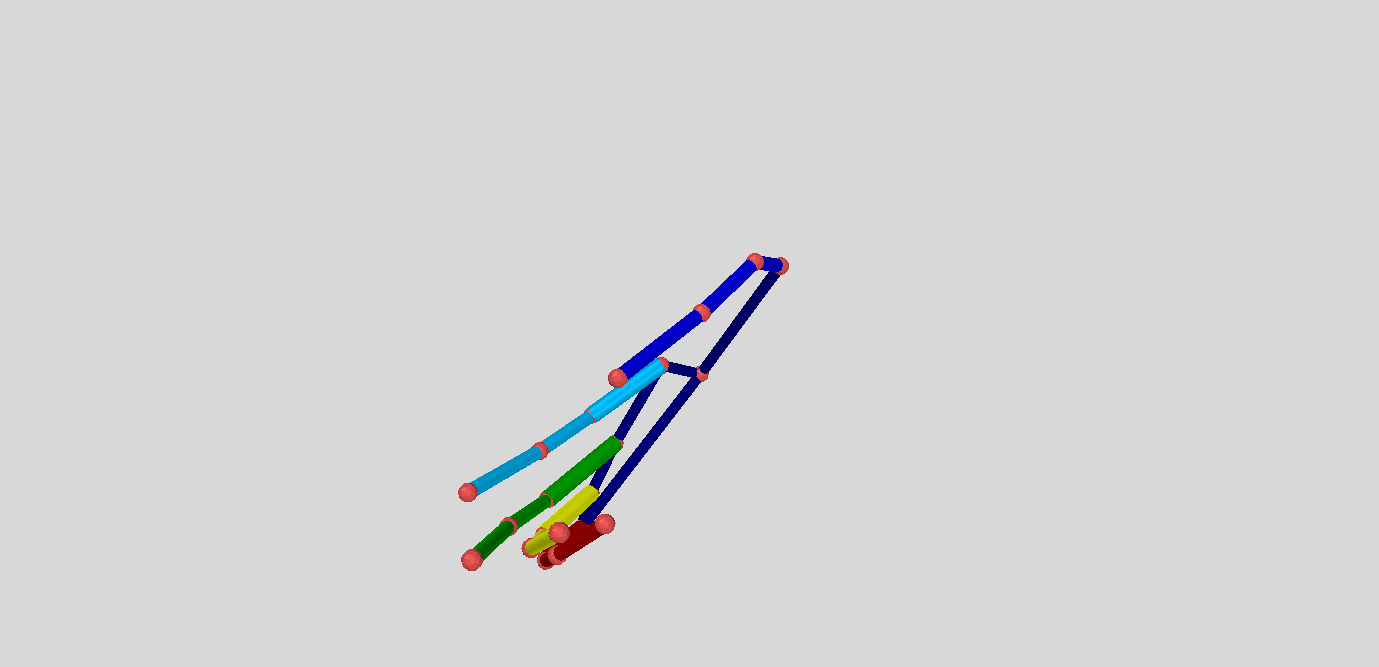}
    \end{subfigure}
    \begin{subfigure}[b]{0.13\linewidth}
        \includegraphics[trim={\cropparamhorz{} \cropparamvert{} \cropparamhorz{} \cropparamvert{}},clip,width=\linewidth]{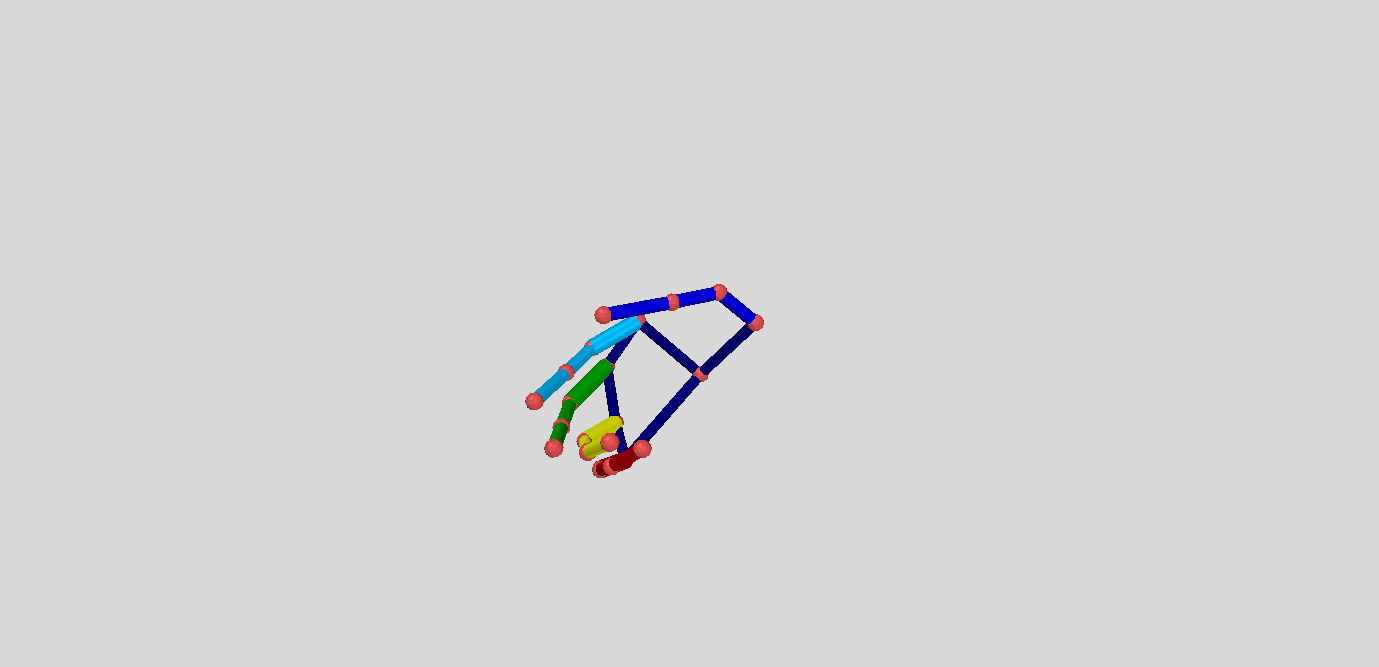}
    \end{subfigure}
    \begin{subfigure}[b]{0.13\linewidth}
        \includegraphics[trim={\cropparamhorz{} \cropparamvert{} \cropparamhorz{} \cropparamvert{}},clip,width=\linewidth]{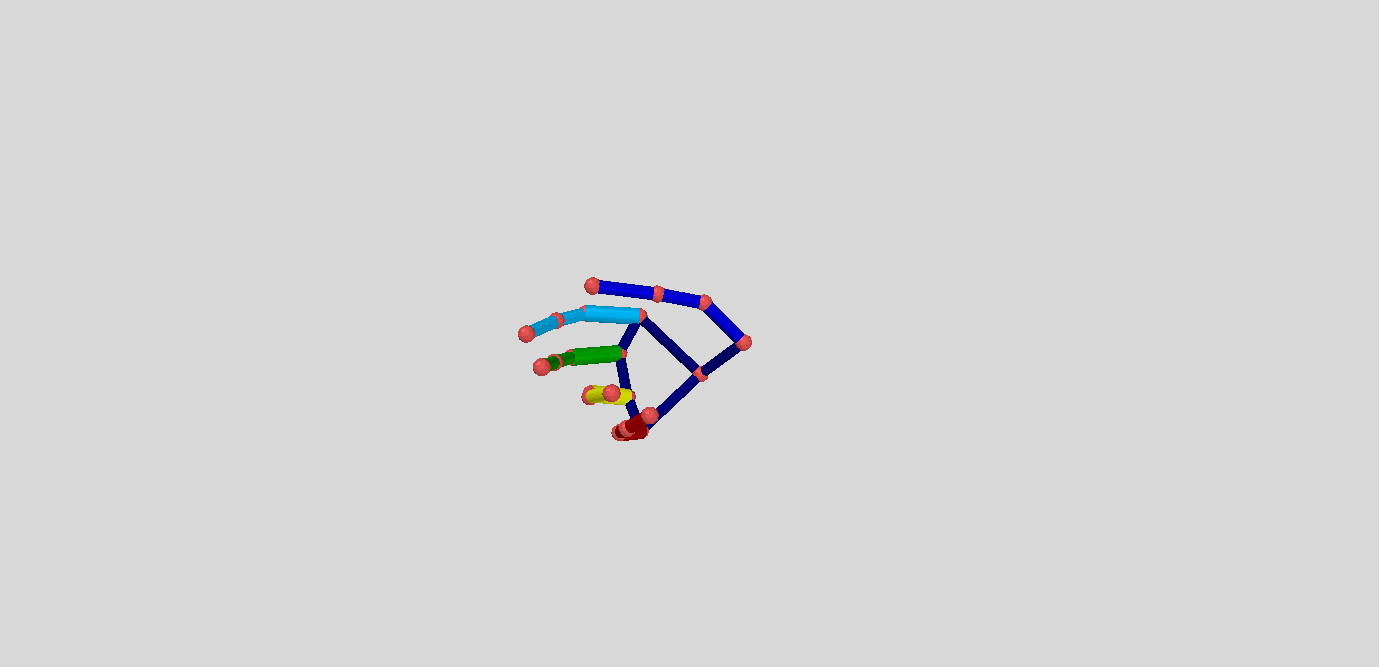}
    \end{subfigure}
    
    \begin{subfigure}[b]{0.13\linewidth}
        \includegraphics[trim={3.6cm 1.8cm 3.6cm 1.8cm},clip,height=0.81\linewidth]{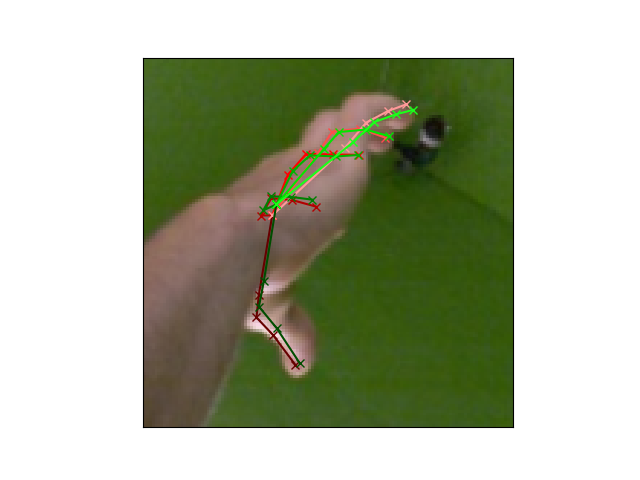}
    \end{subfigure}
    \begin{subfigure}[b]{0.13\linewidth}
        \includegraphics[trim={\cropparamhorz{} \cropparamvert{} \cropparamhorz{} \cropparamvert{}},clip,width=\linewidth]{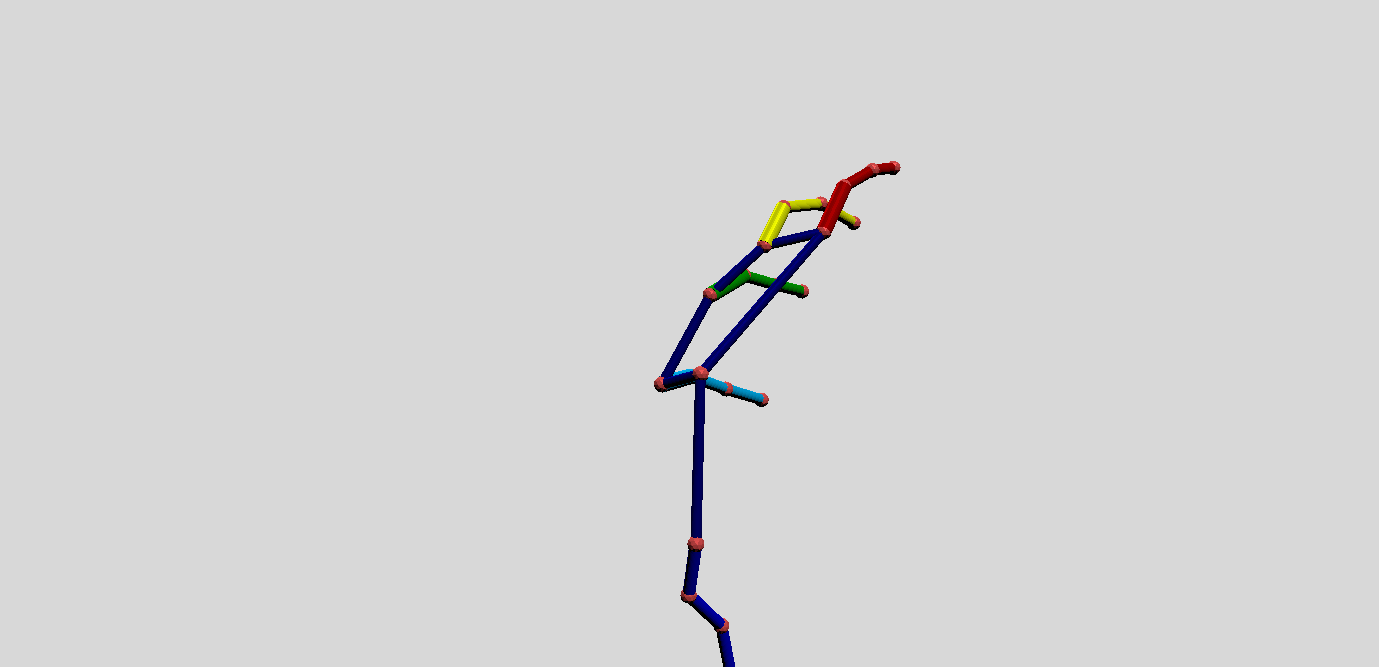}
    \end{subfigure}
    \begin{subfigure}[b]{0.13\linewidth}
        \includegraphics[trim={\cropparamhorz{} \cropparamvert{} \cropparamhorz{} \cropparamvert{}},clip,width=\linewidth]{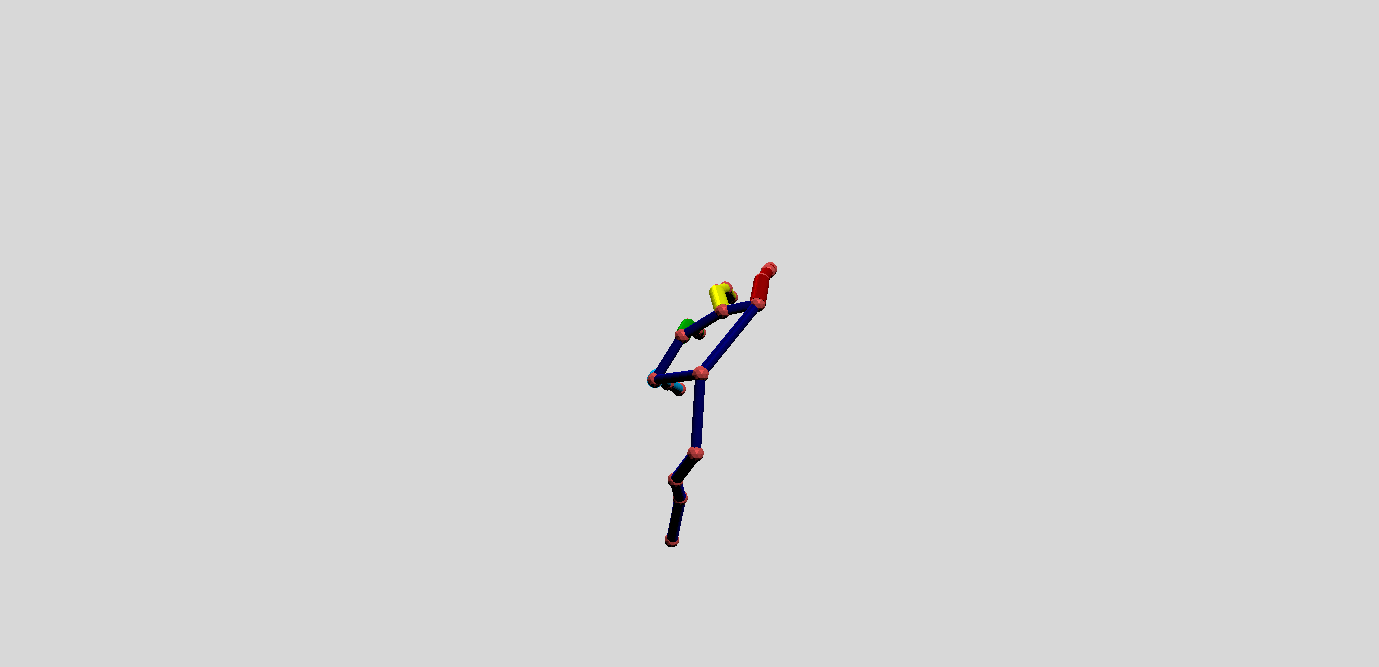}
    \end{subfigure}
    \begin{subfigure}[b]{0.13\linewidth}
        \includegraphics[trim={\cropparamhorz{} \cropparamvert{} \cropparamhorz{} \cropparamvert{}},clip,width=\linewidth]{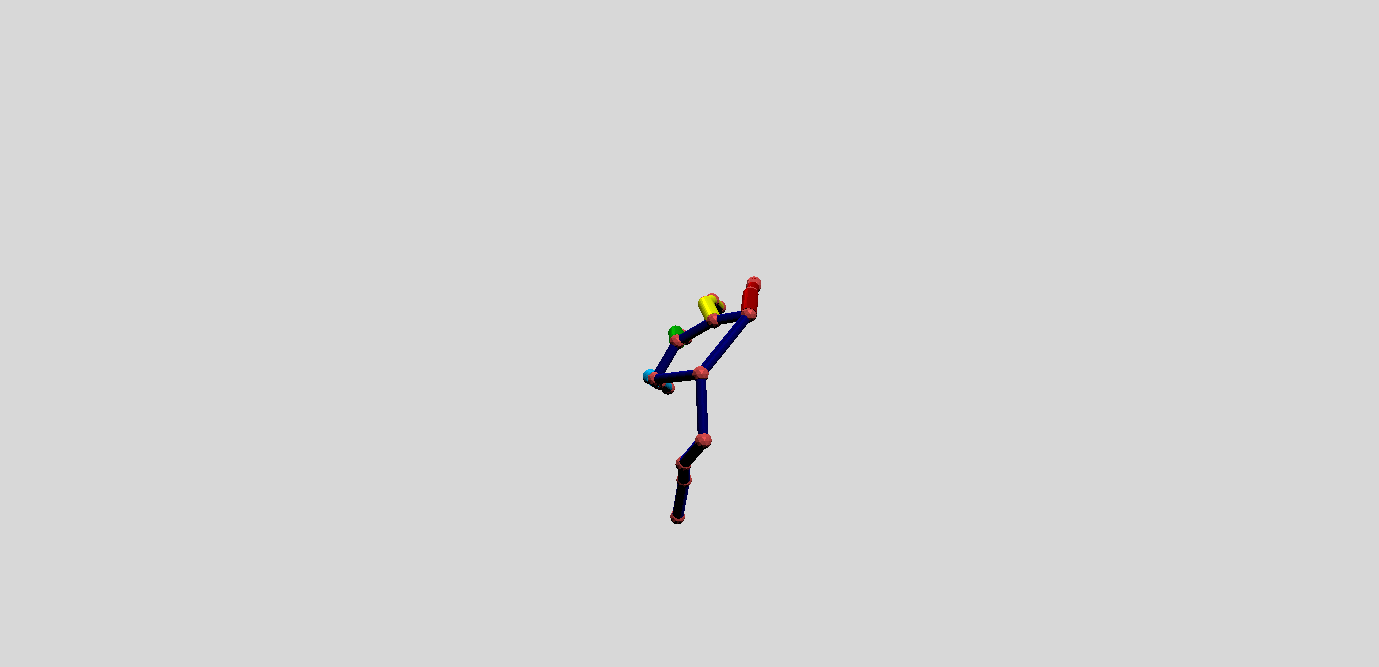}
    \end{subfigure}
    \ \ \ %
    \begin{subfigure}[b]{0.13\linewidth}
        \includegraphics[trim={\cropparamhorz{} \cropparamvert{} \cropparamhorz{} \cropparamvert{}},clip,width=\linewidth]{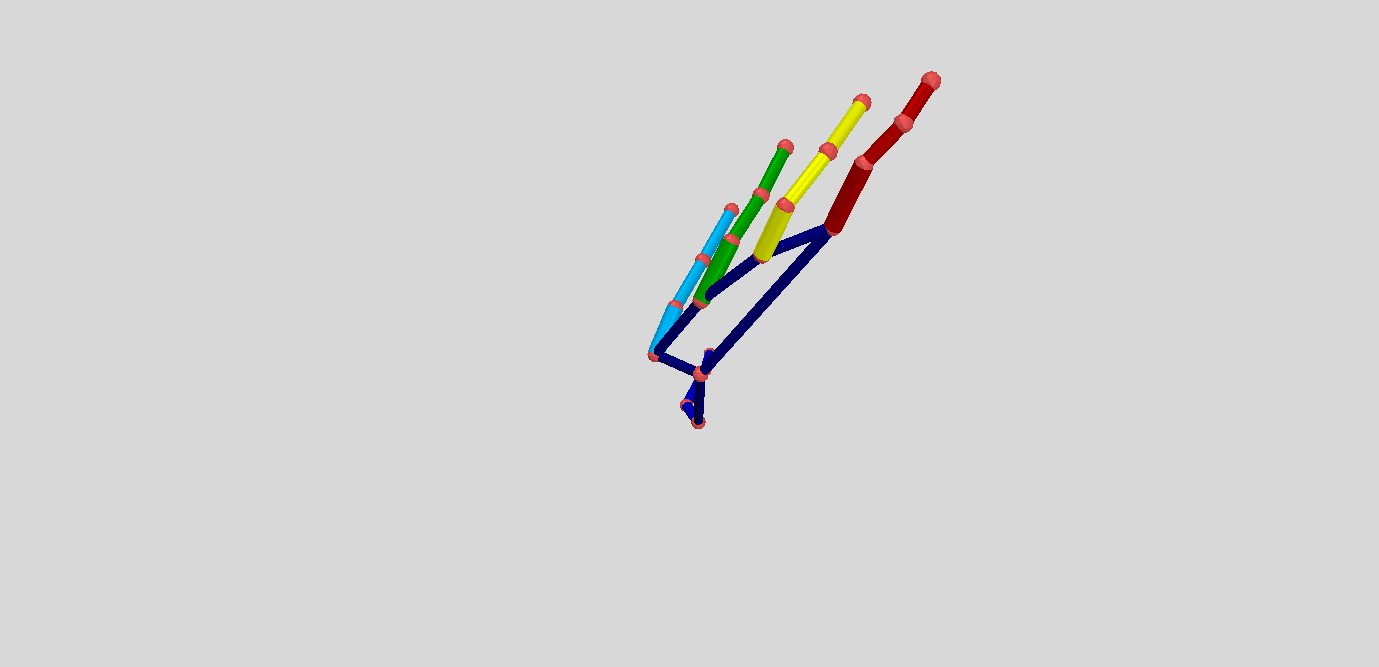}
    \end{subfigure}
    \begin{subfigure}[b]{0.13\linewidth}
        \includegraphics[trim={\cropparamhorz{} \cropparamvert{} \cropparamhorz{} \cropparamvert{}},clip,width=\linewidth]{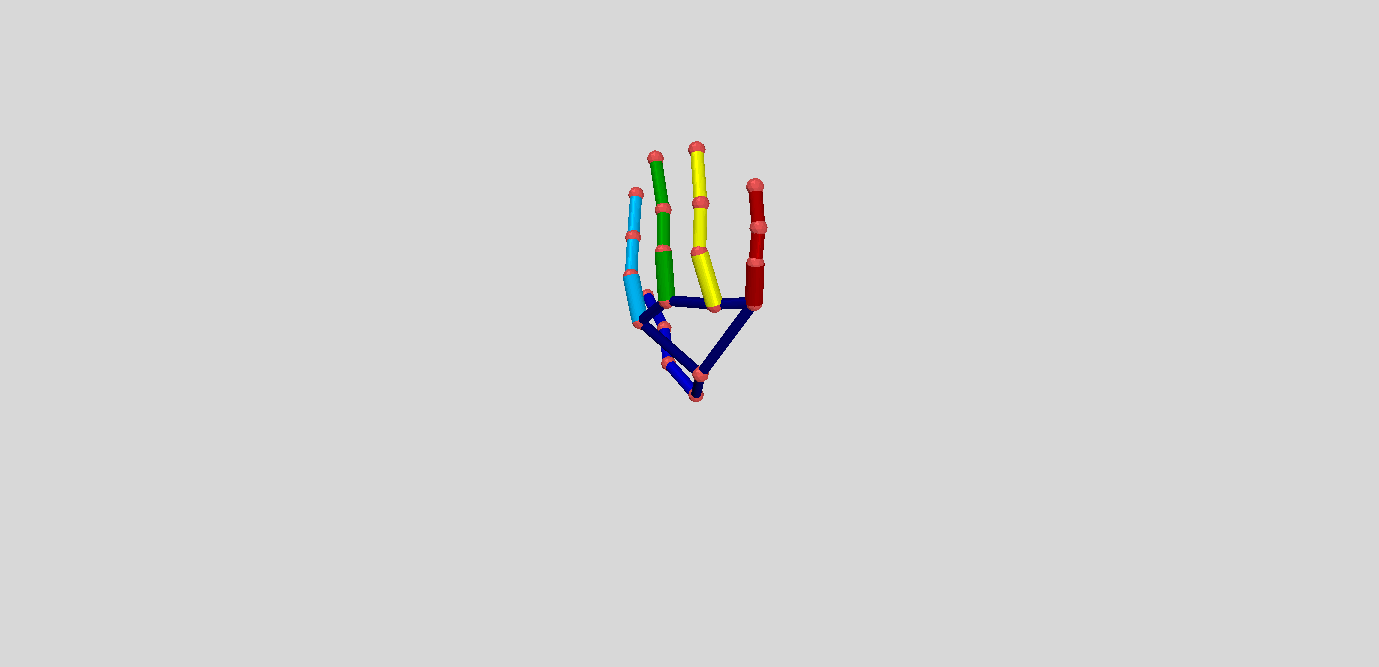}
    \end{subfigure}
    \begin{subfigure}[b]{0.13\linewidth}
        \includegraphics[trim={\cropparamhorz{} \cropparamvert{} \cropparamhorz{} \cropparamvert{}},clip,width=\linewidth]{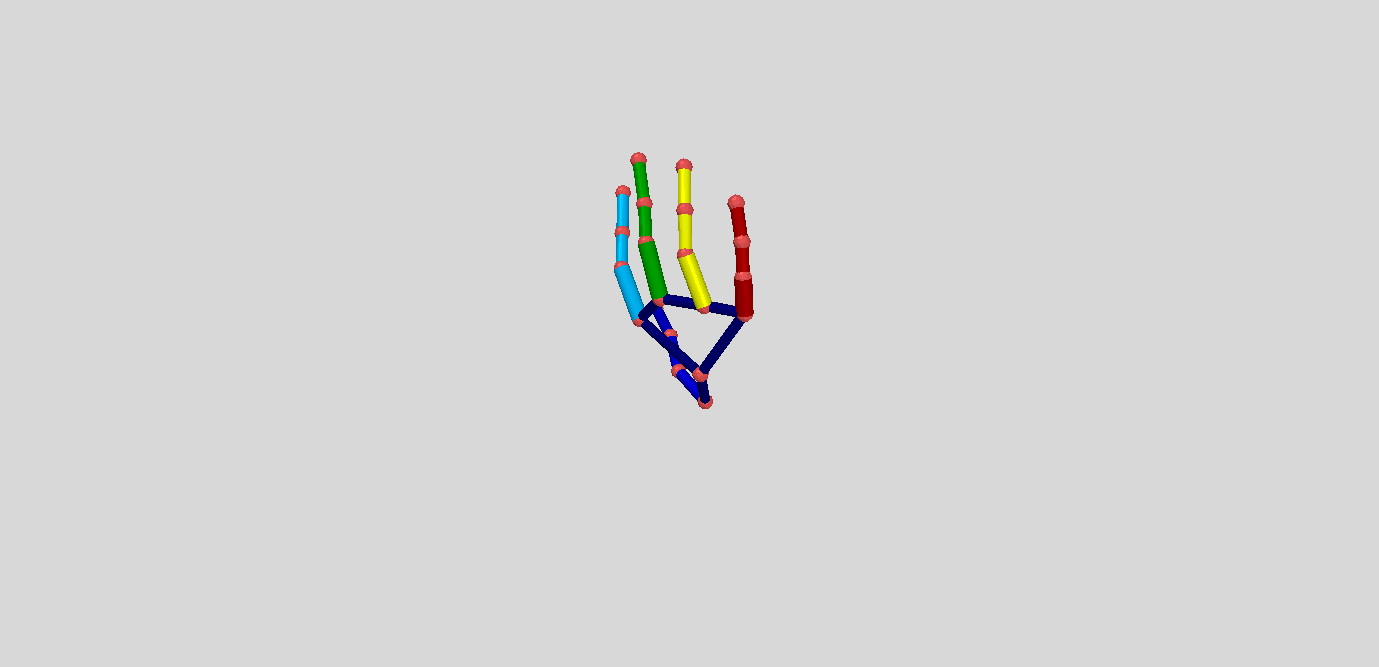}
    \end{subfigure}
    
    \begin{subfigure}[b]{0.13\linewidth}
        \includegraphics[trim={3.6cm 1.8cm 3.6cm 1.8cm},clip,height=0.81\linewidth]{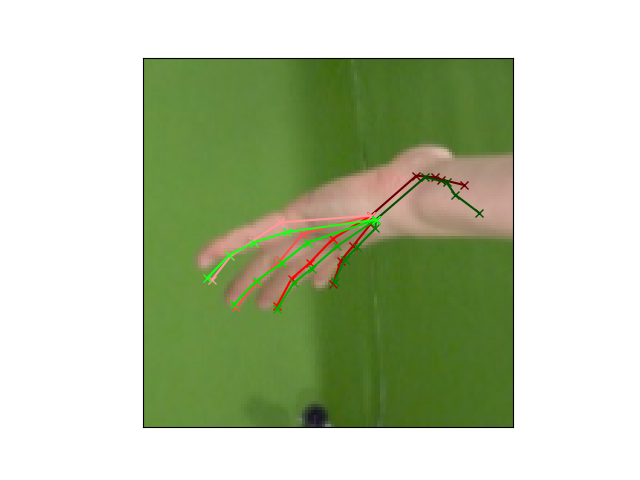}
    \end{subfigure}
    \begin{subfigure}[b]{0.13\linewidth}
        \includegraphics[trim={\cropparamhorz{} \cropparamvert{} \cropparamhorz{} \cropparamvert{}},clip,width=\linewidth]{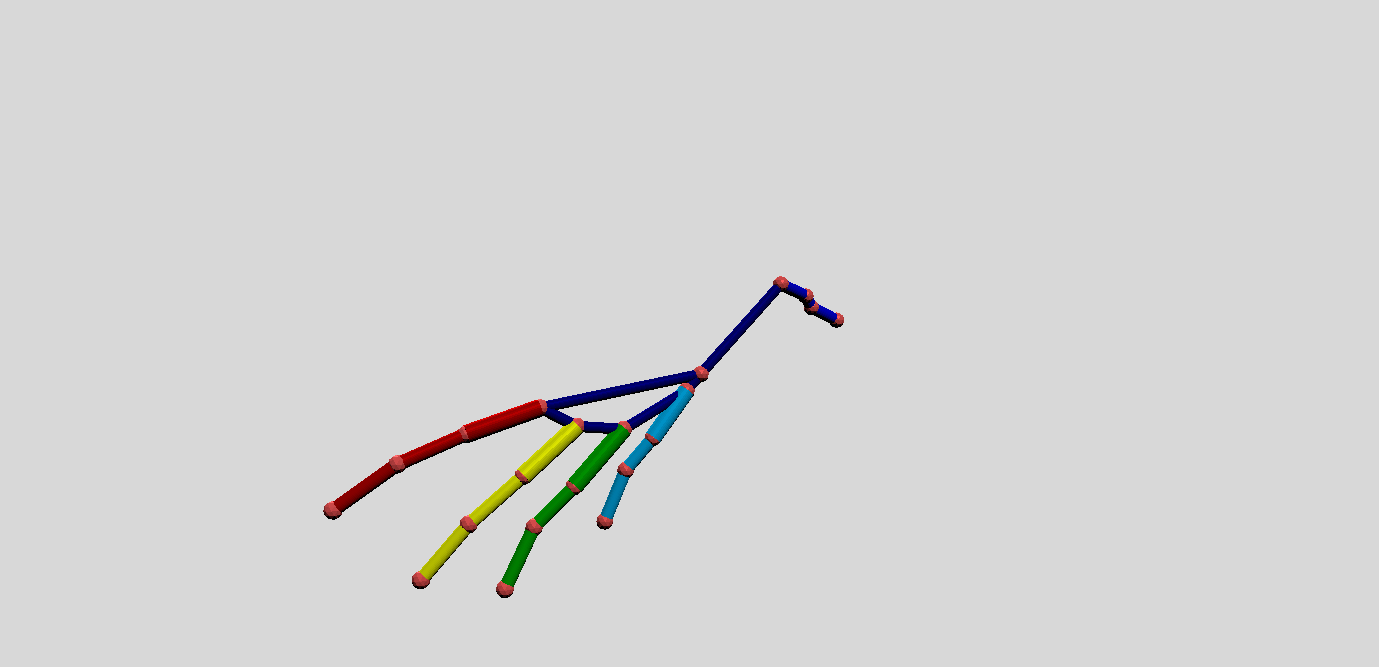}
    \end{subfigure}
    \begin{subfigure}[b]{0.13\linewidth}
        \includegraphics[trim={\cropparamhorz{} \cropparamvert{} \cropparamhorz{} \cropparamvert{}},clip,width=\linewidth]{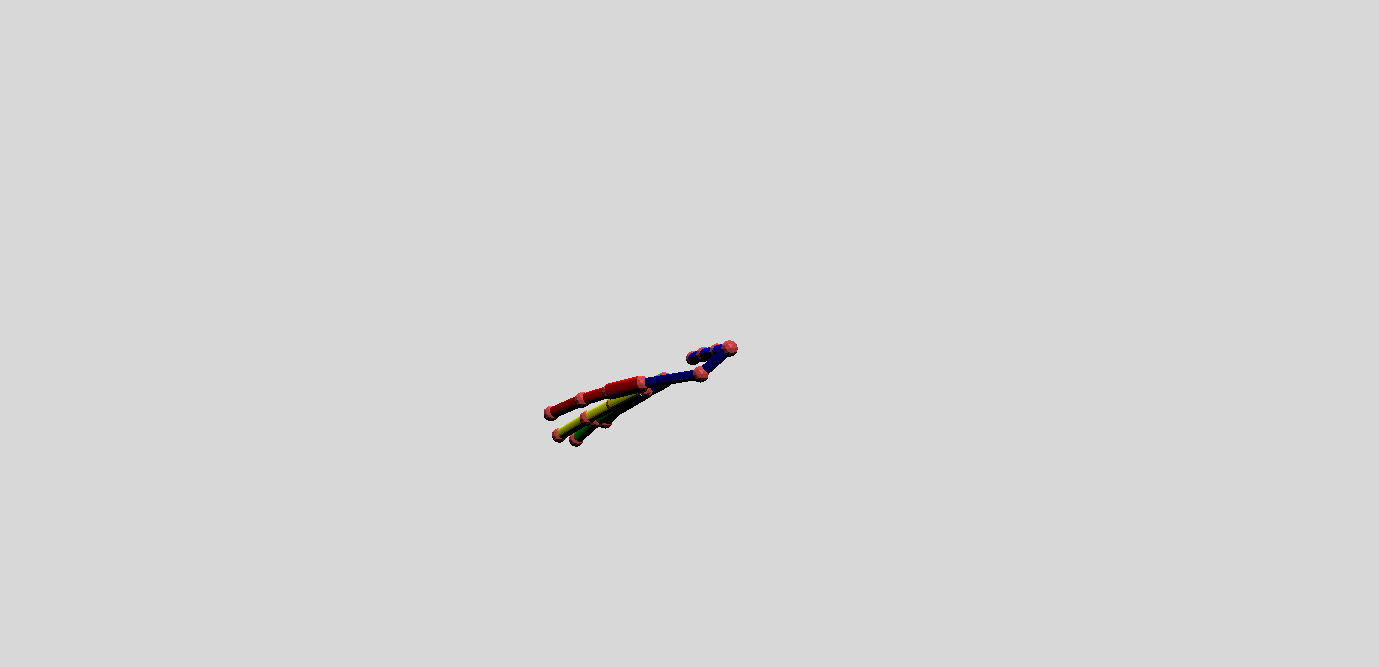}
    \end{subfigure}
    \begin{subfigure}[b]{0.13\linewidth}
        \includegraphics[trim={\cropparamhorz{} \cropparamvert{} \cropparamhorz{} \cropparamvert{}},clip,width=\linewidth]{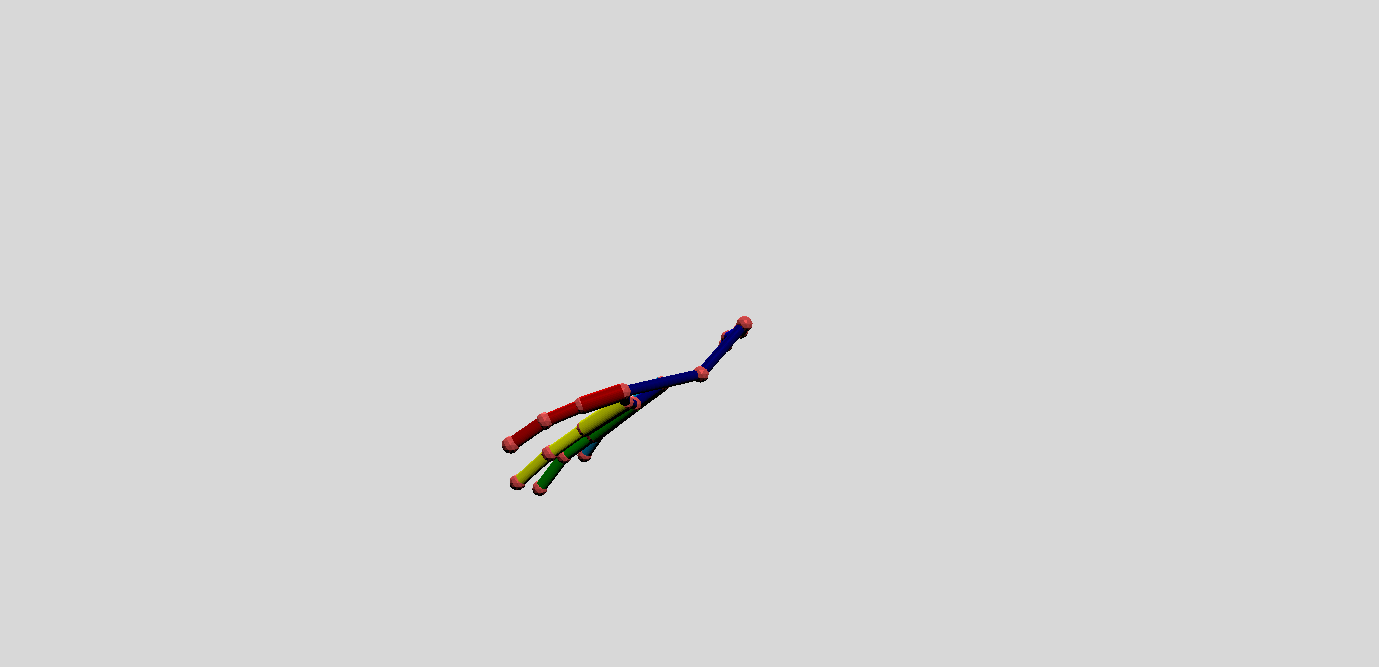}
    \end{subfigure}
    \ \ \ %
    \begin{subfigure}[b]{0.13\linewidth}
        \includegraphics[trim={\cropparamhorz{} \cropparamvert{} \cropparamhorz{} \cropparamvert{}},clip,width=\linewidth]{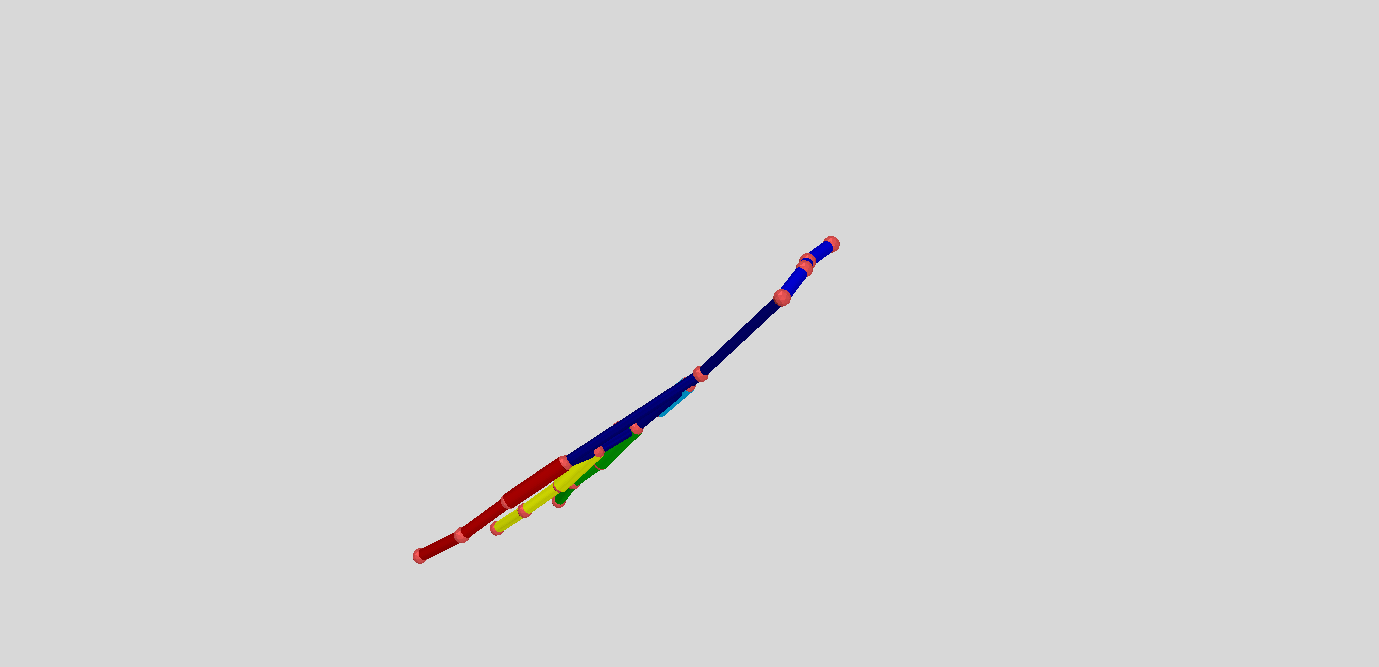}
    \end{subfigure}
    \begin{subfigure}[b]{0.13\linewidth}
        \includegraphics[trim={\cropparamhorz{} \cropparamvert{} \cropparamhorz{} \cropparamvert{}},clip,width=\linewidth]{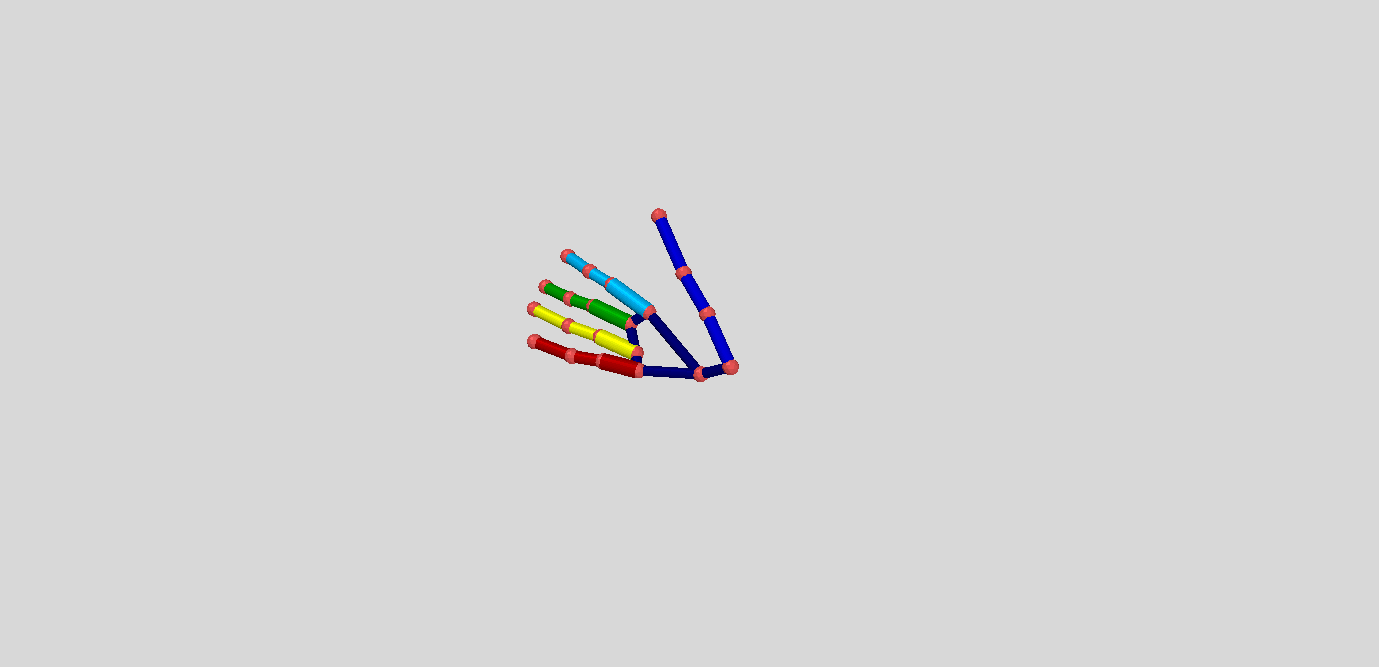}
    \end{subfigure}
    \begin{subfigure}[b]{0.13\linewidth}
        \includegraphics[trim={\cropparamhorz{} \cropparamvert{} \cropparamhorz{} \cropparamvert{}},clip,width=\linewidth]{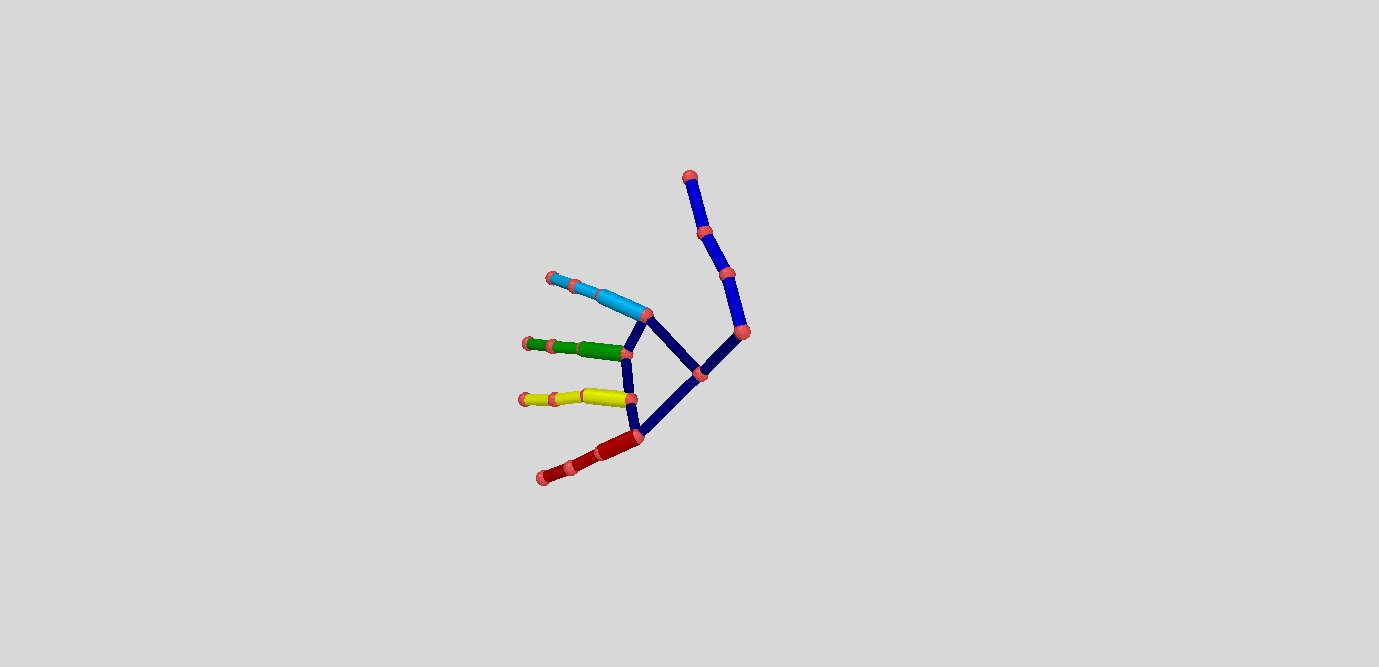}
    \end{subfigure}
    
    \begin{subfigure}[b]{0.13\linewidth}
        \includegraphics[trim={3.6cm 1.8cm 3.6cm 1.8cm},clip,height=0.81\linewidth]{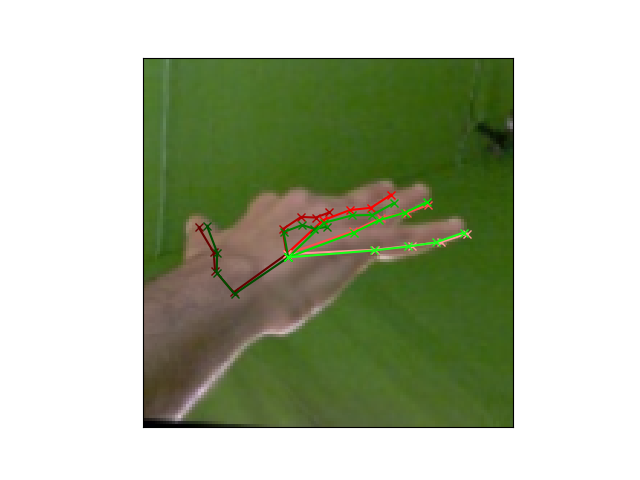}
    \end{subfigure}
    \begin{subfigure}[b]{0.13\linewidth}
        \includegraphics[trim={\cropparamhorz{} \cropparamvert{} \cropparamhorz{} \cropparamvert{}},clip,width=\linewidth]{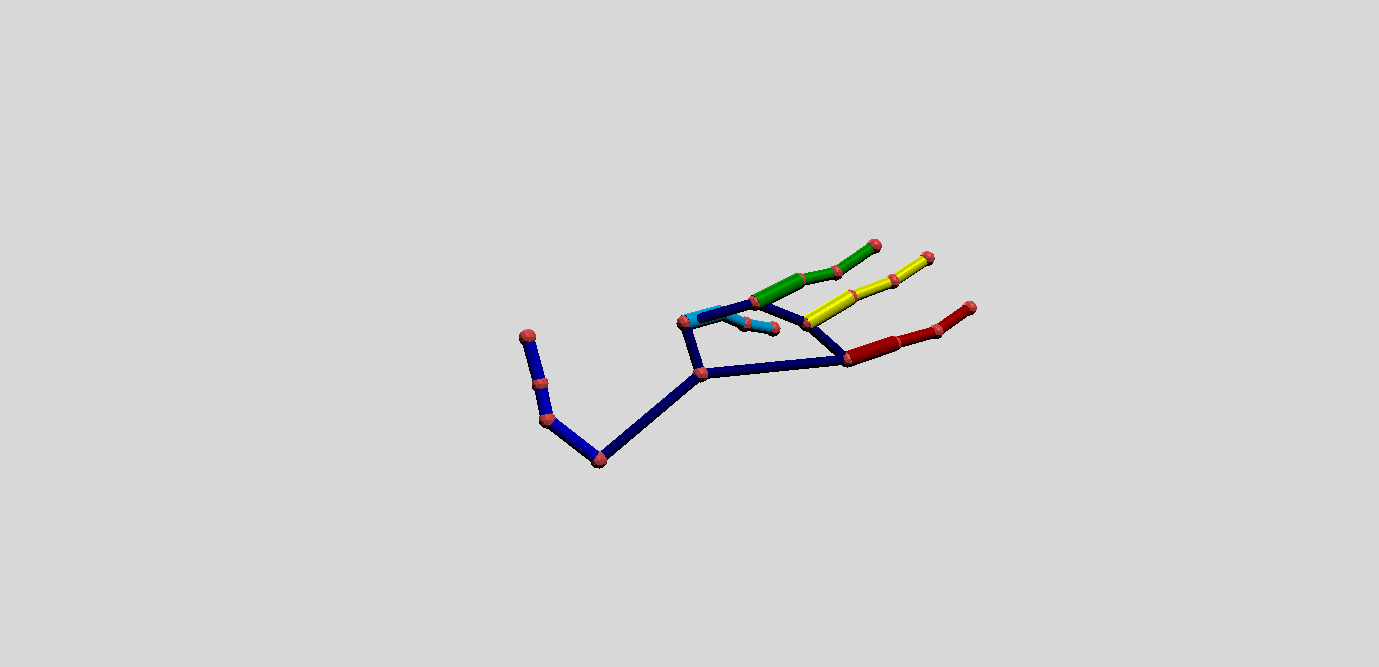}
    \end{subfigure}
    \begin{subfigure}[b]{0.13\linewidth}
        \includegraphics[trim={\cropparamhorz{} \cropparamvert{} \cropparamhorz{} \cropparamvert{}},clip,width=\linewidth]{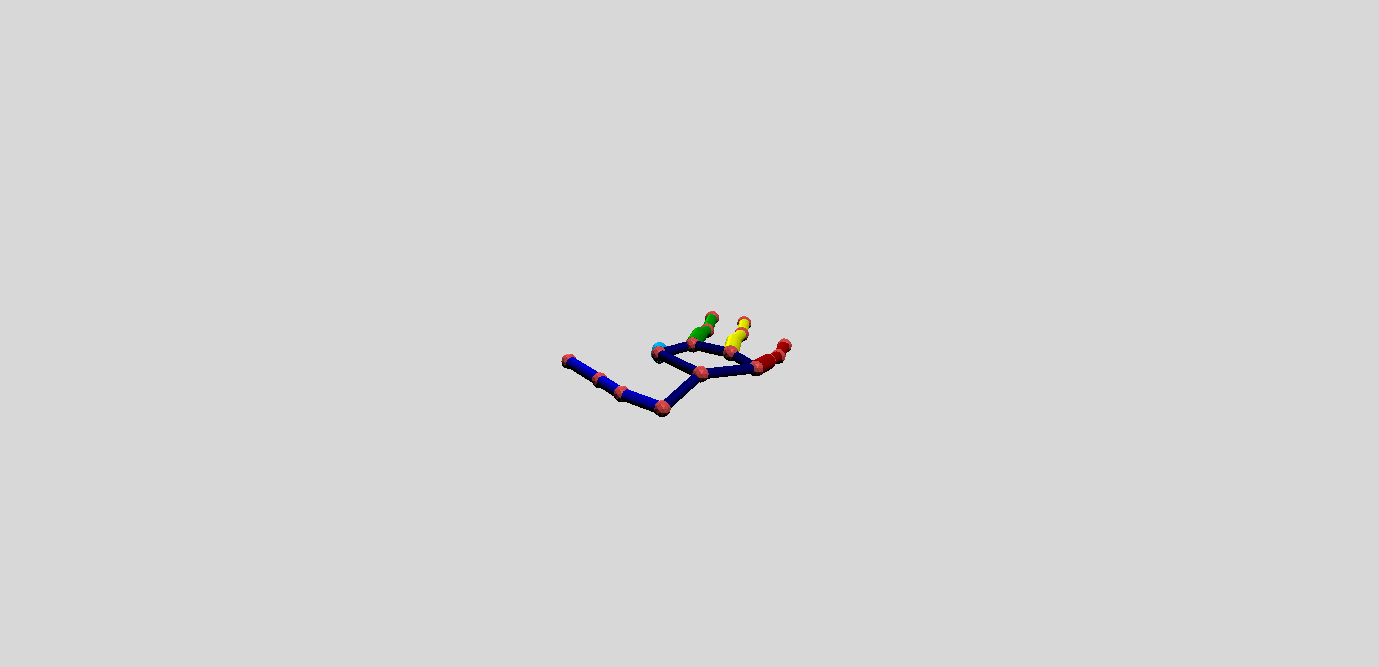}
    \end{subfigure}
    \begin{subfigure}[b]{0.13\linewidth}
        \includegraphics[trim={\cropparamhorz{} \cropparamvert{} \cropparamhorz{} \cropparamvert{}},clip,width=\linewidth]{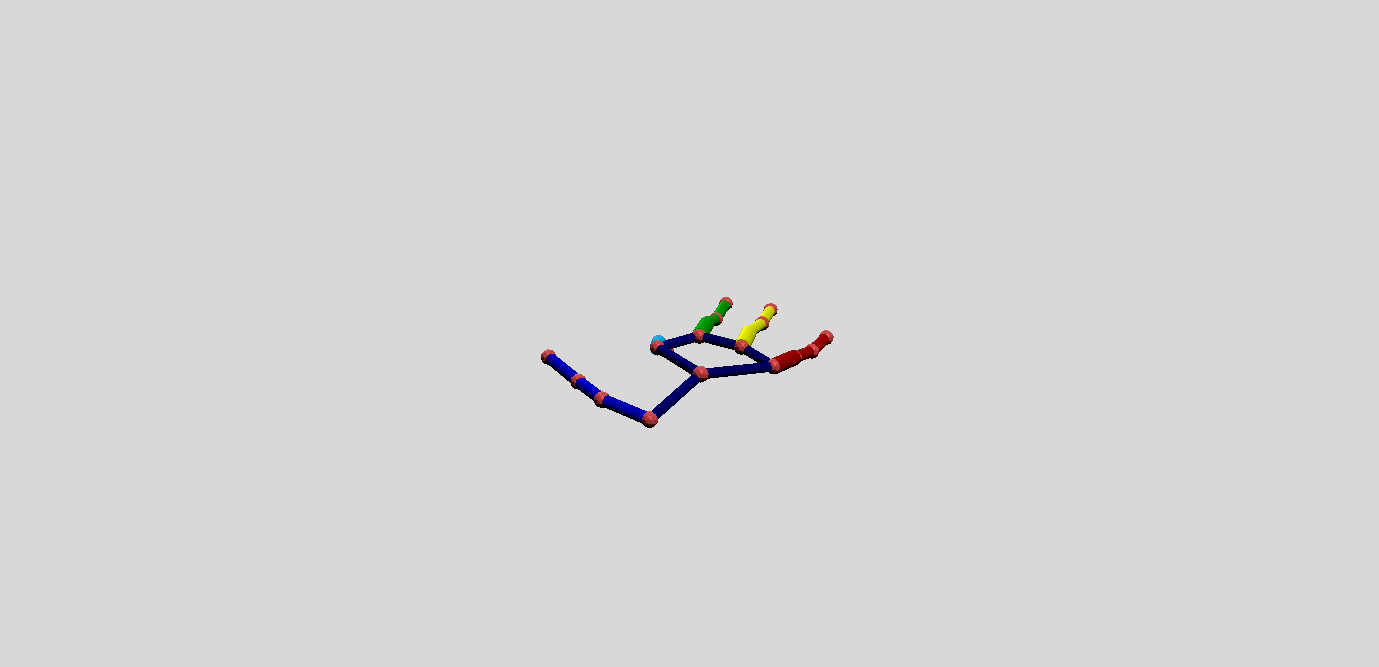}
    \end{subfigure}
    \ \ \ %
    \begin{subfigure}[b]{0.13\linewidth}
        \includegraphics[trim={\cropparamhorz{} \cropparamvert{} \cropparamhorz{} \cropparamvert{}},clip,width=\linewidth]{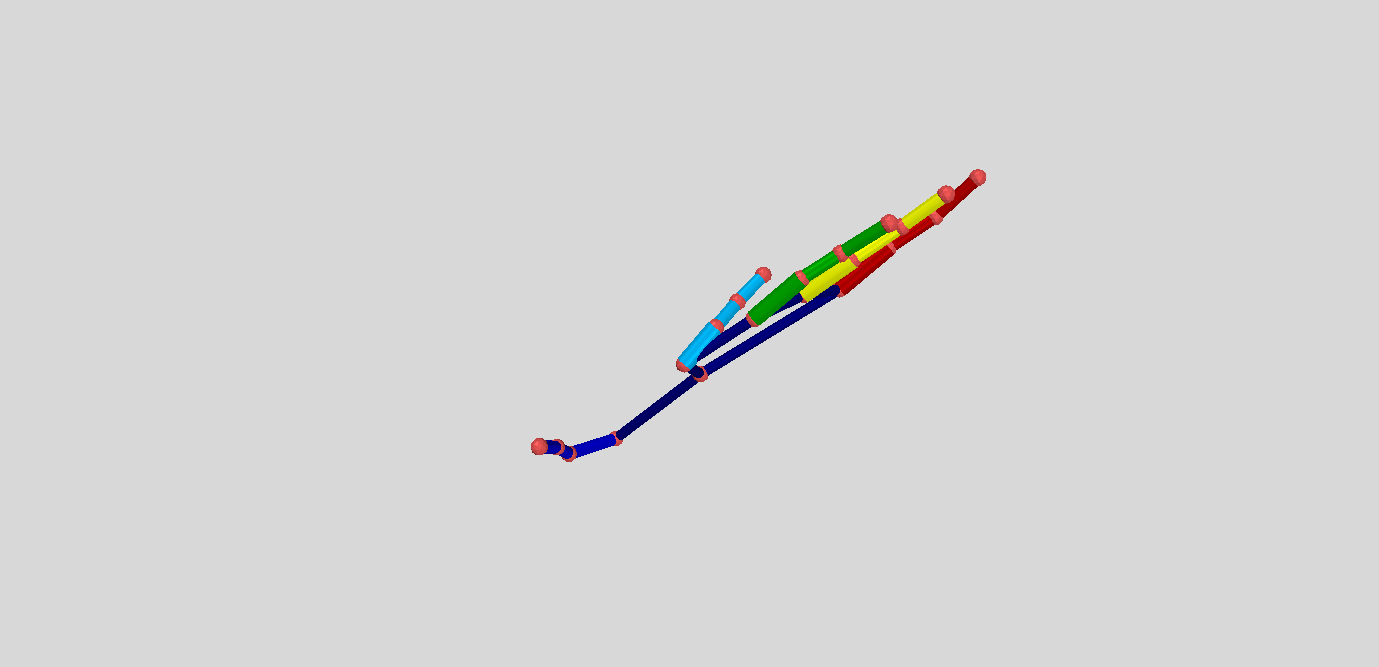}
    \end{subfigure}
    \begin{subfigure}[b]{0.13\linewidth}
        \includegraphics[trim={\cropparamhorz{} \cropparamvert{} \cropparamhorz{} \cropparamvert{}},clip,width=\linewidth]{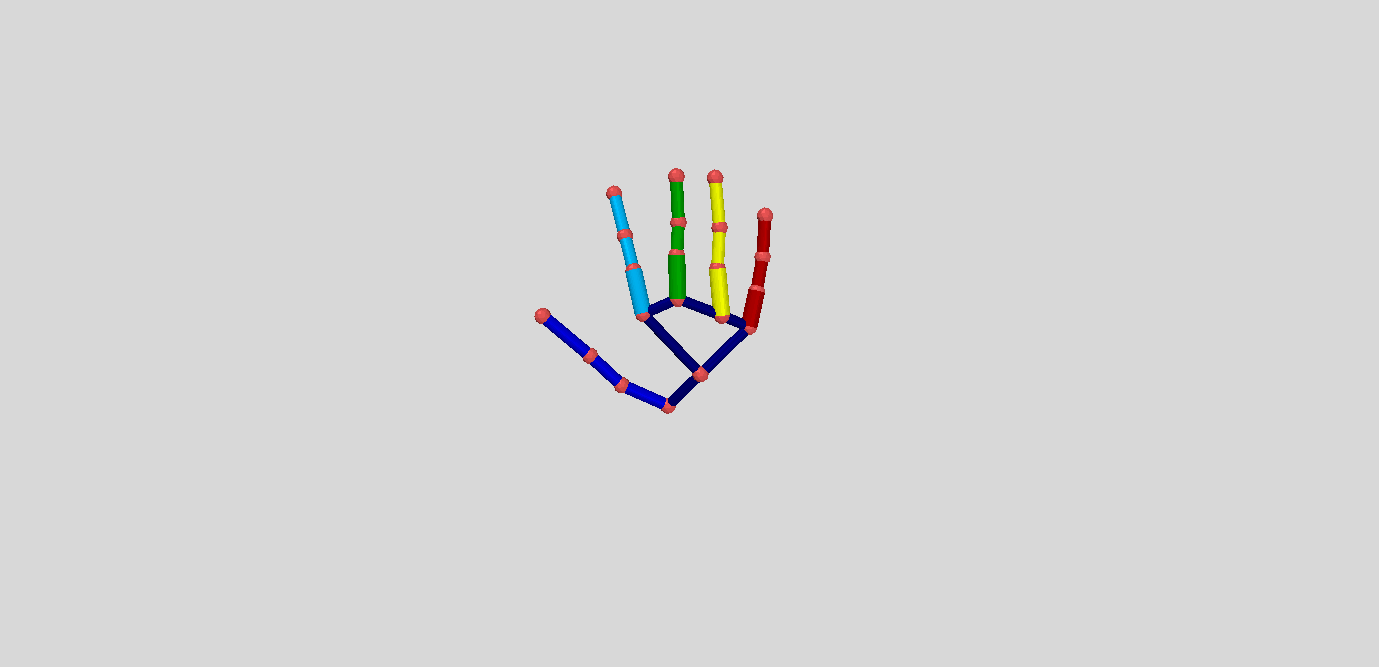}
    \end{subfigure}
    \begin{subfigure}[b]{0.13\linewidth}
        \includegraphics[trim={\cropparamhorz{} \cropparamvert{} \cropparamhorz{} \cropparamvert{}},clip,width=\linewidth]{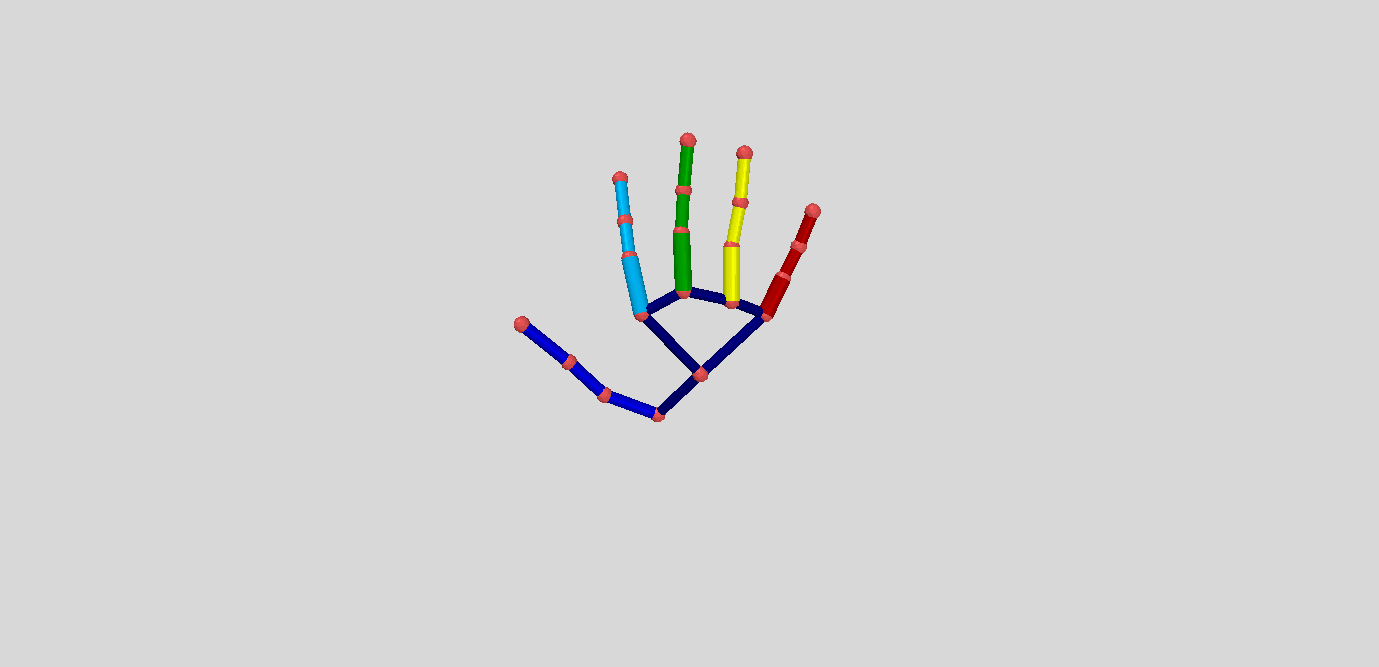}
    \end{subfigure}
    
    \begin{subfigure}[b]{0.13\linewidth}
        \includegraphics[trim={3.6cm 1.8cm 3.6cm 1.8cm},clip,height=0.81\linewidth]{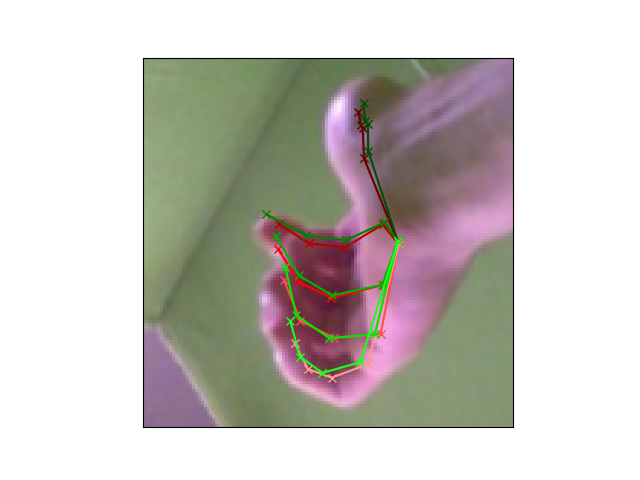}
    \end{subfigure}
    \begin{subfigure}[b]{0.13\linewidth}
        \includegraphics[trim={\cropparamhorz{} \cropparamvert{} \cropparamhorz{} \cropparamvert{}},clip,width=\linewidth]{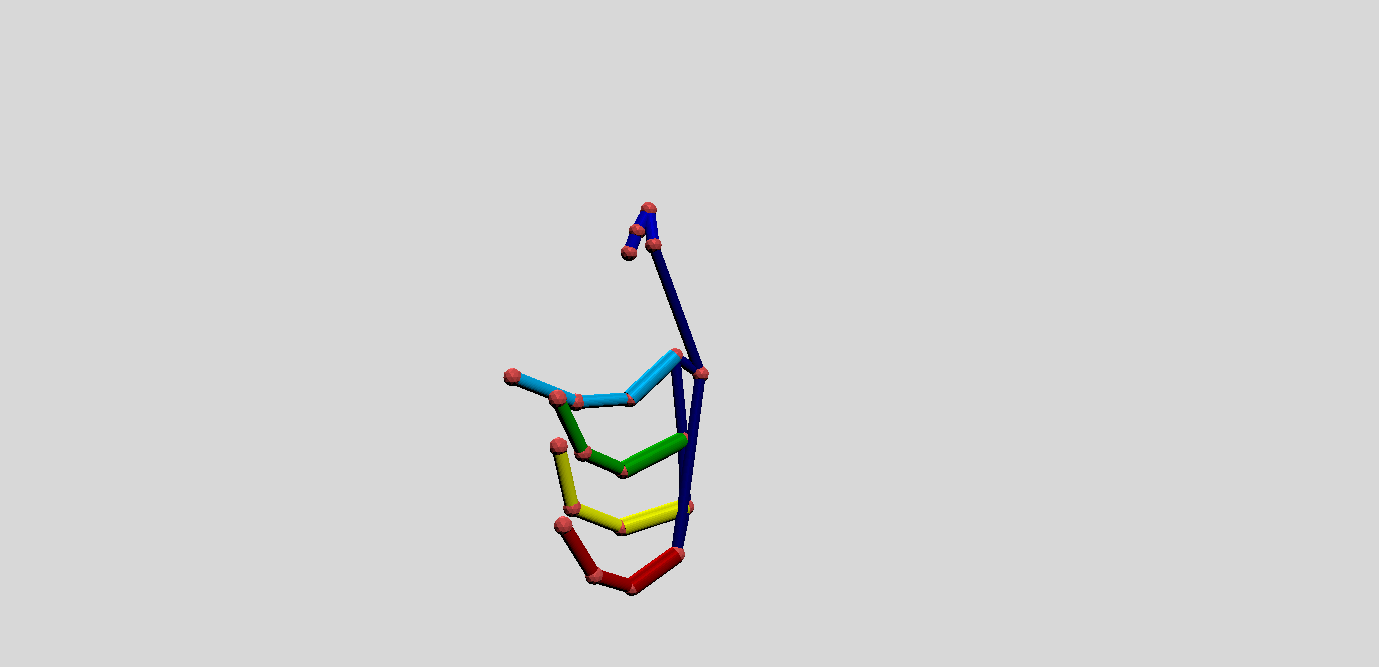}
    \end{subfigure}
    \begin{subfigure}[b]{0.13\linewidth}
        \includegraphics[trim={\cropparamhorz{} \cropparamvert{} \cropparamhorz{} \cropparamvert{}},clip,width=\linewidth]{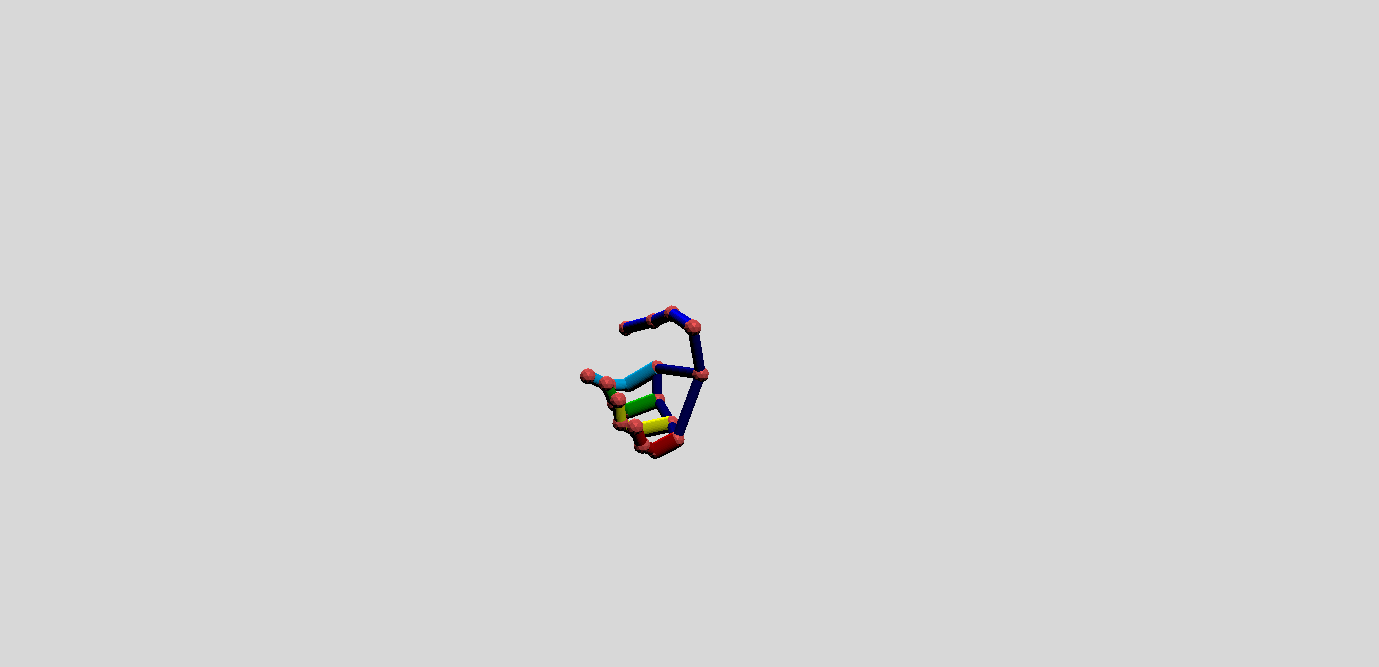}
    \end{subfigure}
    \begin{subfigure}[b]{0.13\linewidth}
        \includegraphics[trim={\cropparamhorz{} \cropparamvert{} \cropparamhorz{} \cropparamvert{}},clip,width=\linewidth]{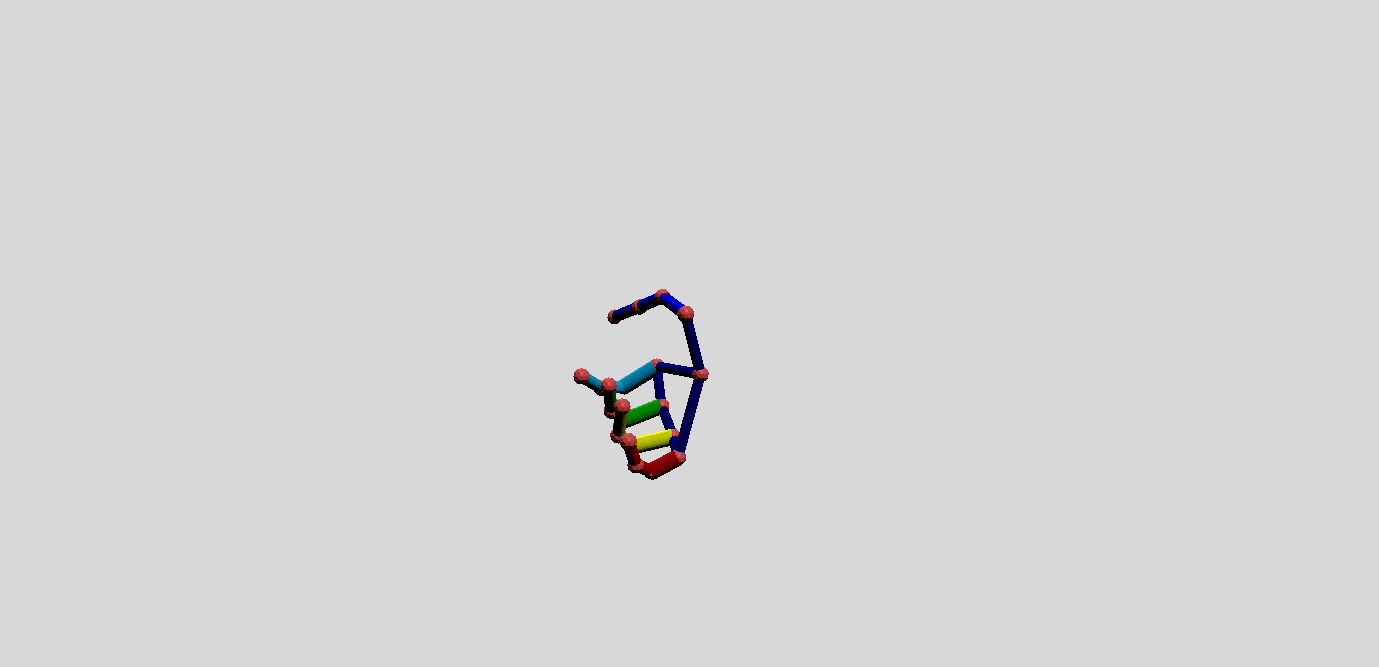}
    \end{subfigure}
    \ \ \ %
    \begin{subfigure}[b]{0.13\linewidth}
        \includegraphics[trim={\cropparamhorz{} \cropparamvert{} \cropparamhorz{} \cropparamvert{}},clip,width=\linewidth]{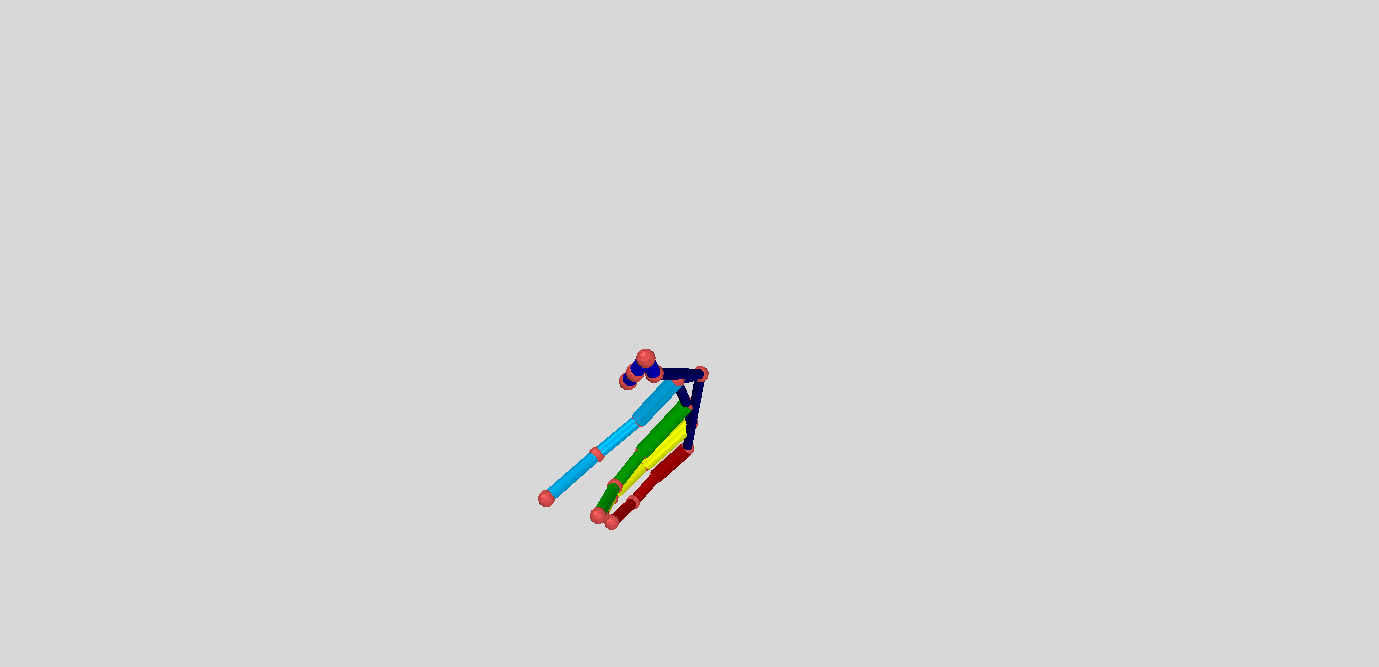}
    \end{subfigure}
    \begin{subfigure}[b]{0.13\linewidth}
        \includegraphics[trim={\cropparamhorz{} \cropparamvert{} \cropparamhorz{} \cropparamvert{}},clip,width=\linewidth]{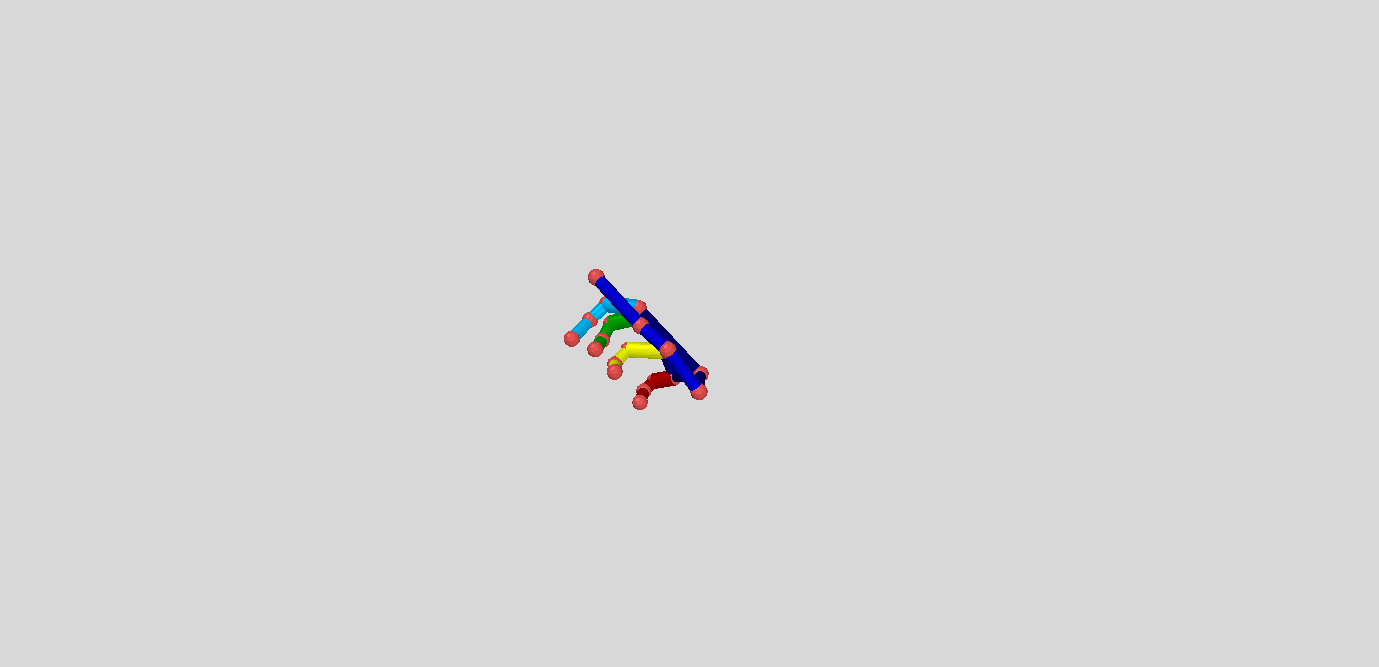}
    \end{subfigure}
    \begin{subfigure}[b]{0.13\linewidth}
        \includegraphics[trim={\cropparamhorz{} \cropparamvert{} \cropparamhorz{} \cropparamvert{}},clip,width=\linewidth]{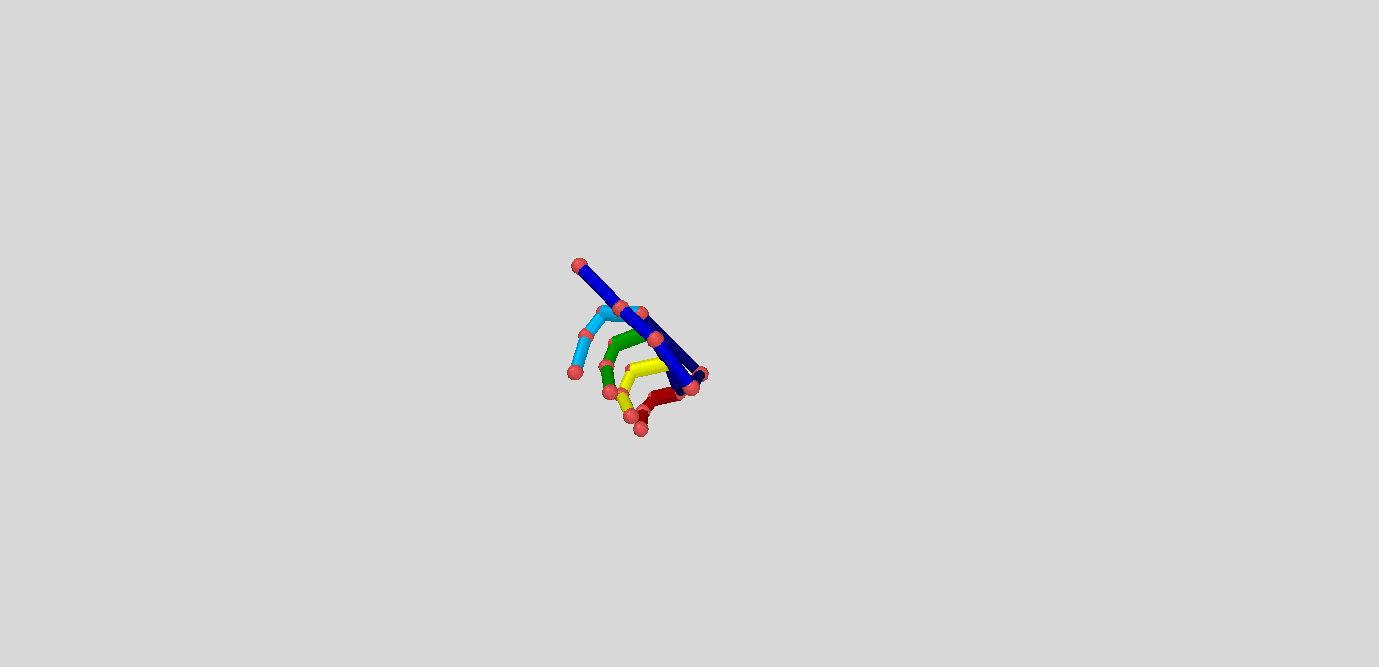}
    \end{subfigure}

    \begin{subfigure}[b]{0.13\linewidth}
        \includegraphics[trim={3.6cm 1.8cm 3.6cm 1.8cm},clip,height=0.81\linewidth]{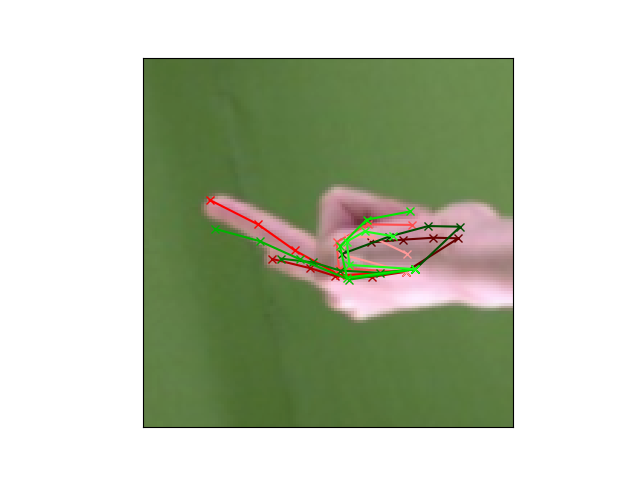}
        \caption{\mysize $\hat{\V{J}}^{2D}$}
        \label{fig:qual_rgb}
    \end{subfigure}
    \begin{subfigure}[b]{0.13\linewidth}
        \includegraphics[trim={\cropparamhorz{} \cropparamvert{} \cropparamhorz{} \cropparamvert{}},clip,width=\linewidth]{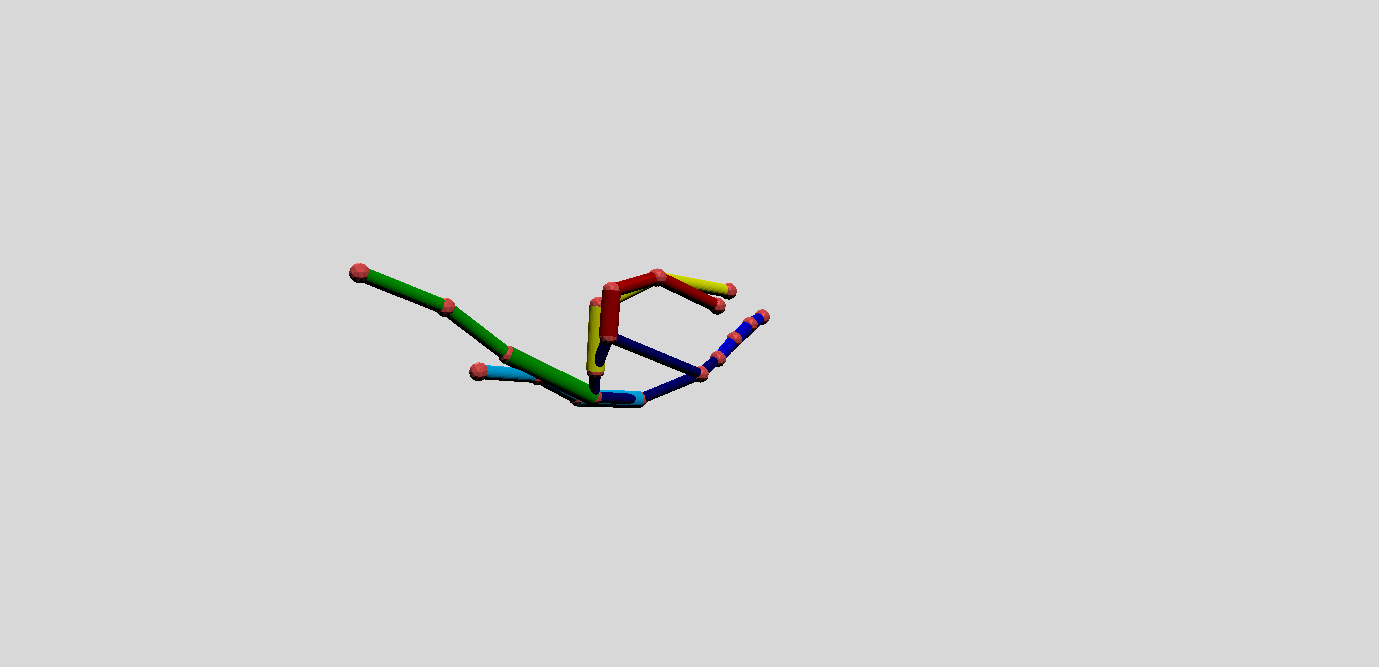}
        \caption{{\mysize w/o $\mathrm{\mathbf{BMC}}$}}
        \label{fig:no_BMC_1}
    \end{subfigure}
    \begin{subfigure}[b]{0.13\linewidth}
        \includegraphics[trim={\cropparamhorz{} \cropparamvert{} \cropparamhorz{} \cropparamvert{}},clip,width=\linewidth]{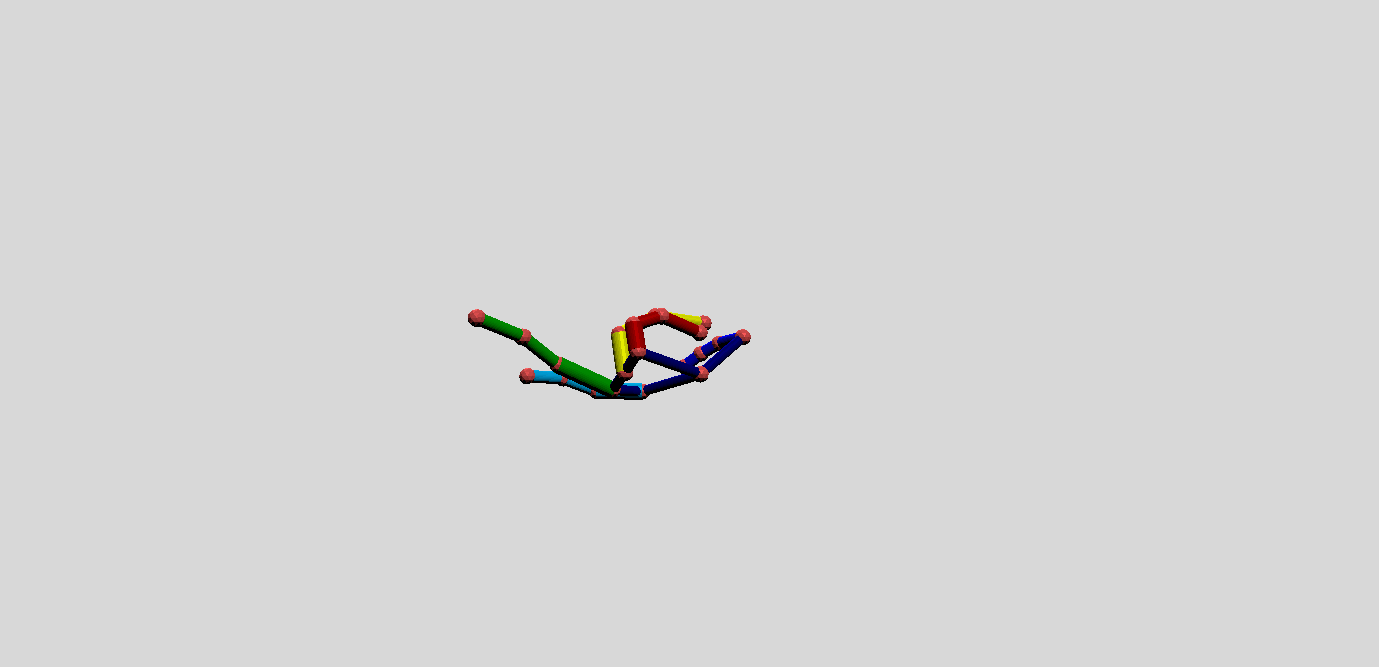}
        \caption{{\mysize w. $\mathrm{\mathbf{BMC}}$}}
        \label{fig:with_BMC_1}
    \end{subfigure}
    \begin{subfigure}[b]{0.13\linewidth}
        \includegraphics[trim={\cropparamhorz{} \cropparamvert{} \cropparamhorz{} \cropparamvert{}},clip,width=\linewidth]{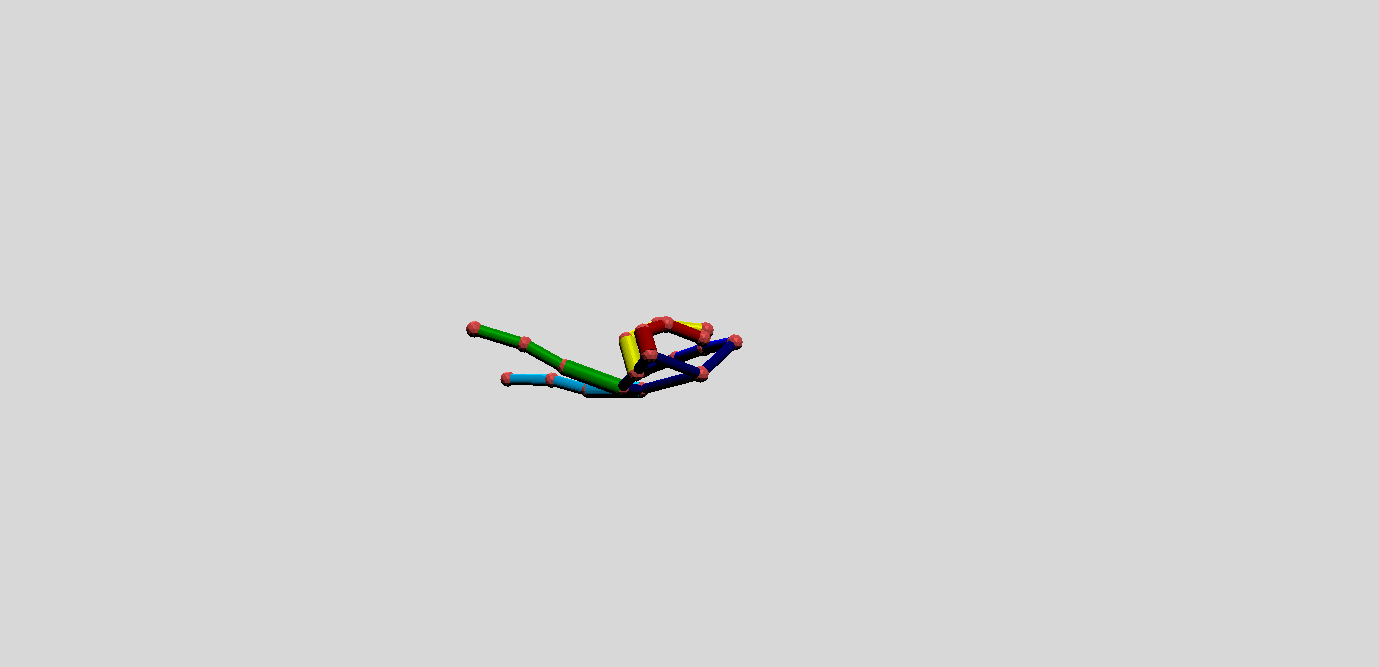}
        \caption{\mysize GT}
        \label{fig:gt_1}
    \end{subfigure}
    \ \ \ %
    \begin{subfigure}[b]{0.13\linewidth}
        \includegraphics[trim={\cropparamhorz{} \cropparamvert{} \cropparamhorz{} \cropparamvert{}},clip,width=\linewidth]{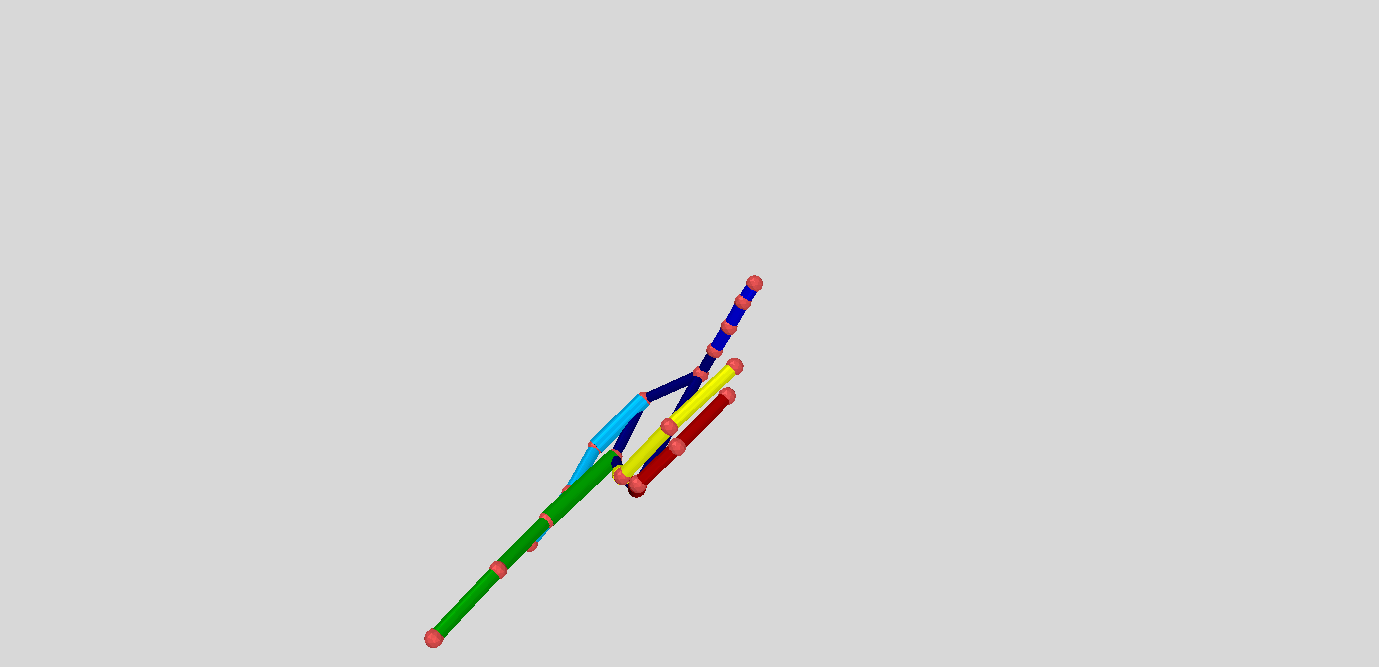}
        \caption{{\mysize w/o $\mathrm{\mathbf{BMC}}$}}
        \label{fig:no_BMC_2}
    \end{subfigure}
    \begin{subfigure}[b]{0.13\linewidth}
        \includegraphics[trim={\cropparamhorz{} \cropparamvert{} \cropparamhorz{} \cropparamvert{}},clip,width=\linewidth]{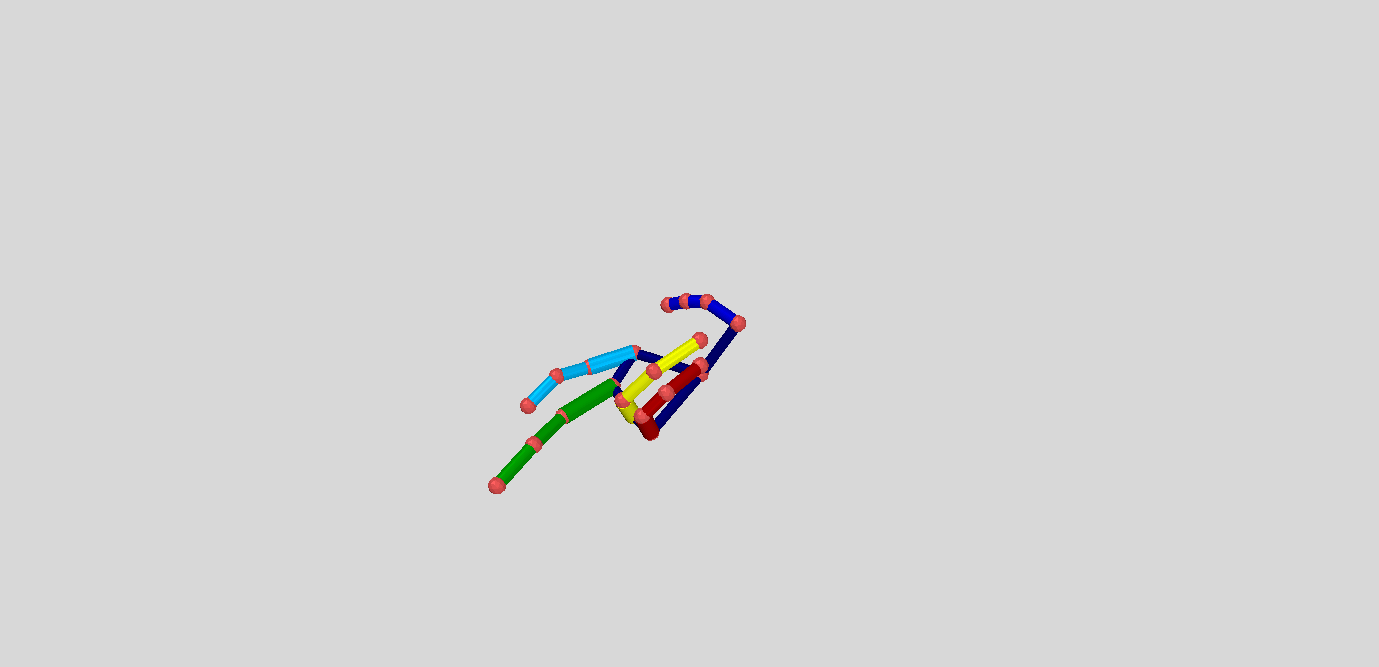}
        \caption{{\mysize w. $\mathrm{\mathbf{BMC}}$}}
        \label{fig:with_BMC_2}
    \end{subfigure}
    \begin{subfigure}[b]{0.13\linewidth}
        \includegraphics[trim={\cropparamhorz{} \cropparamvert{} \cropparamhorz{} \cropparamvert{}},clip,width=\linewidth]{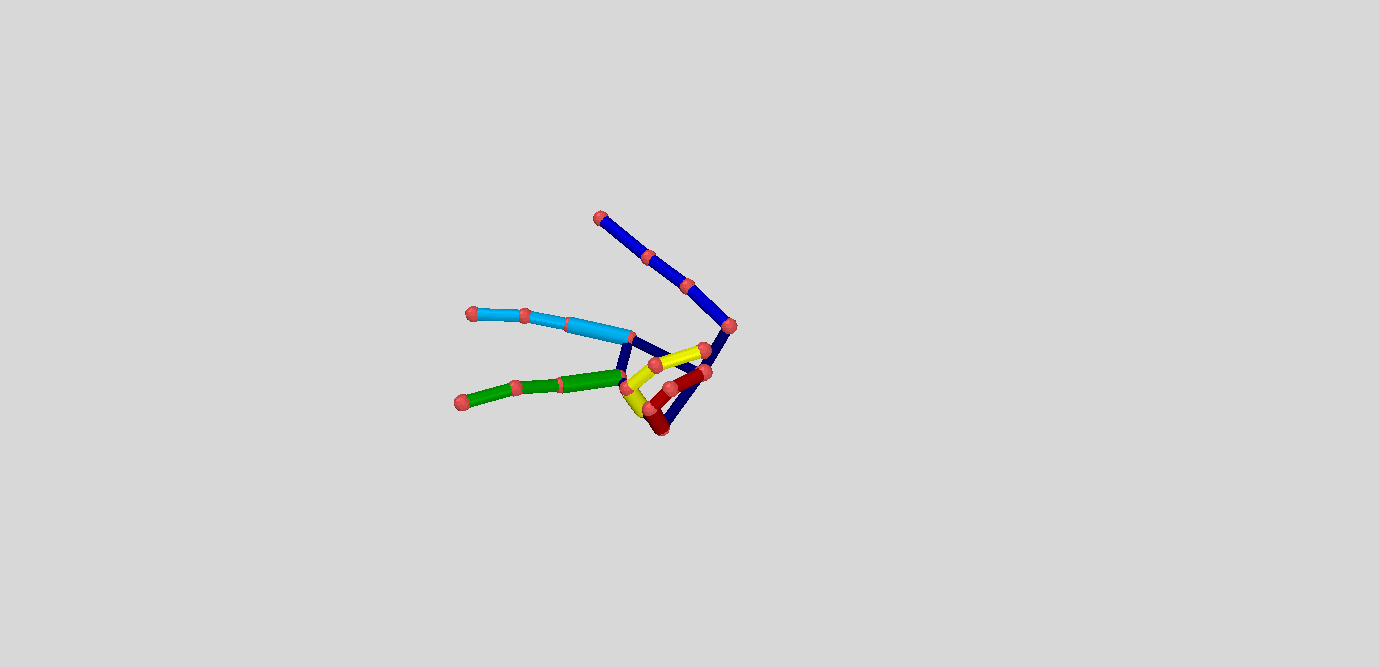}
        \caption{\mysize GT}
        \label{fig:gt_2}
    \end{subfigure}
    
    \caption{Qualitative results of the Boostrapping with Synthetic Data experiment. Testing performed on custom split of FH. \figref{fig:qual_rgb}: We see that the model trained without BMC (green), as well as the model trained with BMC (red), perform equally well on the 2D prediction task. \figref{fig:gt_1}, \figref{fig:gt_2} show the ground-truth joint skeleton from the camera view, as well as the "top" view looking down, respectively. \figref{fig:no_BMC_1} and \figref{fig:no_BMC_2} show the 3D predictions of the model trained fully supervised on RHD and weakly-supervised on FH. Despite the accurate 2D predictions, the 3D pose is incorrect, displaying implausible bio-physical poses. \figref{fig:with_BMC_1} and \figref{fig:with_BMC_2} show the result of incorporating BMC into the model. The predictions are kinematically and structurally sound, and as a result closer to the ground-truth predictions.}
    \label{fig:qualitative_results}
\end{figure*}

We show qualitative results of the Bootstrapping with Synthetic Data experiment in \figref{fig:qualitative_results}. We display the predicted $\V{J}^{3D}$ of both $\mathbf{3D_\mathrm{RHD} + 2D_\mathrm{FH}}$ (w/o $\mathbf{\mathrm{BMC}}$) and $\mathbf{3D_\mathrm{RHD} + 2D_\mathrm{FH}} + \mathcal{L}_\mathbf{\mathrm{BMC}}$ (w. $\mathbf{\mathrm{BMC}}$). Two views are shown. The first displays the view from the front or camera view (looking in direction of the $z$-axis), the second shows the view from the top of the world space, looking down (looking in the \textit{opposite} direction of the $x$-axis). Additionally, we plot the 2D predictions of both models, where green corresponds to without BMC and red is the model using BMC. 

We see that despite both models predicting accurately the 2D pose, its predicted 3D pose are different. Not using BMC, the model predicts bio-physically implausible poses. This is due to unseen 3D poses, views and occlusions. Additionally, the 3D component of the model has only been trained on synthetic data. For example, RHD does not contain object occlusions or ego-centric views. Using BMC, our model can better adapt its depth-component during training to these unseen 3D poses, resulting in more accurate predictions.

\section{Architecture and training}
We use a standard ResNet-50 network for our backbone. We replace the last linear layer to output a $21 \times 3$ dimensional vector. The first two dimensions correspond to the $2D$ keypoints, whereas the last layer corresponds to the root-relative depth $Z^\mathrm{r}$.

Our $Z^\mathrm{root}$ refiner consists of a three layered MLP, using leaky ReLU non-linearity. We used BatchNorm in between all layers except the last. For the cross-dataset evaluation, we empirically found that not using BatchNorm resulted in better accuracy. The exact architecture using BatchNorm is listed in \tabref{tbl:refinement_arch}.

The network was trained for 70 epochs using SGD with a learning rate of $5e\minus3$ and a step-wise learning rate decay of $0.1$ after every 30 epochs.

We set the weight values as follows: $\lambda_{2D} = 1$, $\lambda_{Z^\mathrm{r}}=5$, $\lambda_{Z^\mathrm{root}}=1$. For all experiments using BMC, we set the individual weights of the losses as follows: $\lambda_{\mathrm{BL}} = 0.1$, $\lambda_{\mathrm{RB}}=0.1$, $\lambda_{\mathrm{A}}=0.01$

\begin{table}[t]
\centering
\caption{Architecture of the refinement network. It takes the predicted and calculated values $\V{z}^r \in \R^{21}$, $\V{K}^{-1}\V{J}^{2D} \in \R^{21\times 3}$, $Z^{root} \in \R$ and outputs a residual term $r$ such that $\hat{Z}^{root}_\mathrm{ref} = \hat{Z}^{root} + r$}
\label{tbl:refinement_arch}
\begin{tabular}{|c|}
\hline
\textbf{Refinement Network}         \\ \hline
Linear($85$, $128$)  \\ \hline
LeakyReLU($0.01$) \\ \hline
BatchNorm  \\ \hline
Linear($128$, $128$)  \\ \hline
LeakyReLU($0.01$) \\ \hline
BatchNorm  \\ \hline
Linear($128$, $1$)  \\ \hline
\end{tabular}
\end{table}

\section{HANDS2019 challenge}
The HANDS2019 challenge\footnote{\url{https://competitions.codalab.org/competitions/21116}} was organized to evaluate cutting edge methods for 3D hand pose estimation. The rules of challenge task \#3 required us to train solely on the HO-3D dataset. We trained the proposed model without auxiliary losses. The refinement step was vital for achieving the first place of the competition, demonstrating the performance of the underlying backbone model.
\begin{table}[t]
\centering
\caption{HANDS2019 challenge results on the \textbf{test split} of HO-3D, as evaluated by the \textit{online submission system}. All methods were trained only on HO-3D. We show the top four submission. The winner was selected based on the extrapolation score. Results are given in mm.}
\label{tbl:hands2019_challenge}
\begin{tabular}{lcccc}
\toprule
HO-3D                   & EXTRAP $\downarrow$   & INTERP $\downarrow$   & OBJECT $\downarrow$    & SHAPE $\downarrow$ \\
\midrule
Ours                                    & \textbf{24.74}     & 6.70      &   27.36   &   \textbf{13.21}   \\ 
Nplwe                                   & 29.19     & \textbf{4.06}      &   \textbf{18.39}   &   15.79   \\ 
lin84                                   & 31.51     & 19.15     &   30.59   &   23.47   \\ 
Hasson~et~al.~\cite{hasson2019learning} & 38.42     & 7.38      &   31.82   &   15.61   \\ 
\bottomrule
\end{tabular}

\end{table}

\end{document}